\documentclass{article}
\usepackage{authblk}

\PassOptionsToPackage{numbers, compress}{natbib}

\usepackage[final]{neurips_2022}

\usepackage{preamble}

\usepackage[numbers, compress]{natbib}
\usepackage[utf8]{inputenc} 
\usepackage[T1]{fontenc}    
\usepackage{hyperref}       
\usepackage{url}            
\usepackage{booktabs}       
\usepackage{amsfonts}       
\usepackage{nicefrac}       
\usepackage{microtype}      
\usepackage{xcolor}         
\usepackage{graphicx}
\usepackage{subcaption}
\usepackage[capitalize,noabbrev]{cleveref}
\usepackage{wrapfig}
\usepackage{comment}
\usepackage{enumitem}

\newcommand{\x}[1]{\mathbf{x}_{#1}}
\newcommand{\X}[1]{\mathbf{X}_{#1}}

\newcommand{\Nx}[1]{\hat{\mathbf{x}}_{#1}} 

\def\eg{\emph{e.g.}}

\title{Understanding the Failure of Batch Normalization for Transformers in NLP}

\author{\textbf{Jiaxi Wang$^1$, \ Ji Wu$^{1, 2}$, \ Lei Huang$^3$} \\
$^1$Department of Electronic Engineering, Tsinghua University \\
$^2$Institute for Precision Medicine, Tsinghua University \\
\texttt{\{wjx20@mails, wuji\_ee@mail\}.tsinghua.edu.cn} \\
$^3$SKLSDE, Institute of Artificial Intelligence, Beihang University \\
\texttt{huangleiAI@buaa.edu.cn}
}
\begin{document}

\maketitle

\begin{abstract}
  Batch Normalization (BN) is a core and prevalent technique in accelerating the training of deep neural networks and improving the generalization on Computer Vision (CV) tasks. 
However, it fails to defend its position in Natural Language Processing (NLP), which is dominated by Layer Normalization (LN). 
In this paper, we are trying to answer why BN usually performs worse than LN in NLP tasks with Transformer models. 
We find that the inconsistency between training and inference of BN is the leading cause that results in the failure of BN in NLP. We define Training Inference Discrepancy (TID) to quantitatively measure this inconsistency and reveal that TID can indicate BN's performance, supported by extensive experiments, including image classification, neural machine translation, language modeling, sequence labeling, and text classification tasks. We find that BN can obtain much better test performance than LN when TID keeps small through training.
To suppress the explosion of TID, we propose Regularized BN (RBN) that adds a simple regularization term to narrow the gap between batch statistics and population statistics of BN.
RBN improves the performance of BN consistently and outperforms or is on par with LN on 17 out of 20 settings, involving ten datasets and two common variants of Transformer\footnote{Our code is available at \url{https://github.com/wjxts/RegularizedBN}}. 
\end{abstract}

\section{Introduction}
Deep learning~\cite{Lecun2015DL} has revolutionized Computer Vision (CV)~\cite{Kri2012AlexNet} and Natural Language Processing (NLP)~\cite{Vaswani2017Transformer}. Normalization layers are key components to stabilize and accelerate the training in Deep Neural Networks (DNNs). In CV, Batch Normalization (BN)~\cite{Ioffe2015BN} is the default normalization technique and reveals superior performance over other normalization techniques in image recognition tasks by enforcing the input of a neuron to have zero mean and unit variance within a mini-batch data. Furthermore, a growing number of theoretical works analyze the excellent properties of BN in benefiting optimization~\cite{Ioffe2015BN,San2018BnHelpOpt,Bjorck2018BNLargeLr,Huang2020BNDynamic,Dane2020BNRank1,Dane2021BNRank2}. 
While BN almost dominates in CV with empirical success and theoretical properties,
 Layer Normalization (LN) is the leading normalization technique in NLP, especially for Transformer models that achieve the state-of-the-art performance on extensive tasks, including machine translation~\cite{Vaswani2017Transformer}, natural language understanding~\cite{Devlin2019Bert}, text generation~\cite{Radford2019GPT2}, few shot learning~\cite{Brown2020GPT3}, to name a few.
 As a direct substitute of LN, BN performs poorly in Transformer for neural machine translation~\cite{Shen2020Powernorm}. It remains elusive to explain the failure of BN in NLP community. In this work, we are trying to take a step forward. 
\medskip
Our contributions are summarized as follows:
\vspace{-0.12in}
\begin{itemize}[noitemsep, nolistsep]
    \item We find that the inconsistency between training and inference leads to the failure of BN in NLP, supported by our extensive experiments, including image classification, neural machine translation, language modeling, sequence labeling, and text classification tasks. 

   \item  We define Training Inference Discrepancy (TID) to quantitatively measure this inconsistency and show that TID can serve as an indicator of BN's performance. In particular, BN reaches much better test performance than LN when TID keeps small through training, \eg, in image recognition and language modeling tasks.

	\item We propose Regularized BN (RBN) that adds a regularization term in BN to penalize and reduce the TID when the TID of BN is large. We reveal the optimization advantages of RBN over LN  by exploring the layer-wise training dynamics of Transformer. 
	\item We empirically show that RBN can exceed or match the performance of LN, sometimes with a large margin, on 17 out of 20 settings, involving ten datasets and two common variants of Transformer. Besides, RBN introduces no extra computation at inference compared to LN. 
	\vspace{-0.12in}
\end{itemize}

\section{Related Work}
\label{related_work}
\vspace{-0.12in}
\paragraph{Analyses of BN's Success}
As BN becomes an indispensable component in deep neural networks deployed in CV tasks, a bunch of works explore the theoretical reasons behind its success. From the view of optimization, the original BN paper~\cite{Ioffe2015BN} argues that BN can reduce internal covariate shift and thus stabilize the training, while~\citet{San2018BnHelpOpt} debate that BN could smooth the loss landscape and thus enable training of neural network with larger learning rate~\cite{Bjorck2018BNLargeLr}. \citet{Dane2020BNRank1,Dane2021BNRank2} prove that a stack of randomized linear layers and BN layers will endow the intermediate features of neural network with sufficient numerical rank as depth increases, which is beneficial for optimization and learning discriminative hierarchical features. \citet{Huang2020BNDynamic} show that BN could improve the
layer-wise conditioning of the neural network optimization by exploring the spectrum of Hessian matrix with block diagonal approximation~\cite{Marten2015KFAC}. From the view of generalization, \citet{Ioffe2015BN,Luo2018BNReg,Li2019BNDropout,Wu2021RevistBN} argue that BN serves as regularizer which reduces over-fitting when its stochasticity is small and may have detrimental effect when it is large~\cite{Wu2021RevistBN}. \citet{Huang2019IterNorm} further propose Stochastic Normalization Disturbance (SND) to measure such stochasticity and shows that large SND will hinder the training of neural networks.
\vspace{-0.12in}
\paragraph{Training Inference Inconsistency of BN}
Normalizing along the batch dimension usually introduces training inference inconsistency since mini-batch data is neither necessary nor desirable during inference. BN uses population statistics, estimated by running average over mini-batch statistics, for inference. The training inference inconsistency usually harms the performance of BN for small-batch-size training since the estimation of population statistics could be inaccurate~\cite{2018_ECCV_Wu}. One way to reduce the inconsistency between training and inference is to exploit the estimated population statistics for normalization during training~\cite{2017_NIPS_Ioffe,2019_NeurIPS_Chiley,2020_ICLR_Yan,2020_ECCV_Yong,2021_CVPR_Yao}.
These works may outperform BN when the batch size is small, where inaccurate estimation may be the main issue~\cite{Ioffe2015BN,2018_UAI_Izmailov}, but they usually work inferior to BN under moderate batch-size training~\cite{2019_ICLR_Luo}.  
  Another way to reduce the inconsistency is estimating corrected normalization statistics during inference only, either for domain adaptation~\cite{2017_arxiv_Li}, corruption robustness~\cite{2020_NIPS_Schneider,2020_arxiv_Nado,2021_WACV_Benz}, or small-batch-size training ~\cite{2019_ICCV_Singh,2020_ICLR_Summers}.
 We note that a recent work~\cite{2022_CVPR_Huang} investigates the estimation shift problem of BN. Unlike this work that addresses the accumulated estimation shift due to the stack of BNs for CNNs in CV tasks, our work pays more attention to how the training inference inconsistency of BN correlates with its performances for Transformers in NLP tasks. Besides, the estimation shift of BN defined in~\cite{2022_CVPR_Huang}, which addresses the differences between the estimated population statistics and the expected statistics, differs from our TID of BN that addresses the differences between the mini-batch statistics and populations statistics.  
 
\vspace{-0.12in}
\paragraph{Exploring the Failure of BN in Transformer}
Similar to our work, Power Normalization (PN)~\cite{Shen2020Powernorm} also investigates the reason behind the failure of BN in Transformers. Our work significantly differs from PN~\cite{Shen2020Powernorm} in the following facets. PN attributes the failure of BN to the unstable training of BN incurred by fluctuated forward and backward batch statistics with outlier values, while we observe that the training of BN is as good as LN and the inconsistency between training and inference of BN matters more. Based on our observation, we propose a regularization term to reduce the TID of BN. Compared with PN, which incorporates a layer-scale layer (root mean square layer normalization~\cite{Zhang2019RMSNorm} without affine transformation~\cite{Xu2019UnderstandLN}), our method introduces no extra computation at inference. Besides, we use a more reasonable index to measure inconsistency which is invariant to the scale of data. Furthermore, we show that our RBN can improve the layer-wise training dynamics of LN, which reveals the optimization advantages of RBN.

\section{Analyses of Training Inference Inconsistency in $\text{Transformer}_{BN}$}
\subsection{Preliminary}
\label{ssec:prelinimary}

\begin{figure}[t]
	\captionsetup[subfigure]{justification=centering}
	\centering
	\begin{subfigure}[b]{0.9\columnwidth}
		\centering
		\centerline{\includegraphics[width=\textwidth]{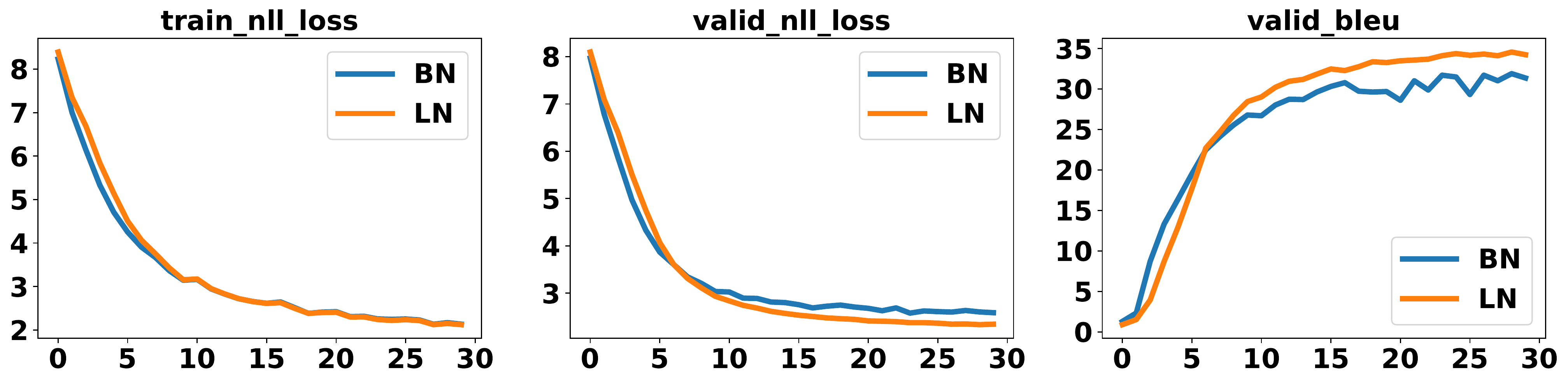}}
	\end{subfigure}
	\caption{Train loss, validation loss/BLEU of Transformer trained on IWSLT14 with BN and LN. The training of $\text{Transformer}_{BN}$ is better than $\text{Transformer}_{LN}$ while the validation loss/BLEU of $\text{Transformer}_{BN}$ underperforms that of $\text{Transformer}_{LN}$ after 8 epoch. At the end of the training, $\text{Transformer}_{BN}$ falls behind $\text{Transformer}_{LN}$ with large BLEU scores. Lower loss and higher BLEU scores indicate better performance. Based on the inconsistency of training and validation performance of BN, we hypothesize that the training inference discrepancy of BN causes its performance degradation.}
	\label{fig:loss}
\end{figure}

Batch Normalization (BN)~\cite{Ioffe2015BN} is typically used to stabilize and accelerate DNN's training. 
Let $\x{} \in \mathbb{R}^d$ denote the $d$-dimensional input to a neural network layer. During training, BN  standardizes each neuron/channel within $m$ mini-batch data by\footnote{BN usually uses extra learnable scale and shift parameters~\cite{Ioffe2015BN} to recover the potentially reduced representation capacity, and we  omit them since they are not relevant to our discussion.}
		\begin{equation}
			\label{eqn:BN}
			\Nx{j}=BN_{train}(\x{j})=  \frac{\x{j} - \mu_{B,j}}{\sqrt{\sigma^2_{B,j}}}, ~~j=1,2, ..., d,
		\end{equation}
where $\mu_{B,j}=\frac{1}{m}  \sum_{i=1}^{m}  \x{j}^{(i)}$ and $\sigma^2_{B,j} = \frac{1}{m}   \sum_{i=1}^{m} (\x{j}^{(i)}-\mu_{B,j})^2 $ are the mini-batch mean and variance for each neuron, respectively. Note that an extra small number $\epsilon$ is usually added to the variance in practice to prevent numerical instability.
	During inference, the population mean $\mu$ and variance $\sigma^2$ of the layer input are required for BN to make a deterministic prediction~\cite{Ioffe2015BN} as:
	\begin{small}
		\begin{equation}
			\label{eqn:BN-inf}
			\Nx{j}=BN_{inf}(\x{j})= \frac{\x{j} - \mu_{j}}{\sqrt{\sigma^2_j}}, ~~j=1,2, ..., d.
		\end{equation}
	\end{small}These population statistics  $\{\mu, \sigma^2\}$ are usually calculated as the running
average of mini-batch statistics over different training iteration $t$ with an update factor $\alpha$ as follows:
		\begin{equation}
			\label{eqn:running-average}
			\begin{aligned}
				\begin{cases}
					\quad \mu^{(t)}=  (1-\alpha) \mu^{(t-1)} + \alpha \mu^{(t)}_B,\\
					\quad (\sigma^2)^{(t)}=  (1-\alpha)(\sigma^2)^{(t-1)} + \alpha (\sigma^2_B)^{(t)}.
				\end{cases}
			\end{aligned}
		\end{equation}
The discrepancy of BN for normalization during training (using Eqn.~\ref{eqn:BN}) and inference (using Eqn.~\ref{eqn:BN-inf}) can produce stochasticity, since the population statistics of BN are estimated from the mini-batch statistics that depend on the sampled mini-batch inputs. This discrepancy is believed to benefit the generalization~\cite{Ioffe2015BN, Huang2019IterNorm} if the stochasticity is well controlled. However, this discrepancy usually harms the performance of small-batch-size training~\cite{2018_ECCV_Wu} since the estimation of population statistics can be inaccurate. To address this problem, a bunch of batch-free normalizations are proposed that use consistent operations during training and inference, \eg, Layer Normalization (LN)~\cite{Ba2016LN}.
\vspace{-0.12in}
\paragraph{Basic Observations}
\begin{figure}[t]
	\captionsetup[subfigure]{justification=centering}
	\centering
	\begin{subfigure}[b]{0.48\columnwidth}
		\centering
		\centerline{\includegraphics[width=\textwidth]{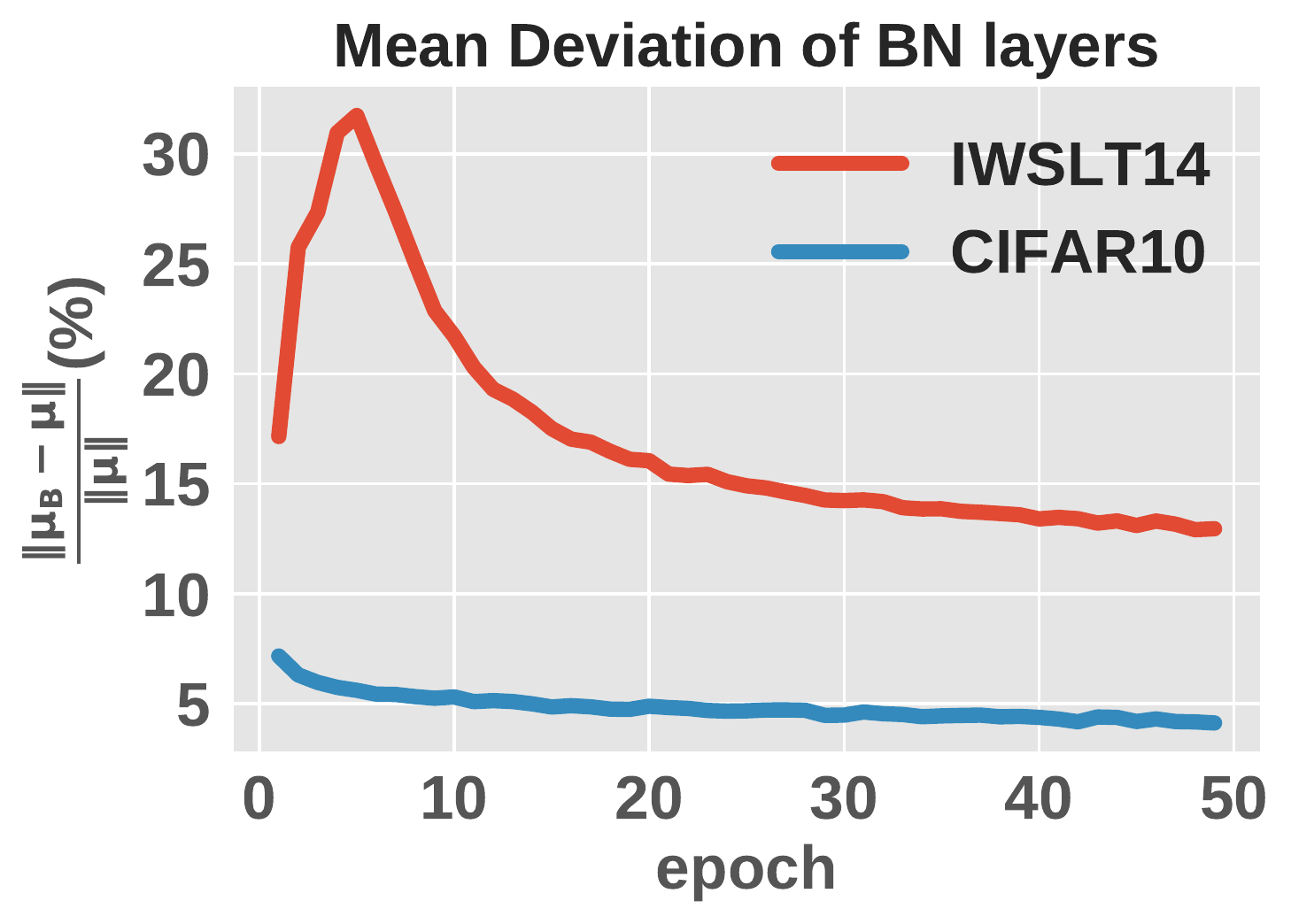}}
	\end{subfigure}
	\hfill
	\begin{subfigure}[b]{0.48\columnwidth}
		\centering
		\includegraphics[width=\textwidth]{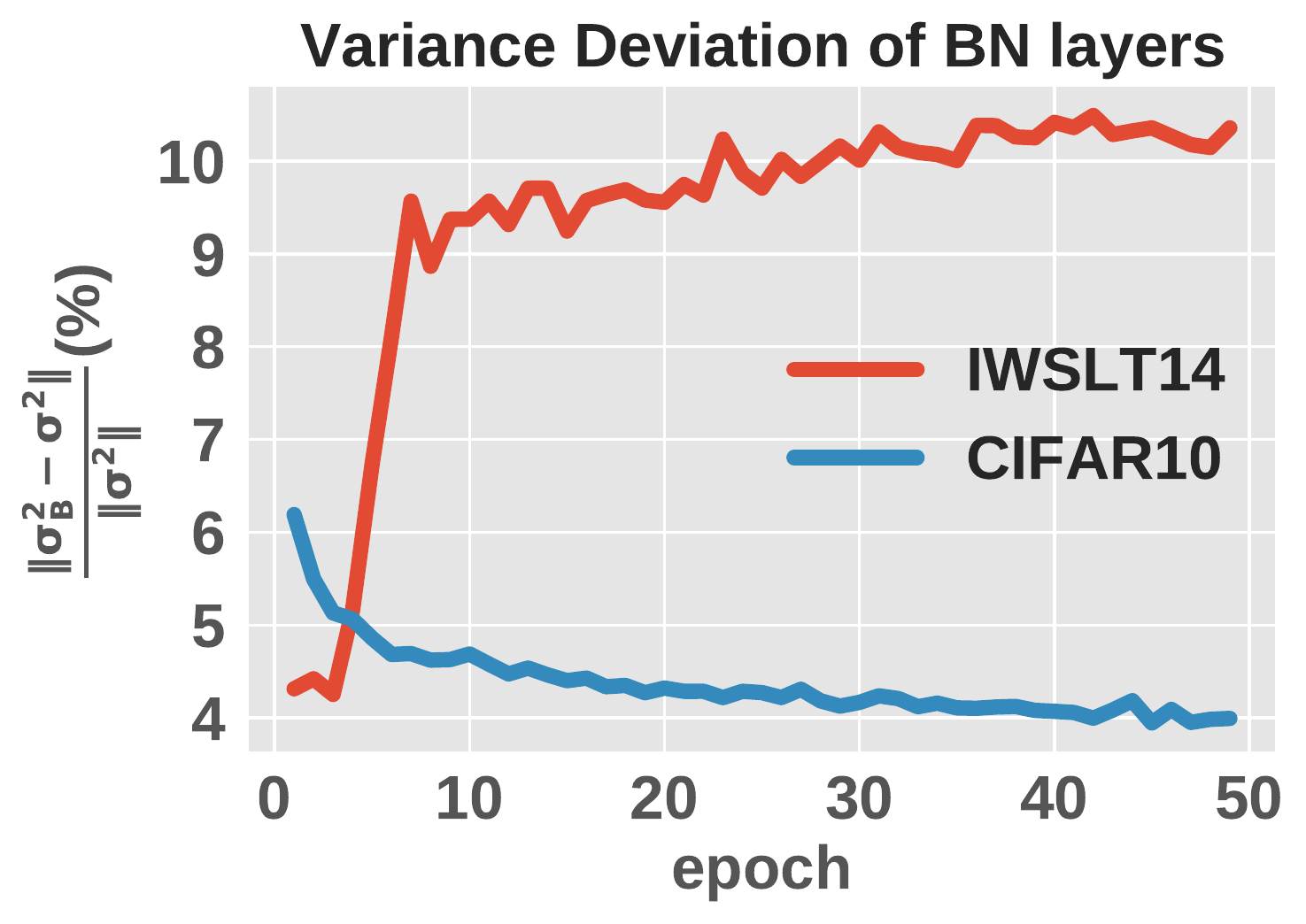}
	\end{subfigure}

\begin{subfigure}[b]{0.48\columnwidth}
	\centering
	\centerline{\includegraphics[width=\textwidth]{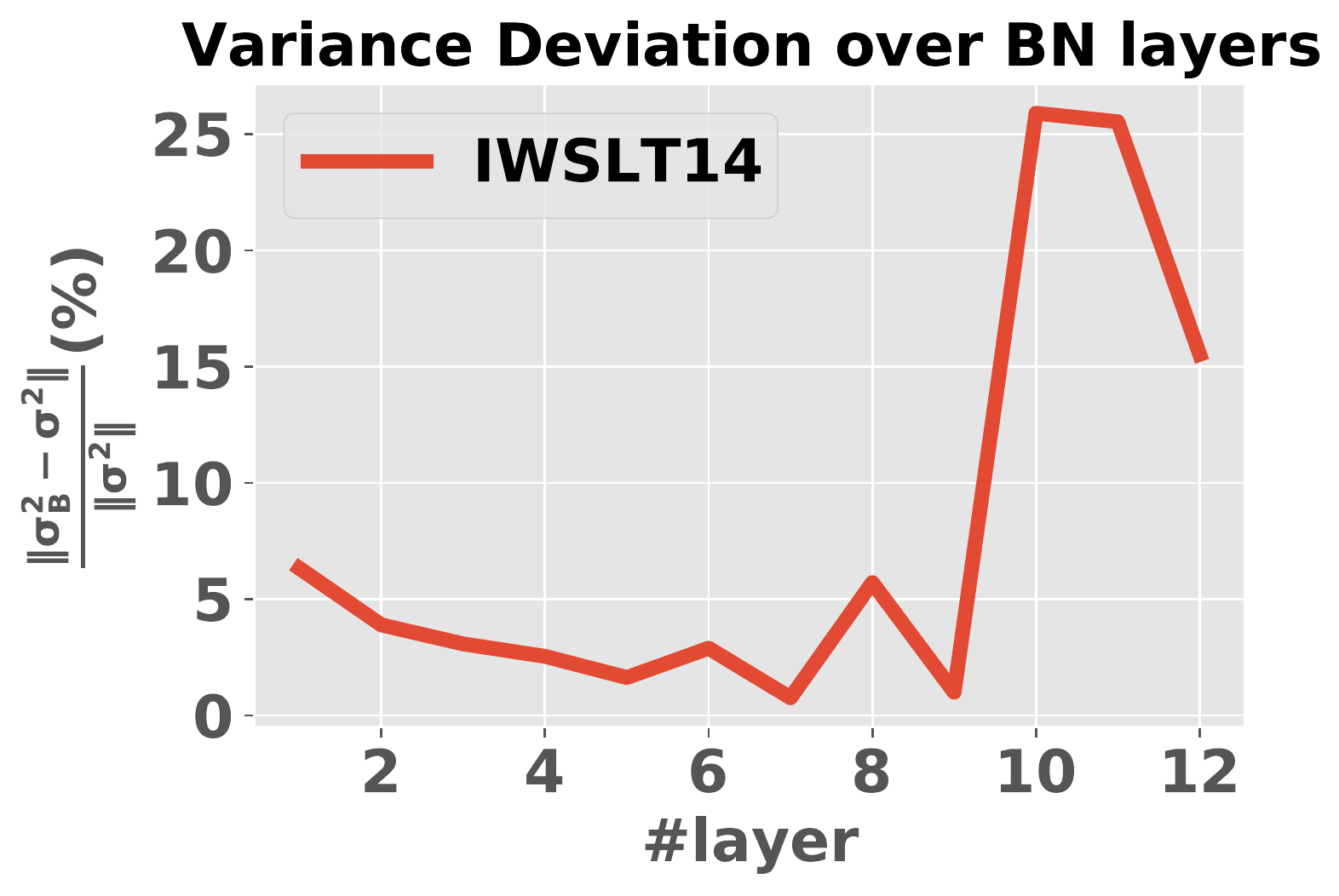}}
\end{subfigure}
\hfill
\begin{subfigure}[b]{0.48\columnwidth}
	\centering
	\includegraphics[width=\textwidth,height=0.7\textwidth]{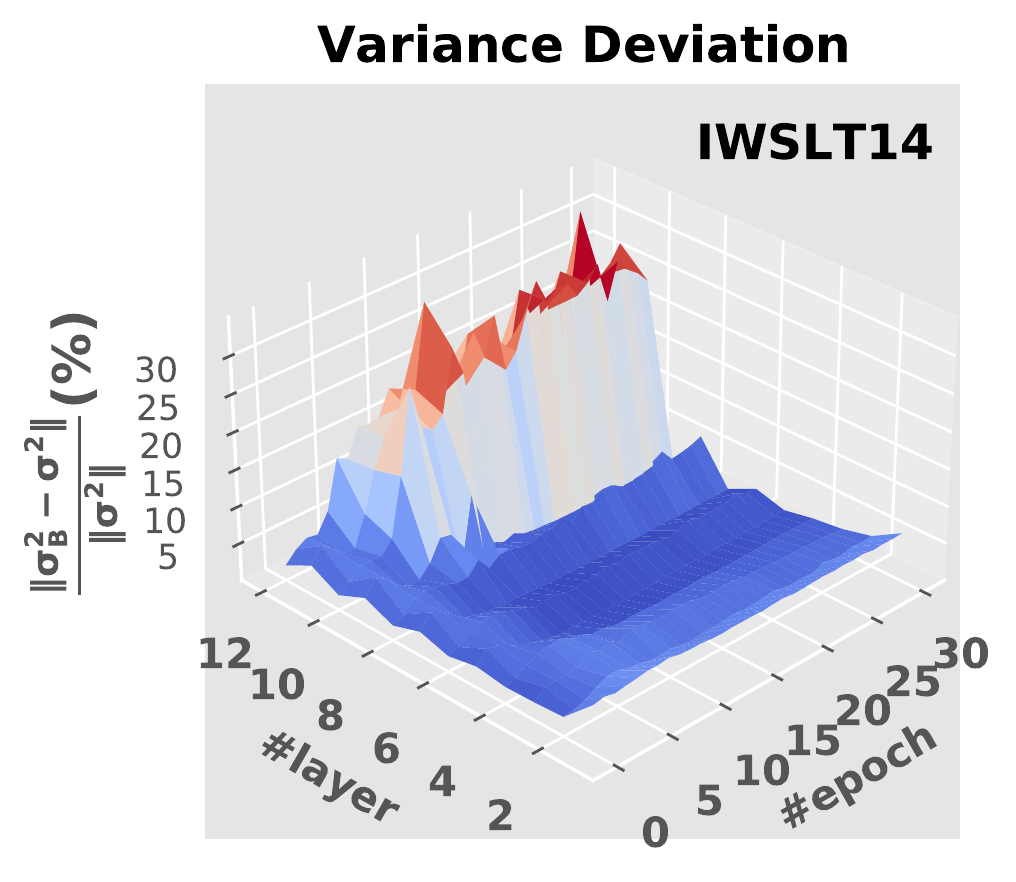}
\end{subfigure}
	\caption{Top: The average deviation of batch mean $\mu_B$ (left figure) and batch variance $\sigma_{B}^2$ (right figure) to population mean $\mu$ and population variance $\sigma^2$ of all BN layers through training in ResNet18 and $\text{Transformer}_{BN}$. There are 21 BN layers in ResNet18 and 12 BN layers in the encoder of $\text{Transformer}_{BN}$. At the end of training, ResNet18 has mean/variance deviation of around $4\%$/$4\%$  and those in $\text{Transformer}_{BN}$ are around $11\%$/$13\%$. Large deviation of statistics hurts the performance of $\text{Transformer}_{BN}$. Bottom: Variance deviation of BN layers with different depths (left) at the end of training and variance deviation over depth and training progress (right).}
	\label{fig:average_dev_cifar_iwslt}
\end{figure}
To analyze the failure of BN in NLP tasks, 
we first plot the training loss and validation loss/BLEU~\cite{Pap2002Bleu} of BN and LN on IWSLT14 (De-En) dataset with the original Transformer model (see \cref{fig:loss}). We observe that the training of $\text{Transformer}_{BN}$ is faster than $\text{Transformer}_{LN}$. The training nll\_loss of BN is even smaller than that of LN, especially at the beginning. However, validation loss/BLEU of BN is worse than that of LN after around the seventh epoch. This phenomenon can not be attributed to over-fitting since BN introduces more stochasticity than LN in the training phase. The inconsistency between training and inference of BN may play a role. \par
Since BN in ResNet18 also involves training inference inconsistency, we guess the degree of such inconsistency has a difference between ResNet18 and $\text{Transformer}_{BN}$. Therefore, we plot the deviation of batch statistics to population statistics of BN in ResNet18 and $\text{Transformer}_{BN}$ in \cref{fig:average_dev_cifar_iwslt} (top) to make a comparison. ResNet18 is trained on CIFAR-10~\cite{Krizhevsky2009CIFAR10} and accuracy will drop 2 percent if we replace BN with LN. We find that at the end of the training, $\text{Transformer}_{BN}$ has a much bigger mean and variance deviation than ResNet18. Besides, the last several BN layers that are close to the output in $\text{Transformer}_{BN}$ have large variance deviation (\cref{fig:average_dev_cifar_iwslt} (bottom)), which negatively impact the model output. Furthermore, the performance degradation of $\text{Transformer}_{BN}$ coincides with the increase of variance deviation by comparing \cref{fig:loss} (right) and \cref{fig:average_dev_cifar_iwslt} (bottom right). Based on these observations, \textit{we hypothesize that the inconsistency between training and inference of BN causes BN's performance degradation in neural machine translation}. We first mathematically define the training inference discrepancy of BN in the next subsection.
\vspace{-0.12in}
\subsection{Training Inference Discrepancy}

By observing Eqns.~\ref{eqn:BN} and~\ref{eqn:BN-inf}, the normalized output during training can be calculated as:
	\begin{equation}
	 \frac{\x{j} - \mu_{B,j}}{\sigma_{B,j}}= \left(\frac{\x{j}-\mu_j}{\sigma_j}+\frac{\mu_j-\mu_{B,j}}{\sigma_j} \right) \frac{\sigma_j}{\sigma_{B,j}}, ~~j=1,2, ..., d,
	\label{eq:decom_tid}
\end{equation}
where $\sigma_{B,j}>0$ and $\sigma_j>0$ are the standard deviation for the $j$-th dimension. We can see $\frac{\mu_j-\mu_{B,j}}{\sigma_j}$ and $\frac{\sigma_j}{\sigma_{B,j}}$ can be viewed as random variables. Their magnitude can characterize the diversity of mini-batch examples during training and indicate how hard the estimation of population statistics is. We thus define the training inference discrepancy to quantitatively measure the inconsistency as follows. 
\begin{definition}[Training Inference Discrepancy (TID)] Let $p_B$ be the distribution of batch data. Given a mini-batch data $X$
sampled from $p_B$,	we define the TID of its mean and variance (with respect to model parameter $\theta$) as: 
	\begin{equation}
	\begin{aligned}
	\text{Mean TID} & = \Expect_{X \sim p_{B} } \frac{\Vert \mu_{B} -  \mu \Vert_2}{\Vert \sigma \Vert_2 } \\
	\text{Variance TID} & = \Expect_{X \sim p_{B} } \frac{\Vert \sigma_{B}-\sigma \Vert_2}{\Vert \sigma \Vert_2 } \\
	\end{aligned}
	\label{eq:tid}
	\end{equation}
\end{definition}

In terms of computing the TID in practice, we add a small positive constant in the denominator to avoid numerical instability. We save the checkpoint at the end of each epoch and before training. We first estimate the population statistics by running forward propagation one epoch and then compute mean and variance TID by another epoch.

We omit $\theta$ when it can be inferred from context without confusion. We compute the average mean and variance TID of all BN layers in ResNet18 trained on CIFAR10 and that of $\text{Transformer}_{BN}$ trained on IWSLT14 throughout training. At the end of the training, the average mean/variance TID of BN in ResNet18 is approximately $0.8\%$/$0.9\%$, while that in Transformer is around $2.8\%$/$4.1\%$. TID in Transformer is much larger than that in ResNet18. The trends are the same as basic observations in \cref{ssec:prelinimary}. We will use \cref{eq:tid} to compute TID in the subsequent analysis due to its better theoretical formulation (\cref{eq:decom_tid}).
\subsection{Comprehensive Validation}
To further verify our hypothesis that large inconsistency between training and inference of BN causes BN's degraded performance, we conduct experiments on Neural Machine Translation (NMT), Language Modeling (LM), Named Entity Recognition (NER), and Text Classification (TextCls) tasks. We test both Post-Norm~\cite{Vaswani2017Transformer} and Pre-Norm~\cite{Xiong2020NoWarmUp} Transformers. 
\vspace{-0.12in}
\paragraph{Experimental Setup}
We briefly illustrate the experimental settings. More detailed description can be found in \cref{sec:app_exp}. For neural machine translation, we use IWSLT14 German-to-English (De-En) and WMT16 English-to-German (En-De) datasets, following the settings in \citet{Shen2020Powernorm}. Our code is based on \textit{fairseq}~\cite{Ott2019Fairseq}\footnote{https://github.com/pytorch/fairseq. MIT license.}. For language modeling, we conduct experiments on PTB~\cite{Mikolov2011PTB} and WikiText-103 (WT103)~\cite{MerityX2016WikiText}. We follow the experimental settings in \citet{Shen2020Powernorm,Ma2019TensorTrans}. For named entity recognition, we choose CoNLL2003 (English)~\cite{Sang2003CoNLL} and Resume (Chinese)~\cite{Zhang2018Resume} datasets. We mainly follow the experimental settings in \citet{Yan2019TENER}. For text classification, we select one small scale dataset (IMDB)~\cite{Maas2011IMDB} and three large scale datasets (Yelp, DBPedia, Sogou News). We use the code\footnote{https://github.com/declare-lab/identifiable-transformers. Apache-2.0 license.} and follow most configurations in \citet{Bhardwaj2021TextClsTrans}. 


\begin{table}[t]
	\centering
	\caption{Results for performance and TID of last BN layer with Post-Norm (top) and Pre-Norm (bottom) Transformers on four tasks containing ten datasets. We use BLEU scores (\%)/perplexity/F1 score (\%)/accuracy (\%) to measure the model performance on neural machine translation/language modeling/named entity recognition/text classification. "+" ("-") means the bigger (smaller) the better. Post-LN means the Post-Norm Transformer with LN. Performance gap is the difference between performance of BN and LN. Positive (Negative) Performance gap indicates BN performs better (worse) than LN.}
	\resizebox{\columnwidth}{!}{
		\begin{tabular}{@{}ccccccccccc@{}}
			\toprule
			Task                  & \multicolumn{2}{c}{NMT (+)}       & \multicolumn{2}{c}{LM (-)}        & \multicolumn{2}{c}{NER (+)}       & \multicolumn{4}{c}{TextCls (+)}                                   \\ \midrule
			Datasets              & IWSLT14       & WMT16         & PTB           & WT103         & Resume        & CoNLL         & IMDB          & Sogou         & DBPedia       & Yelp          \\ \midrule
			Post-LN               & 35.5 & 27.3 & 53.2          & 20.9          & 94.8 & 91.3          & 84.1          & 94.6          & 97.5          & 93.3          \\
			Post-BN               & 34.0          & 25.0          & 45.9          & 17.2          & 94.5          & 90.9          & 84.0          & 94.3          & 97.5          & 93.3          \\
			Performance Gap                   & -1.5          & -2.3          & 7.3           & 3.7           & -0.3          & -0.4          & -0.1          & -0.3          & 0             & 0             \\
			Mean TID of $\text{BN}_{last}$  & 1.5\%         & 4.2\%         & 0.9\%         & 1.8\%         & 1.7\%         & 4.2\%         & 1.8\%         & 1.8\%         & 2.2\%         & 3.1\%         \\
			Var TID of $\text{BN}_{last}$   & 10.6\%         & 17.9\%        & 1.1\%         & 2.0\%         & 3.7\%         & 9.5\%         & 3.9\%         & 4.3\%         & 3.5\%         & 4.0\%         \\ \midrule \midrule
			Pre-LN                & 35.5          & 27.3 & 54.5          & 24.6          & 94.0 & 91.0 & 84.1          & 94.5          & 97.5 & 93.3          \\
			Pre-BN                & 34.8          & 25.2          & 45.9          & 17.8          & 93.2          & 89.9          & 84.0          & 94.3          & 97.5 & 93.3          \\
			Performance Gap                   & -0.7          & -2.1          & 8.6           & 6.8           & -0.8          & -1.1          & -0.1          & -0.2          & 0             & 0             \\
			Mean TID of $\text{BN}_{last}$  & 3.4\%         & 7.9\%         & 1.6\%         & 2.4\%         & 9.6\%         & 10.0\%        & 2.9\%         & 7.5\%         & 3.9\%         & 12.1\%        \\
			Var TID of $\text{BN}_{last}$   & 4.6\%         & 30.1\%        & 1.7\%         & 2.5\%         & 6.5\%         & 6.4\%         & 6.2\%         & 7.1\%         & 3.3\%         & 8.6\%         \\ \bottomrule
		\end{tabular}
	}
	
	\label{table:basic_result}
\end{table}
\vspace{-0.12in}
\paragraph{Performance Result}
We first verify the inefficiency of BN compared to LN on four natural language tasks. Results for Post-Norm and Pre-Norm Transformers are listed in \cref{table:basic_result}. BN performs much worse than LN on NMT, slightly worse on NER and TextCls tasks, but performs much better on LM. Although BN performs worse in most cases, it has remarkable improvement over LN on LM, raising the question: what contributes to the failure or success of BN? 
\vspace{-0.12in}
\paragraph{Analyzing the Statistics of BN}
\begin{figure}[t]
	\captionsetup[subfigure]{justification=centering}
	\centering
	\begin{subfigure}[b]{0.48\columnwidth}
		\centering
		\centerline{\includegraphics[width=\textwidth]{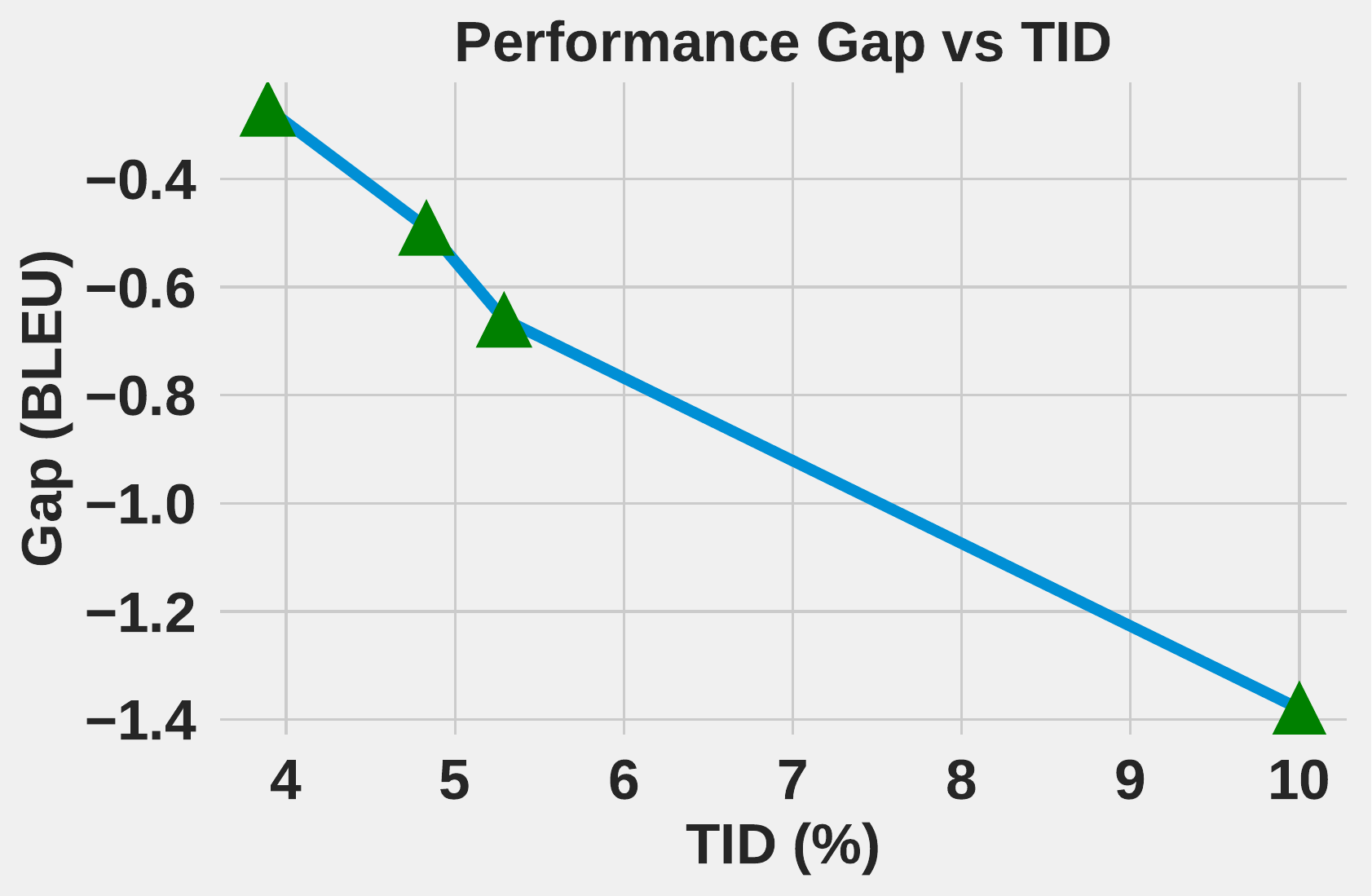}}
	\end{subfigure}
	\hfill
	\begin{subfigure}[b]{0.48\columnwidth}
		\centering
		\includegraphics[width=\textwidth]{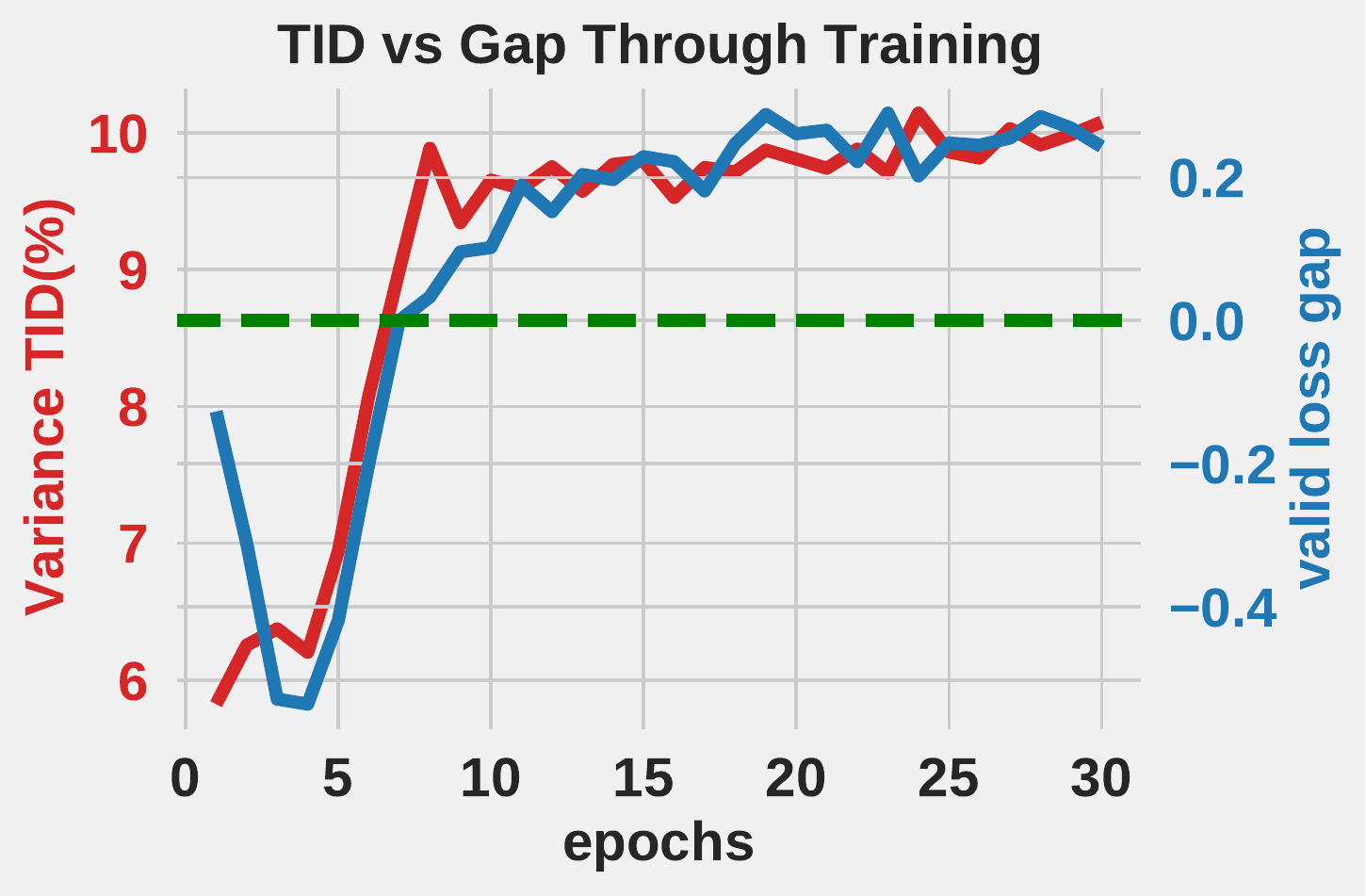}
	\end{subfigure}
	\caption{Left: Variance TID and BLEU gap between $\text{Transformer}_{BN}$ and $\text{Transformer}_{LN}$ when replacing different numbers of LN layers with BN. Right: Variance TID and valid loss gap of Post-Norm Transformer through training.}
	\vspace{-0.15 in}
	\label{fig:tid_vs_gap}
\end{figure}

We compute the TID of the last BN layer in \cref{table:basic_result} and leave the average TID of all BN layers in \cref{sec:train_tid}. The last BN layer, which is close to the output, significantly impacts the model prediction. We observe that TID is highly correlated with the performance gap between BN and LN. When TID is large, e.g., on WMT16, BN performs much worse than LN. However, when the TID of BN is negligible, e.g., on PTB and WT103, BN performs better than LN with a large margin. We select one dataset from each task with Pre-Norm Transformer and define the total TID as the sum of mean and variance TID. At the end of the training, the total TID of the last BN layer for WMT16/CoNLL/IMDB/WT103 is around 38\%/16\%/9\%/5\%, and the performance gap is -2.1 BLEU scores/-1.1 F1 score/-0.1\% accuracy/6.8 perplexity (PPL). Larger TID tends to hurt BN's performance. \par
To explore the quantitative relation between TID and performance gap, we substitute $L=3\sim6$ LN layers with BN layers from the bottom in the Post-Norm Transformer encoder on IWSLT14. As $L$ increases, the variance TID of the last BN layer grows, and the BLEU scores of $\text{Transformer}_{BN}$ drops off. We plot the variance TID and BLEU gap between $\text{Transformer}_{BN}$ and $\text{Transformer}_{LN}$ in \cref{fig:tid_vs_gap} (left). We can see that the two quantities are highly correlated. \par
In \cref{fig:tid_vs_gap} (right), we plot the variance TID of the last BN layer and the validation loss gap between $\text{Transformer}_{BN}$ and $\text{Transformer}_{LN}$ on IWSLT14 through training. The validation loss gap is calculated by subtracting loss of $\text{Transformer}_{LN}$ from $\text{Transformer}_{BN}$. At the beginning of training, BN performs better than LN. When the TID begins to explode, BN's performance starts to degrade. \par
Based on the results in \cref{table:basic_result} and observations in \cref{fig:tid_vs_gap}, we argue that TID serves as an indicator of BN's performance in Transformers. Large TID hurts BN's performance, while BN with small TID performs better than LN due to its more efficient optimization (see experimental validation in \cref{ssec:analysis_rbn}).

\vspace{-0.06in}
\section{Suppressing High TID by RBN}
\vspace{-0.06in}
\label{sec:rbn}
In this section, we are devoted to reducing the TID of BN when it is large. If TID is suppressed, the performance of BN will be improved and may exceed LN due to the training efficiency of BN.
\vspace{-0.06in}
\subsection{Regularized Batch Normalization}
\vspace{-0.06in}
Assume there are $H$ layers of BN in a neural network. We denote the batch statistics and running statistics of each layer by $\mu_{B}^{i}$, $\sigma_{B}^{i}$, and $\mu^{i}$, $\sigma^{i}$, $i=1\ldots,H$. Assume the Cross-Entropy (CE) loss with respect to the neural network parameters $\theta$ is denoted by ${\cal L}(\theta)$. To avoid undesirable training inference discrepancy, we pose the optimization as a constrained problem: 
\begin{equation}
\begin{aligned}
\min\limits_{\theta}\quad & {\cal L}(\theta) \\
s.t. \quad & \E_{p_B}  d_{\mu}(\mu_{B}^{i}, \mu^{i})  \leq \epsilon_i, \ i=1,\ldots,H \\
& \E_{p_B} d_{\sigma}(\sigma_{B}^{i},\sigma^{i})  \leq \eta_i, \ i=1,\ldots,H
\end{aligned}
\end{equation}
where $d_{\mu}$ and $d_{\sigma}$ measure the inconsistency of mean and variance. 
This is equivalent to 
\begin{equation}
\min\limits_{\theta}\quad {\cal L}(\theta)+\sum\limits_{i=1}^{H}\lambda_i \E d_{\mu}(\mu_{B}^{i}, \mu^{i}) +\nu_i \E d_{\sigma}(\sigma_{B}^{i},\sigma^{i}) 
\end{equation}
To simplify the problem, we set $\lambda_i=\lambda$, $\nu_i = \nu$, for $\ i=1,\ldots,H$. \\
 When handling batch data, we apply gradient-based optimization to the following loss (${\cal L}_{B}(\theta)$ is the  batch CE loss):
\[
{\cal L}_{B}(\theta)+\sum\limits_{i=1}^{H}\lambda d_{\mu}(\mu_{B}^{i}, \mu^{i})+\nu d_{\sigma}(\sigma_{B}^{i},\sigma^{i}) 
\]
In particular, we choose $d_{\mu}\left(\mu_{B}, \mu \right) = \|\mu_{B}-\mu \|_2^2$ and $d_{\sigma}\left( \sigma_{B},\sigma\right) = \|\sigma_{B}-\sigma \|_2^2$. The sensitivity analysis of hyperparameter is given in \cref{ssec:analysis_rbn}. Since back propagating through the running statistics $\mu$ and $\sigma$ would trace back to the first batch of data which is impractical, we simply stop the gradient of $\mu$ and $\sigma$ in back propagation. 
\subsection{Experimental Result for RBN}
\begin{table}[t]
	\centering
	\caption{Results for the performance of Post-Norm (top) and Pre-Norm (bottom) Transformers with LN/BN/RBN. RBN consistently improves BN and could match or exceed LN on 17 out of 20 settings.}
	\resizebox{\columnwidth}{!}{
		\begin{tabular}{@{}ccccccccccc@{}}
			\toprule
			Task                  & \multicolumn{2}{c}{NMT (+)}       & \multicolumn{2}{c}{LM (-)}        & \multicolumn{2}{c}{NER (+)}       & \multicolumn{4}{c}{TextCls (+)}                                   \\ \midrule
			Datasets              & IWSLT14       & WMT16         & PTB           & WT103         & Resume        & CoNLL         & IMDB          & Sogou         & DBPedia       & Yelp          \\ \midrule
			Post-LN               & \textbf{35.5} & \textbf{27.3} & 53.2          & 20.9          & \textbf{94.8} & 91.3          & 84.1          & 94.6          & 97.5          & 93.3          \\
			Post-BN               & 34.0          & 25.0          & 45.9          & 17.2          & 94.5          & 90.9          & 84.0          & 94.3          & 97.5          & 93.3          \\
			Post-RBN              & \textbf{35.5} & 26.5          & \textbf{44.6} & \textbf{17.1} & \textbf{94.8} & \textbf{91.4} & \textbf{84.5} & \textbf{94.7} & \textbf{97.6} & \textbf{93.6} \\
			\midrule \midrule
			Pre-LN                & 35.5          & \textbf{27.3} & 54.5          & 24.6          & \textbf{94.0} & \textbf{91.0} & 84.1          & 94.5          & \textbf{97.5} & 93.3          \\
			Pre-BN                & 34.8          & 25.2          & 45.9          & 17.8          & 93.2          & 89.9          & 84.0          & 94.3          & \textbf{97.5} & 93.3          \\
			Pre-RBN               & \textbf{35.6} & 26.2          & \textbf{43.2} & \textbf{17.1} & \textbf{94.0} & 90.6          & \textbf{84.4} & \textbf{94.7} & \textbf{97.5} & \textbf{93.5} \\
			
			\bottomrule
		\end{tabular}
	}
	
	\label{table:rbn_result}
\end{table}

We choose $\lambda, \nu$ both from $\left\{0,0.01,0.1,1\right\}$ by validation loss. Results are shown in \cref{table:rbn_result}. The optimal hyperparameters are listed in \cref{sec:optimal_hyperpara}. 
\paragraph{Neural Machine Translation}
On IWSLT14 datasets, we see that RBN significantly improves BN and can exceed LN with 0.1 BLEU scores with Pre-Norm Transformer and match LN with Post-Norm Transformer. On WMT16 dataset, although RBN still falls behind LN, it could improve 1.5/1.0 BLEU scores over BN in Post-Norm/Pre-Norm setting. The reason is that even though RBN can suppress a large amount of TID, the remaining is still large since the original TID is huge. We speculate that the high data diversity in WMT16 contributes to the explosive TID of BN, which is hard to remove. We leave the verification as future work.
\vspace{-0.08in}
\paragraph{Language Modeling}
On Post-Norm Transformer, BN could boost the testing PPL of LN from 53.2 to 45.9 on PTB and from 20.9 to 17.2 on WikiText-103. Furthermore, substituting RBN for BN improves the testing PPL to 44.6 on PTB and 17.1 on WikiText-103. On Pre-Norm Transformer, BN elevates the testing PPL of LN from 54.5 to 45.9 on PTB and from 24.6 to 17.8 on WikiText-103. Moreover, replacing BN with RBN improves the testing PPL to 43.2 on PTB and 17.1 on WikiText-103.
Overall, RBN exceeds LN with 8.6/3.8 testing PPL with Post-Norm Transformer and 11.3/7.5 testing PPL with Pre-Norm Transformer on PTB/WikiText-103. 
\vspace{-0.08in}
\paragraph{Named Entity Recognition}
BN performs worse than LN on both Resume and CoNLL2003 datasets, especially for Pre-Norm Transformer. RBN improves BN in all settings, matches or exceeds LN in three out of four settings. By taking the better performance of Post-Norm and Pre-Norm, RBN matches the performance of LN on Resume and exceeds LN on CoNLL2003.
\vspace{-0.08in}
\paragraph{Text Classification}
We find that BN performs similar to/worse than LN on 4/4 settings. RBN improves the performance of BN consistently and can match/exceed LN on 1/7 settings. RBN improves BN with 0.3\% accuracy on average, which shows the benefit of our regularization. We do not intend to achieve the state-of-the-art performance but to verify the efficacy of RBN.
\vspace{-0.08in}
\paragraph{Comparison to BN's Variants}
We compare our RBN with Power Normalization (PN)~\cite{Shen2020Powernorm}, Batch
Renormalization (BRN)~\cite{2017_NIPS_Ioffe}, and Moving Averaing Batch Normaliazation
(MABN)~\cite{2020_ICLR_Yan} in \cref{table:compare_bn_variants}. These methods incorporate population statistics of BN in training, which is beneficial for alleviating training inference inconsistency of BN. PN and MABN are implemented by their official codes\footnote{https://github.com/sIncerass/powernorm. GPL-3.0 license. https://github.com/megvii-model/MABN. MIT license.}. BRN is implemented according to their paper~\cite{2017_NIPS_Ioffe}. The configurations of PN, BRN, and MABN are given in \cref{sec:bn_variants_config}. We highlight that PN incorporates layer scaling (LS)~\cite{Zhang2019RMSNorm}, which is important for stabilizing training, as shown in the supplementary materials and
official code of PN. We thus report the results for PN only and PN with layer scaling (PN+LS). We can see that RBN performs the best in most settings. PN and MABN is not
stable without layer scaling, especially for Post-Norm Transformers. 

\begin{table}[t]
	\centering
	\caption{Results for the performance of Post-Norm (top) and Pre-Norm (bottom) Transformers with PN/BRN/MABN/RBN.}
	\resizebox{\columnwidth}{!}{
		\begin{tabular}{@{}ccccccccccc@{}}
			\toprule
			Task                  & \multicolumn{2}{c}{NMT (+)}       & \multicolumn{2}{c}{LM (-)}        & \multicolumn{2}{c}{NER (+)}       & \multicolumn{4}{c}{TextCls (+)}                                   \\ \midrule
			Datasets              & IWSLT14       & WMT16         & PTB           & WT103         & Resume        & CoNLL         & IMDB          & Sogou         & DBPedia       & Yelp          \\ \midrule
			Post-PN-only &   0        &      0     &    254.6       &    inf       &     94.4     &      67.1     &   84.2       &     90.6      &    97.1       & 89.6 \\
			Post-PN+LS &   \textbf{35.6}        &    0       &   49.8        &    21.0       &      94.3    &     90.9      &     84.0     &   94.6        &   97.4        &  93.2 \\
			
			Post-BRN &    35.3     &    25.8     &   45.1      &    17.3     &         93.6      &  89.9    &     83.6     &   94.5       &    97.5      &  93.3  \\
			Post-MABN &   0      &    0     &    47.4     &   33.6      &    94.4     &  90.8    &     84.1     &     94.5     &    97.5      & 93.5  \\
			Post-RBN              & 35.5 & \textbf{26.5}          & \textbf{44.6} & \textbf{17.1} & \textbf{94.8} & \textbf{91.4} & \textbf{84.5} & \textbf{94.7} & \textbf{97.6} & \textbf{93.6} \\
			\midrule \midrule
			Pre-PN-only &     34.5      &    26.0       &      48.6     &   inf        &     5.0     &    11.1       &   84.2       &     94.4      &    97.4       &  93.3\\
			Pre-PN+LS  &    \textbf{35.6}       &   \textbf{27.2}        &     59.8      &    20.9       &    93.3      &    54.1       &     83.3     &     94.4      &     97.3      &  93.4\\

			Pre-BRN &    35.2     &    25.3     &   45.7      &    17.5     & 94.1        &   \textbf{91.1}   &    84.3      &  94.5      &    97.4     & 93.4 \\
			Pre-MABN &    35.0     &    26.8     &  48.7       &    inf     &   \textbf{94.8}      & 90.9     &     \textbf{84.4}     &   94.6       &     97.5     & 93.3 \\
			Pre-RBN               & \textbf{35.6} & 26.2          & \textbf{43.2} & \textbf{17.1} & 94.0 & 90.6          & \textbf{84.4} & \textbf{94.7} & \textbf{97.5} & \textbf{93.5} \\
			
			\bottomrule
		\end{tabular}
	}
	
	\label{table:compare_bn_variants}
\end{table}

\subsection{Analysis}
\label{ssec:analysis_rbn}
\begin{figure}[t]
	\captionsetup[subfigure]{justification=centering}
	\centering
	\begin{subfigure}[b]{0.24\columnwidth}
		\centering
		\centerline{\includegraphics[width=\textwidth]{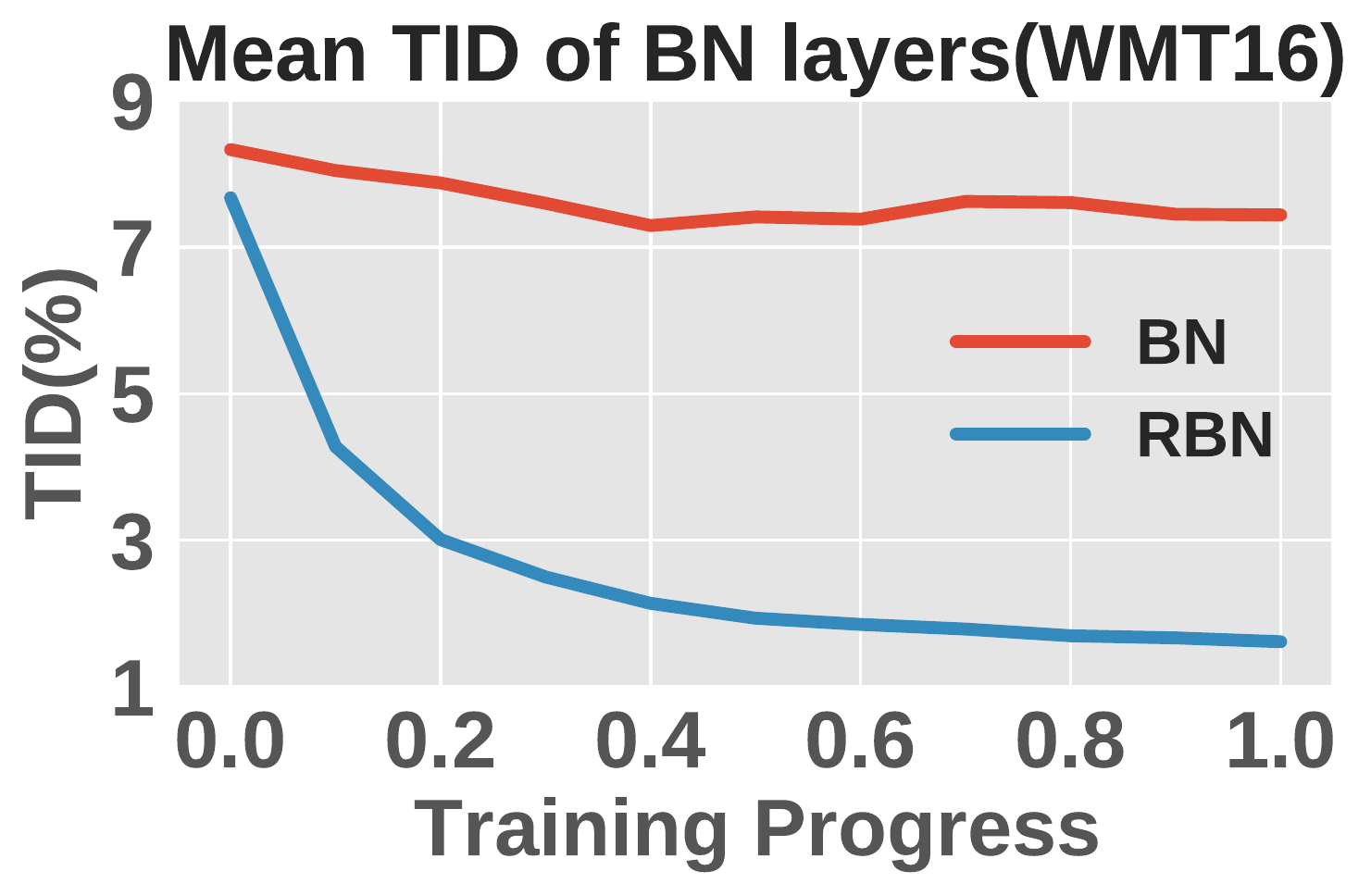}}
	\end{subfigure}
	\hfill
	\begin{subfigure}[b]{0.24\columnwidth}
		\centering
		\includegraphics[width=\textwidth]{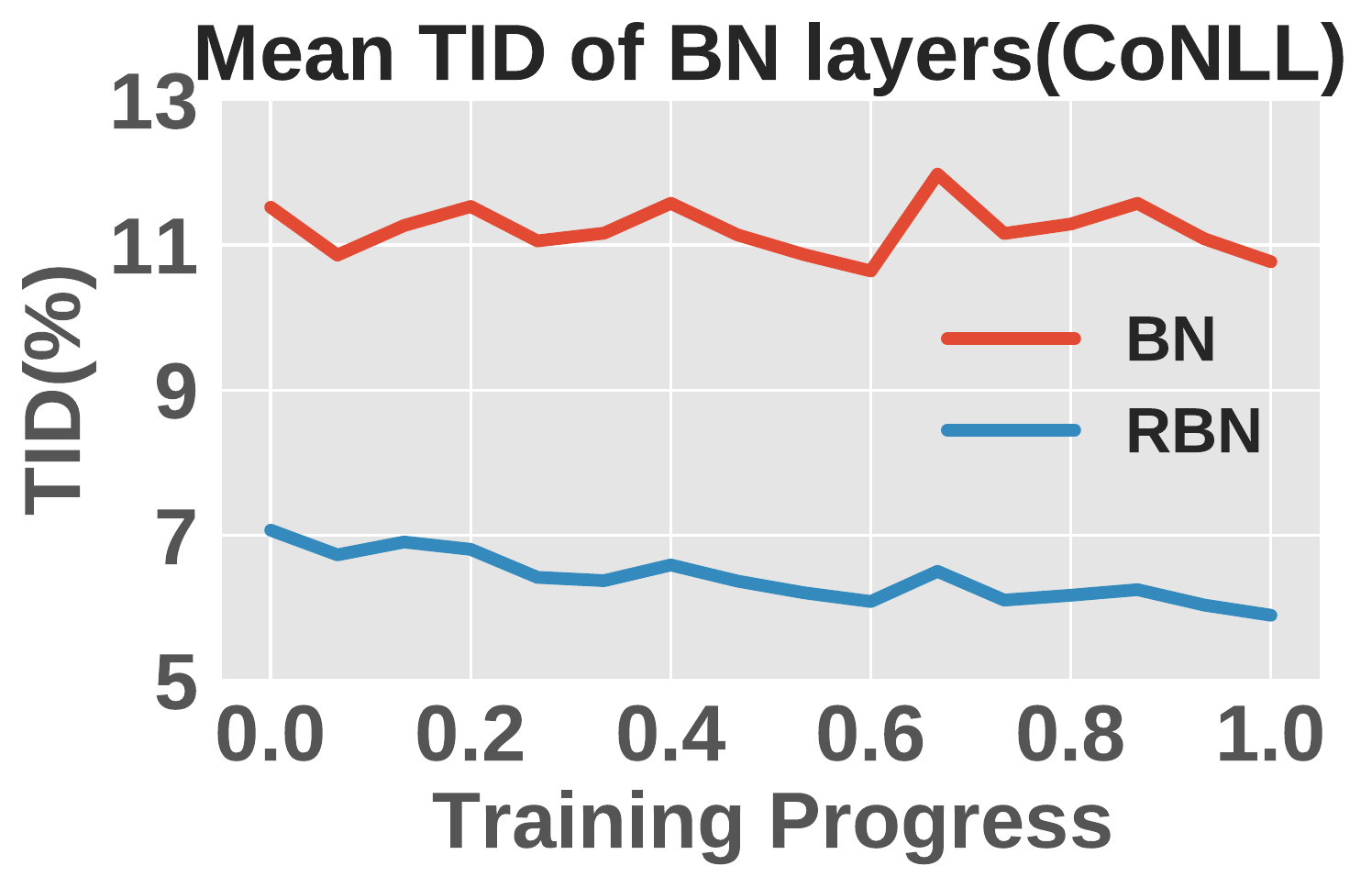}
	\end{subfigure}
	\hfill
	\begin{subfigure}[b]{0.24\columnwidth}
		\centering
		\includegraphics[width=\textwidth]{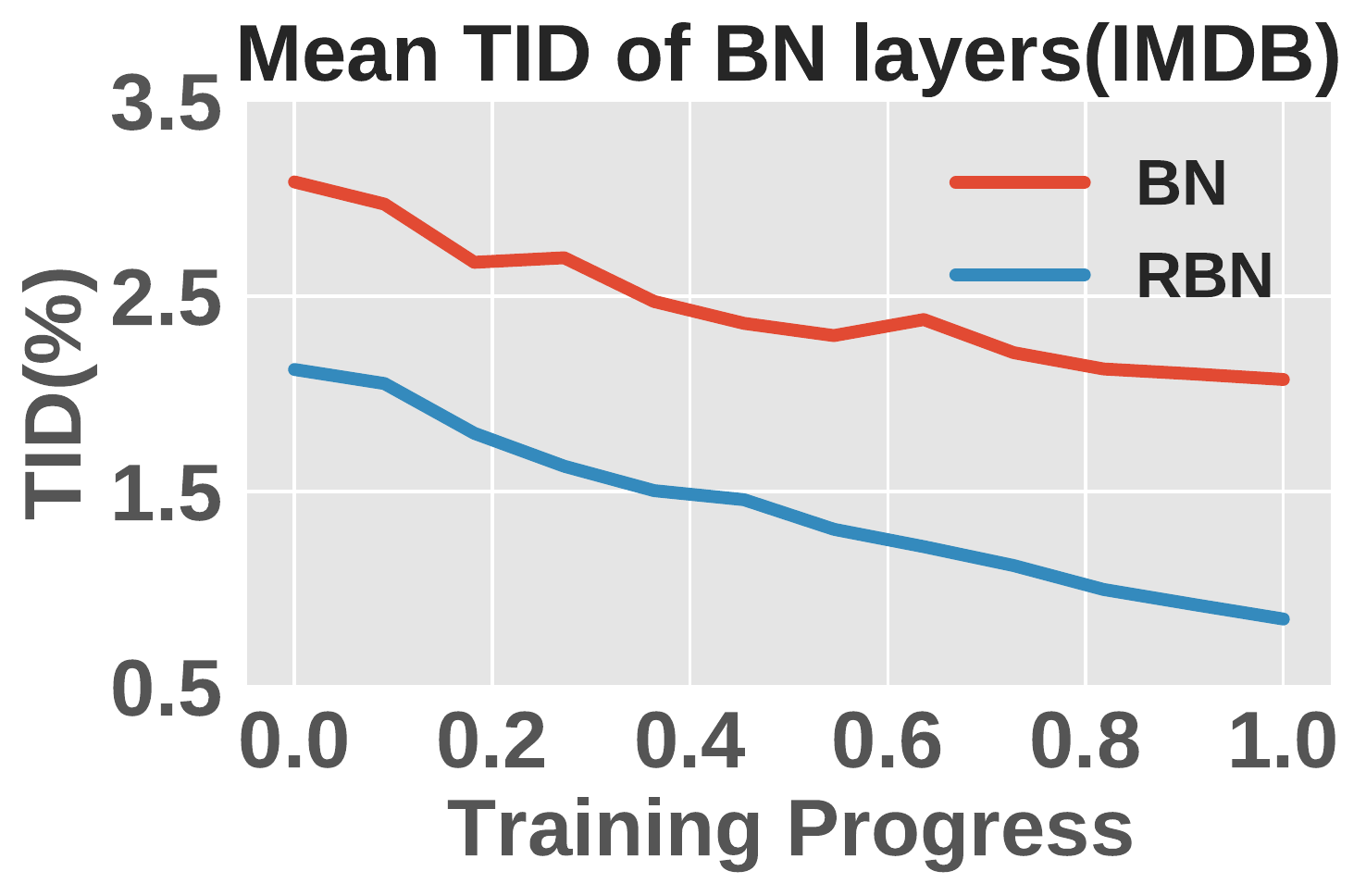}
	\end{subfigure}
	\hfill
	\begin{subfigure}[b]{0.24\columnwidth}
		\centering
		\includegraphics[width=\textwidth]{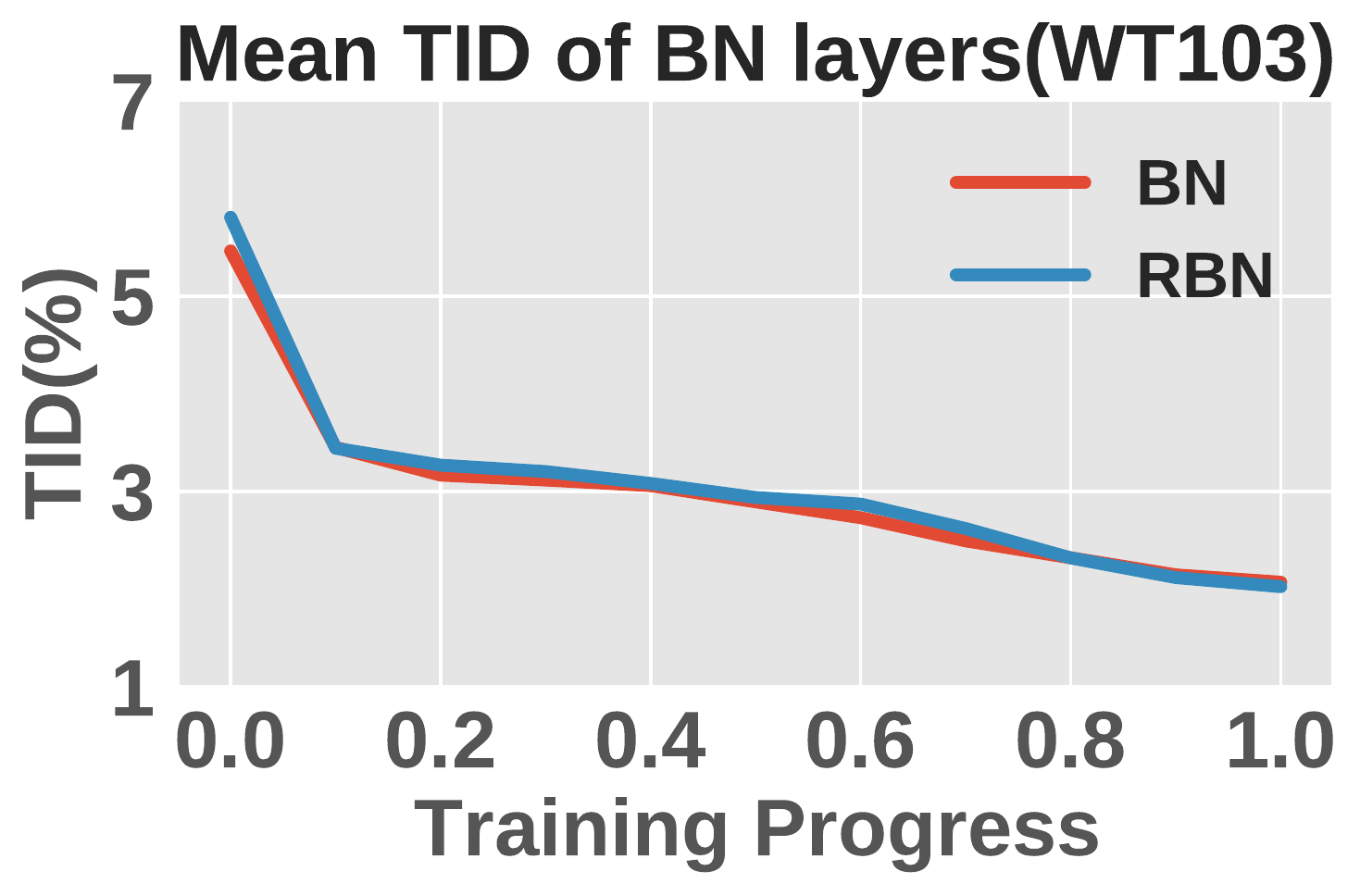}
	\end{subfigure}
	\begin{subfigure}[b]{0.24\columnwidth}
		\centering
		\centerline{\includegraphics[width=\textwidth]{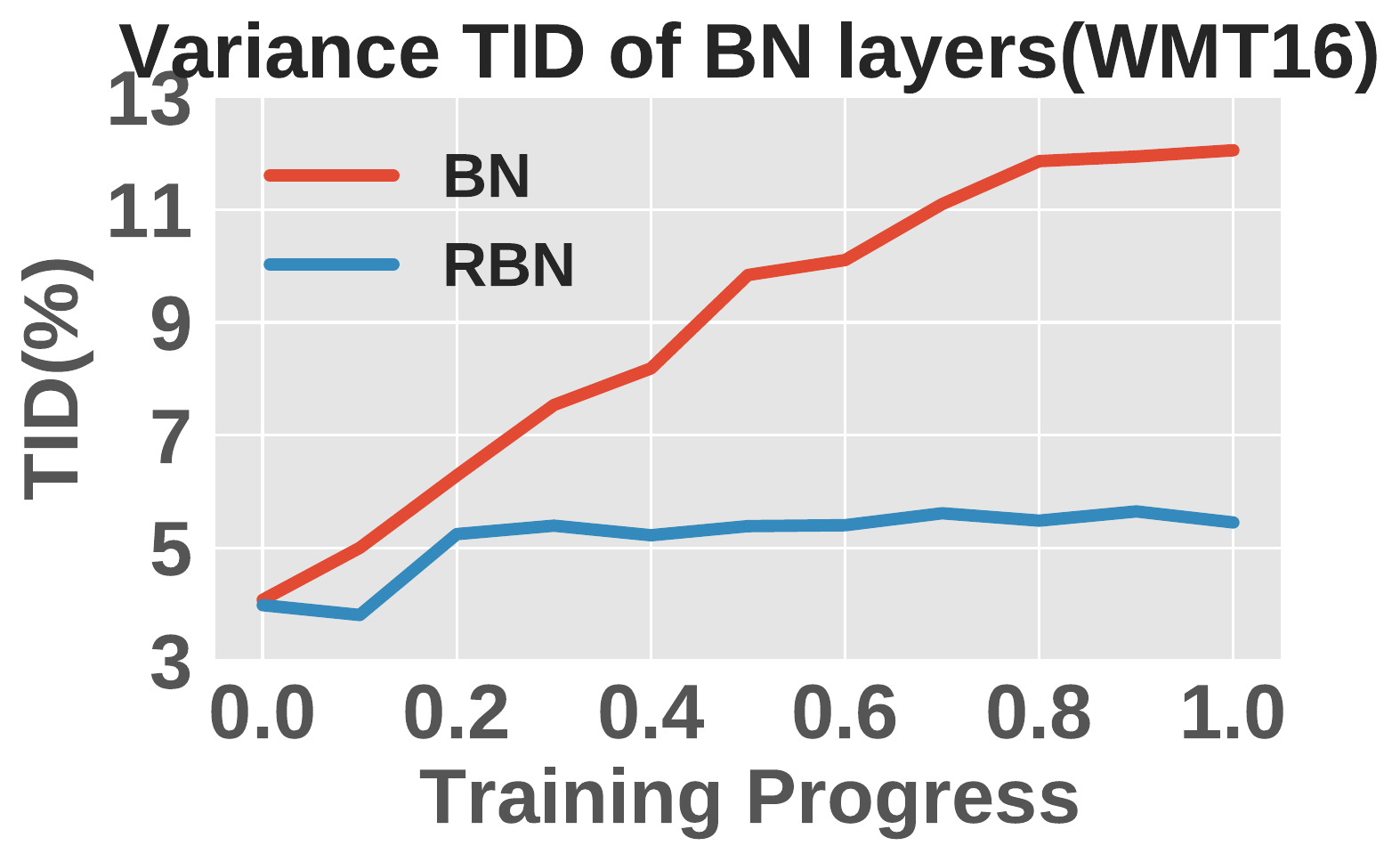}}
	\end{subfigure}
	\hfill
	\begin{subfigure}[b]{0.24\columnwidth}
		\centering
		\includegraphics[width=\textwidth]{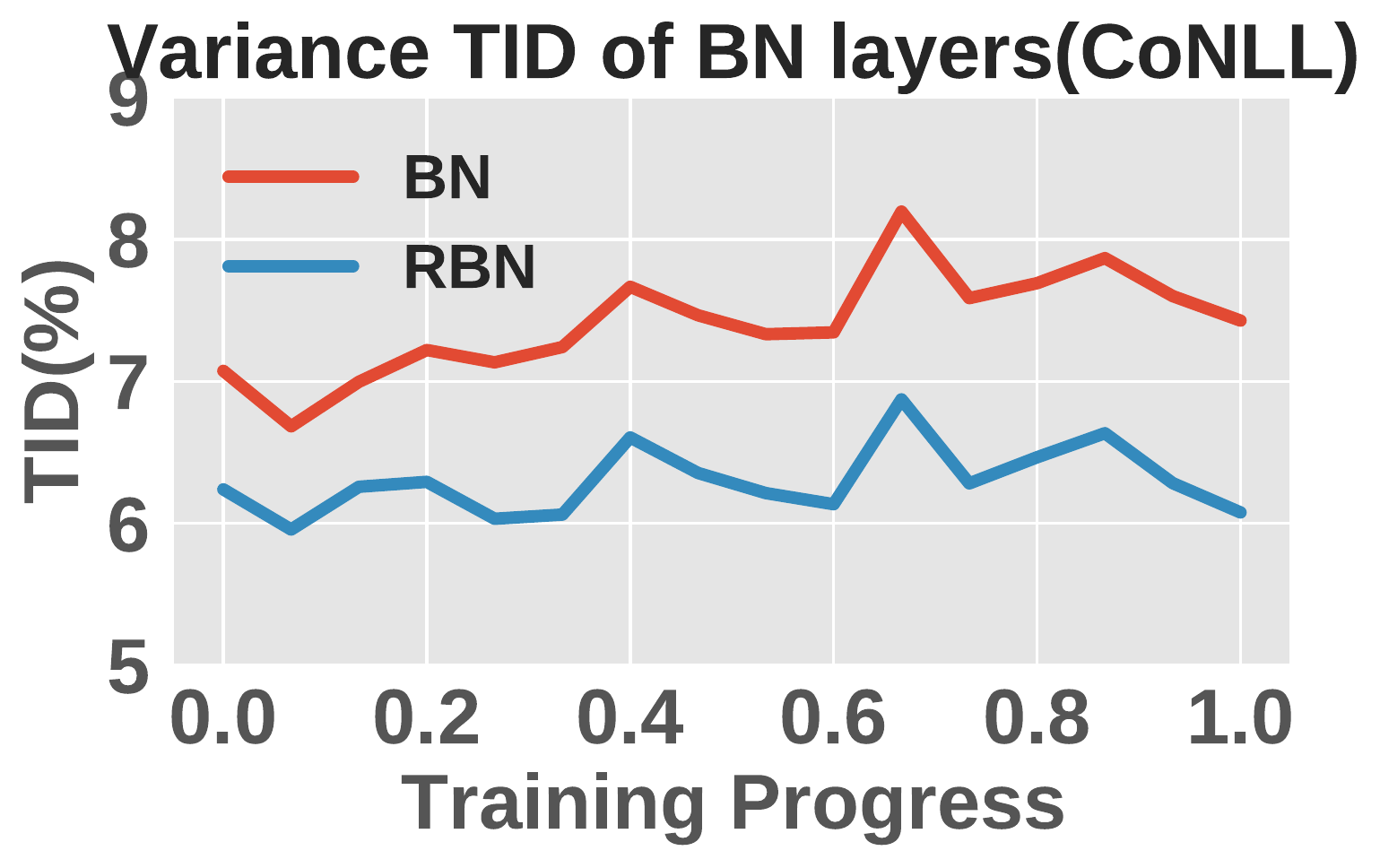}
	\end{subfigure}
	\hfill
	\begin{subfigure}[b]{0.24\columnwidth}
		\centering
		\includegraphics[width=\textwidth]{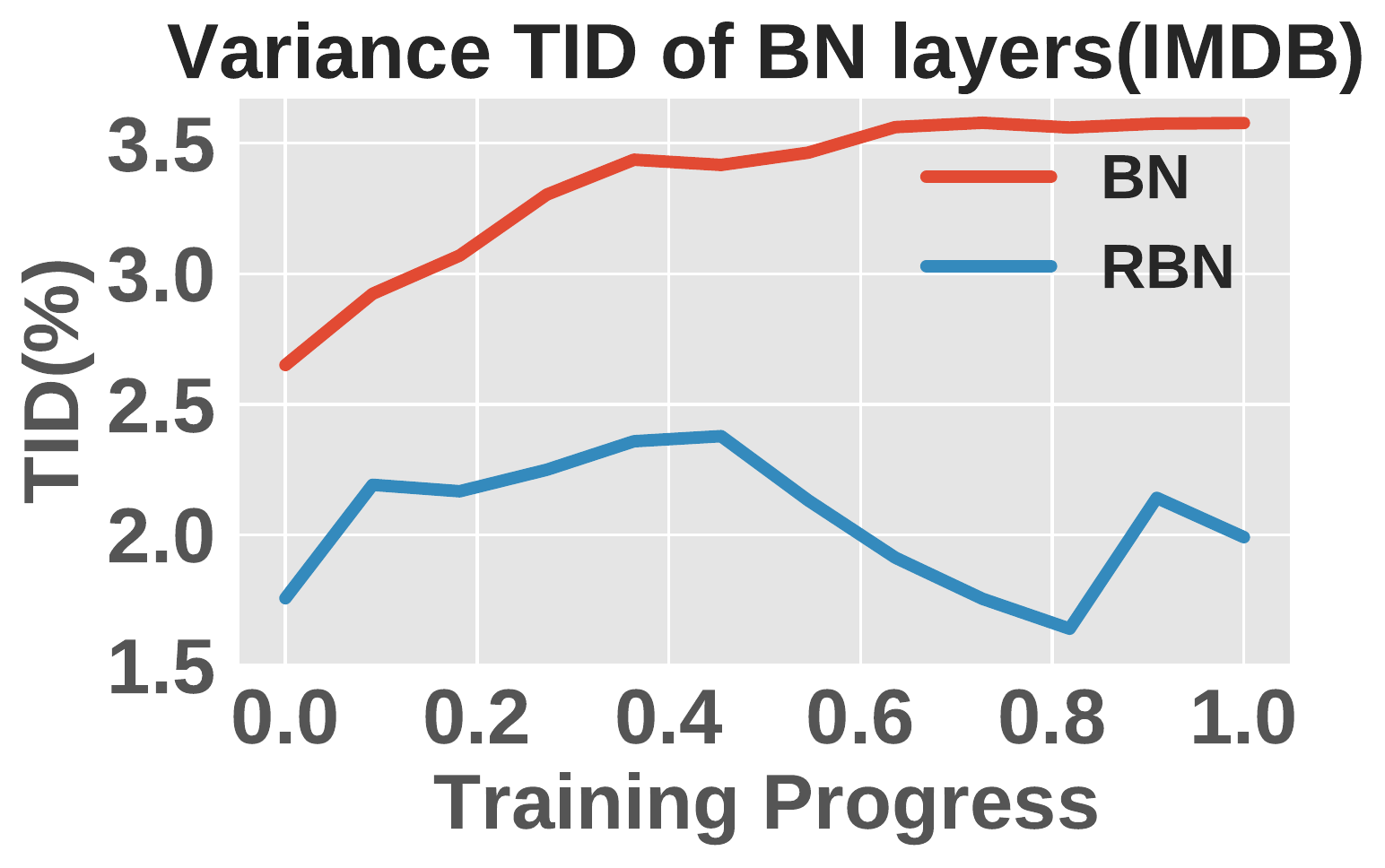}
	\end{subfigure}
	\hfill
	\begin{subfigure}[b]{0.24\columnwidth}
		\centering
		\includegraphics[width=\textwidth]{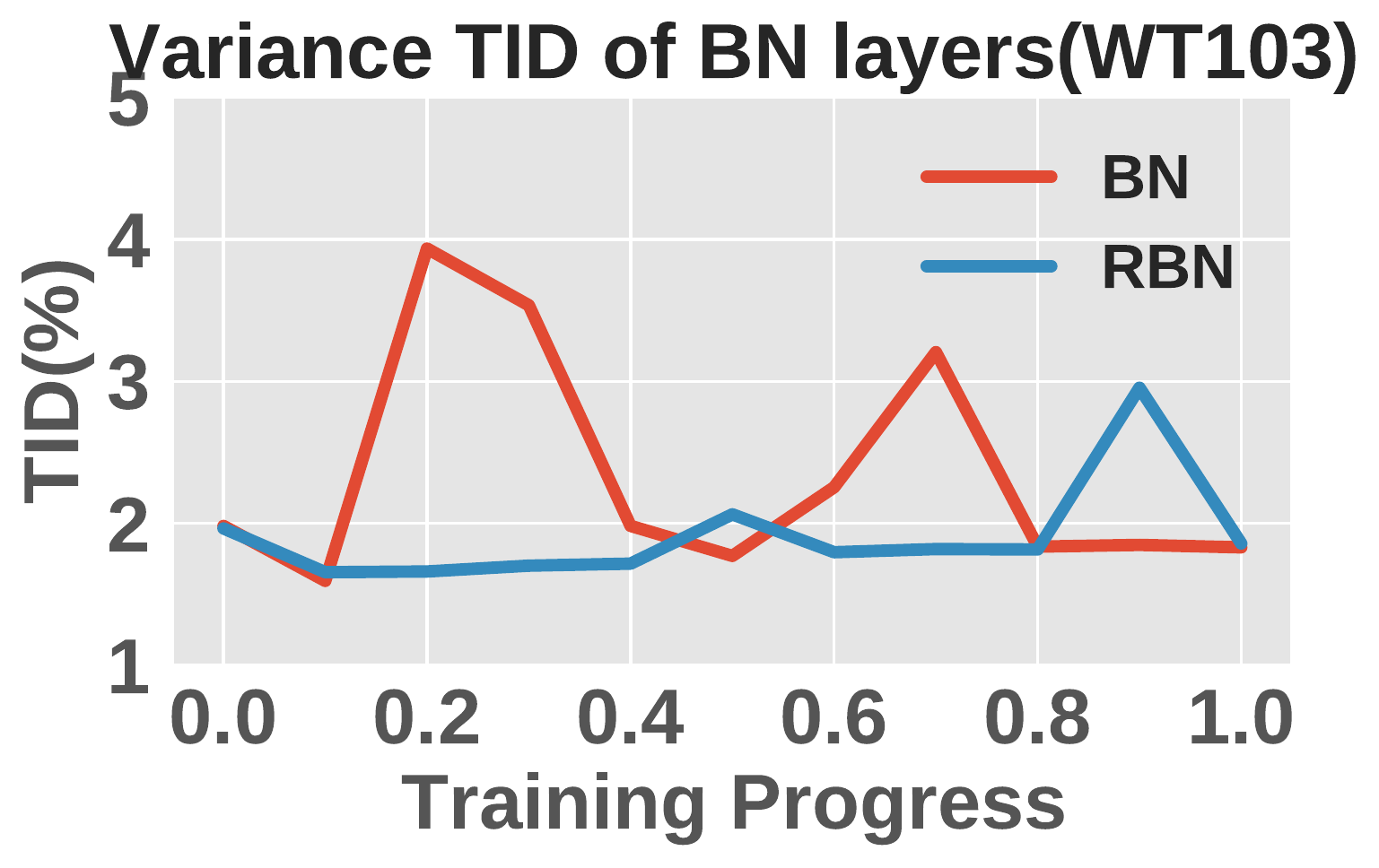}
	\end{subfigure}
	\caption{Average Mean and Variance TID on WMT16/CoNLL/IMDB/WT103 for Pre-Norm Transformer with BN and RBN. RBN reduces the Mean and Variance TID of BN at the end of the training and leads to better performance.}
	\label{fig:mean_var_tid}
\end{figure}
\paragraph{Training Inference Inconsistency}
\begin{table}[t]
	\centering
	\vspace{-0.2in}
	\caption{TID of the last BN/RBN layer in Post-Norm and Pre-Norm Transformers on various NLP tasks. RBN reduces the TID of BN effectively.}
	\resizebox{\columnwidth}{!}{
		\begin{tabular}{@{}ccccccccccc@{}}
			\toprule
			Task                  & \multicolumn{2}{c}{NMT}       & \multicolumn{2}{c}{LM}        & \multicolumn{2}{c}{NER}       & \multicolumn{4}{c}{TextCls}                                   \\ \midrule
			Datasets              & IWSLT14       & WMT16         & PTB           & WT103         & Resume        & CoNLL         & IMDB          & Sogou         & DBPedia       & Yelp          \\ \midrule
			\multicolumn{11}{c}{Post-Norm Transformer} \\
			\midrule
			Mean TID of $\text{BN}_{last}$  & 1.5\%         & 4.2\%         & 0.9\%         & 1.8\%         & 1.7\%         & 4.2\%         & 1.8\%         & 1.8\%         & 2.2\%         & 3.1\%         \\
			Mean TID of $\text{RBN}_{last}$ & 0.8\%         & 2.3\%         & 0.9\%         & 1.8\%         & 1.4\%         & 1.9\%         & 0.2\%         & 0.2\%         & 0.3\%         & 0.2\%         \\
			Var TID of $\text{BN}_{last}$   & 10.6\%         & 17.9\%        & 1.1\%         & 2.0\%         & 3.7\%         & 9.5\%         & 3.9\%         & 4.3\%         & 3.5\%         & 4.0\%         \\
			
			Var TID of $\text{RBN}_{last}$  & 6.7\%         & 7.7\%         & 1.1\%         & 1.7\%         & 3.0\%         & 5.0\%         & 1.2\%         & 0.2\%         & 0.3\%         & 0.1\%         \\ \midrule \midrule
			\multicolumn{11}{c}{Pre-Norm Transformer} \\ \midrule
			Mean TID of $\text{BN}_{last}$  & 3.4\%         & 7.9\%         & 1.6\%         & 2.4\%         & 9.6\%         & 10.0\%        & 2.9\%         & 7.5\%         & 3.9\%         & 12.1\%        \\
			Mean TID of $\text{RBN}_{last}$ & 3.2\%         & 1.3\%         & 1.6\%         & 2.4\%         & 4.5\%         & 4.0\%         & 0.7\%         & 1.0\%         & 1.1\%         & 1.0\%         \\
			Var TID of $\text{BN}_{last}$   & 4.6\%         & 30.1\%        & 1.7\%         & 2.5\%         & 6.5\%         & 6.4\%         & 6.2\%         & 7.1\%         & 3.3\%         & 8.6\%         \\
			Var TID of $\text{RBN}_{last}$  & 1.5\%         & 12.1\%        & 1.7\%         & 2.4\%         & 6.3\%         & 5.6\%         & 4.7\%         & 0.4\%         & 0.5\%         & 0.5\%         \\ \bottomrule
		\end{tabular}
	}
	
	\label{table:rbn_tid_result}
\end{table}

We compute the TID of the last BN layer ($\text{BN}_{last}$) in \cref{table:rbn_tid_result} and plot the average TID of BN and RBN on WMT16, WT103, CoNLL2003, and IMDB datasets for Pre-Norm Transformers through training in \cref{fig:mean_var_tid}. Figures of TID for other datasets and Post-Norm Transformer can be found in \cref{sec:avg_tid_compare}. We can see that RBN reduces BN's mean and variance TID at the end of training. On neural machine translation and named entity recognition tasks, the original TID is large. RBN significantly decreases the TID of BN and improves BN's performance by a clear margin. For language modeling and text classification tasks, RBN also reduces the moderate TID of BN and gets better PPL or accuracy. 
\begin{wrapfigure}{r}{0.42\textwidth}
	%
	\captionsetup[subfigure]{justification=centering}
	\begin{center}
		\includegraphics[width=0.4\textwidth]{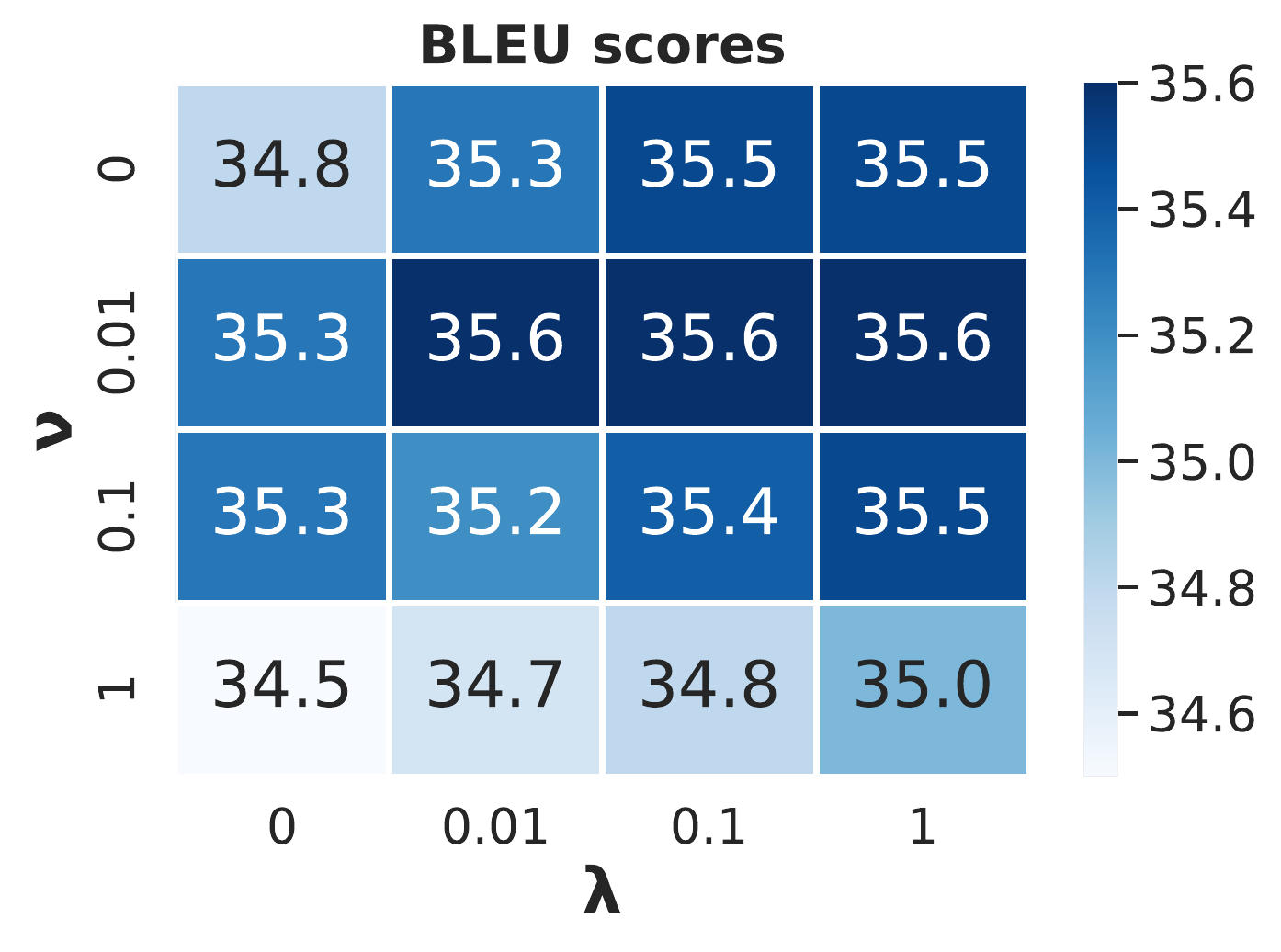}
	\end{center}
	\caption{The BLEU scores on IWSLT14 with different mean ($\lambda$) and variance ($\nu$) discrepancy penalty of RBN.}
	\label{fig:penaltybleu}
\end{wrapfigure}
\vspace{-0.25in}
\paragraph{Sensitivity to Hyperparameters}
We test different penalty coefficients for RBN on neural machine translation with Pre-Norm Transformer. The results are shown in \cref{fig:penaltybleu}. Penalizing the mean and variance discrepancy can both improve the performance of BN. Combining them with moderate coefficients achieves the best performance.
\vspace{-0.12in}
\paragraph{Training Dynamics}
\label{sssec:training_dynamic}
\begin{figure}[t]
	\captionsetup[subfigure]{justification=centering}
	\centering

	\begin{subfigure}[b]{0.32\columnwidth}
		\centering
		\centerline{\includegraphics[width=\textwidth]{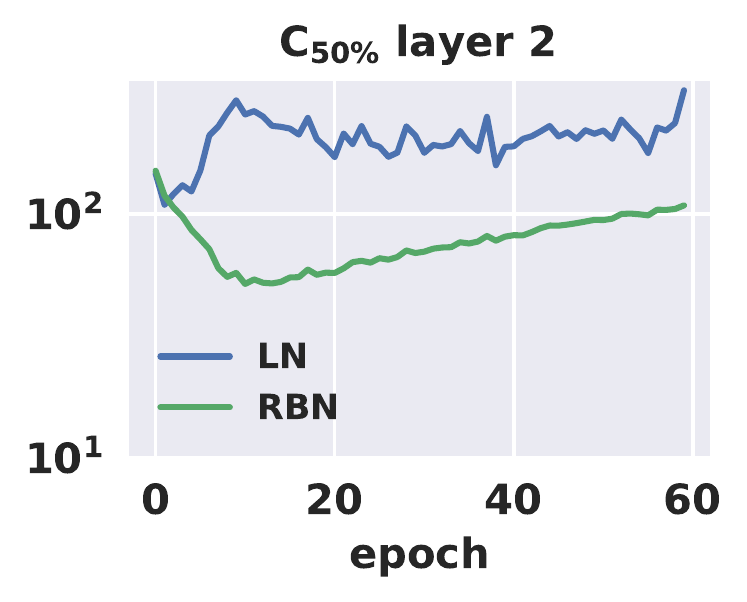}}
	\end{subfigure}
	\hfill
	\begin{subfigure}[b]{0.32\columnwidth}
		\centering
		\includegraphics[width=\textwidth]{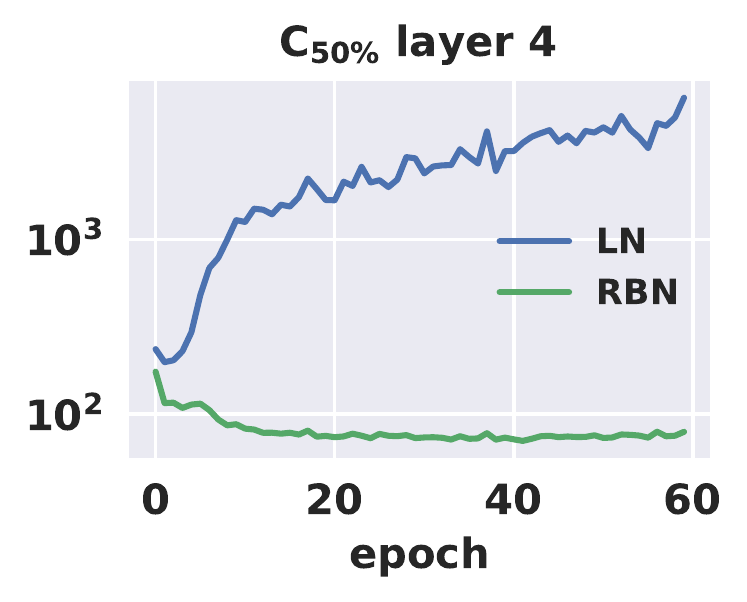}
	\end{subfigure}
	\hfill
	\begin{subfigure}[b]{0.32\columnwidth}
		\centering
		\includegraphics[width=\textwidth]{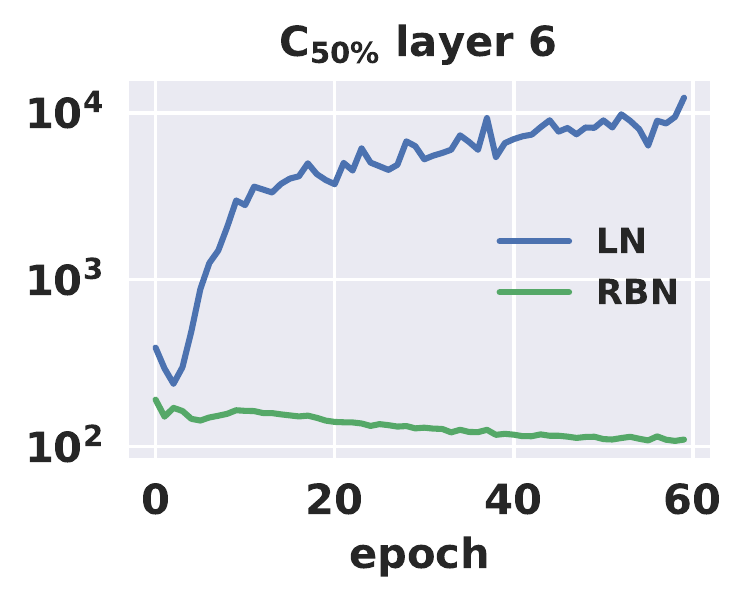}
	\end{subfigure}
	
	\begin{subfigure}[b]{0.32\columnwidth}
		\centering
		\centerline{\includegraphics[width=\textwidth]{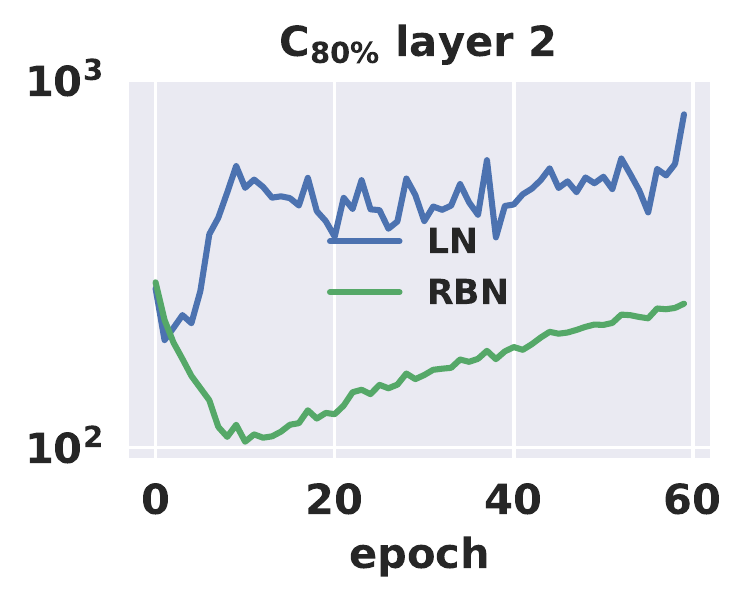}}
	\end{subfigure}
	\hfill
	\begin{subfigure}[b]{0.32\columnwidth}
		\centering
		\includegraphics[width=\textwidth]{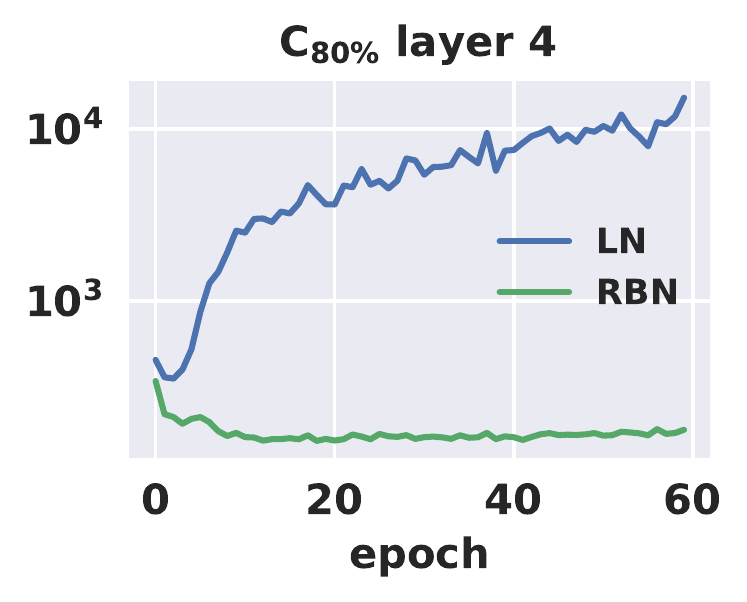}
	\end{subfigure}
	\hfill
	\begin{subfigure}[b]{0.32\columnwidth}
		\centering
		\includegraphics[width=\textwidth]{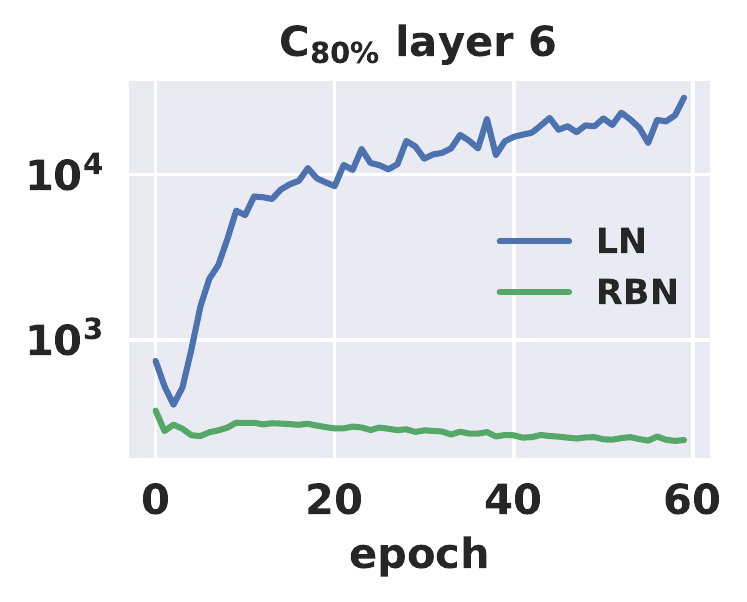}
	\end{subfigure}
	
	\caption{$C_{50\%}$ (top), and $C_{80\%}$ (bottom) of input features of Transformer encoder layer 2/4/6. RBN improves the $C_{50\%}$ and $C_{80\%}$ of LN, especially for deep layers (2 orders of magnitude at layer 6).}
	\label{fig:Cp_condition}
\end{figure}

\begin{figure}[t]
	\captionsetup[subfigure]{justification=centering}
	\centering
	\begin{subfigure}[b]{0.32\columnwidth}
		\centering
		\centerline{\includegraphics[width=\textwidth]{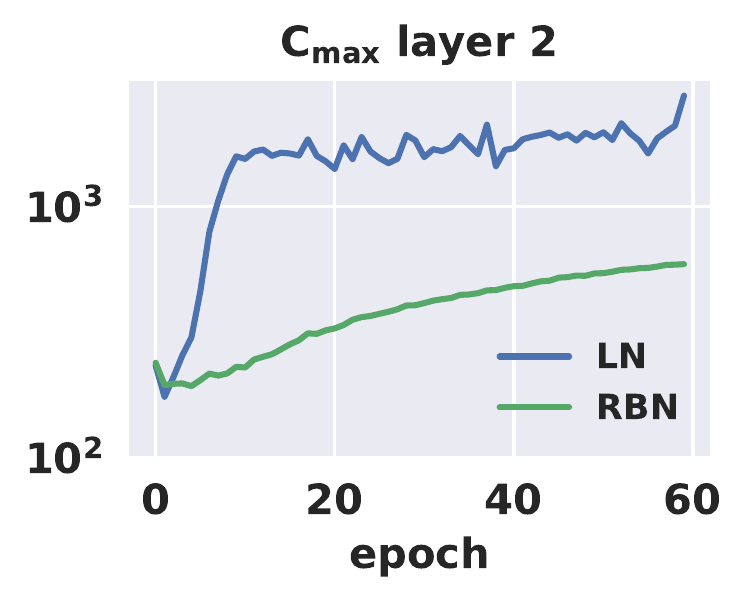}}
	\end{subfigure}
	\hfill
	\begin{subfigure}[b]{0.32\columnwidth}
		\centering
		\includegraphics[width=\textwidth]{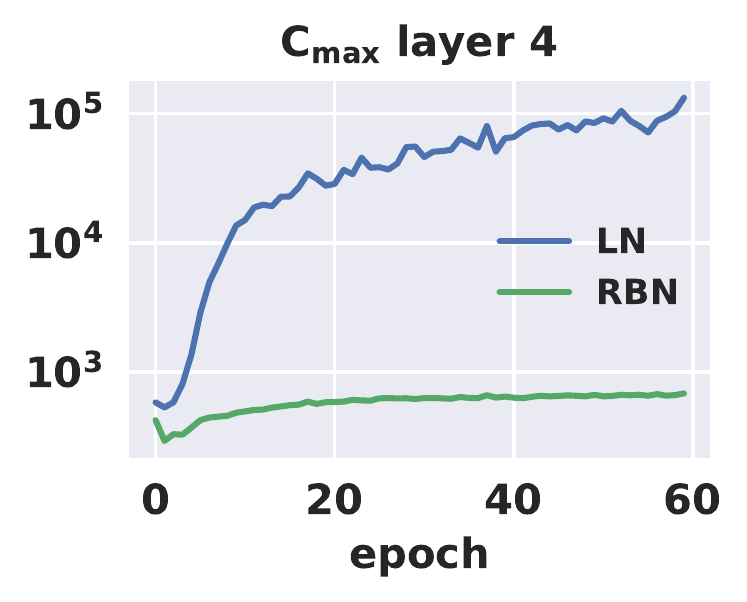}
	\end{subfigure}
	\hfill
	\begin{subfigure}[b]{0.32\columnwidth}
		\centering
		\includegraphics[width=\textwidth]{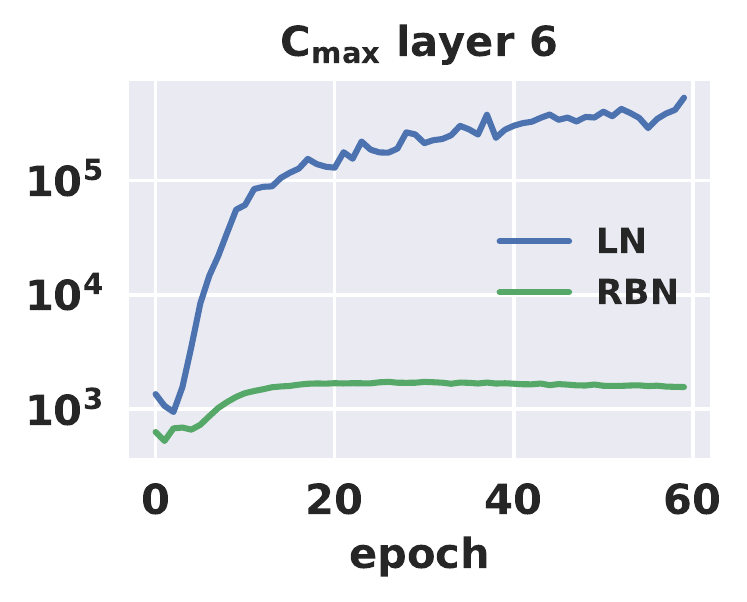}
	\end{subfigure}
	\caption{$\text{C}_{max}$ of input features of Transformer encoder layer 2/4/6 through training.}
	\label{fig:cmax}
\end{figure}
To show the optimization advantages of RBN over LN, we explore the layer-wise training dynamics of LN and RBN in Pre-Norm Transformer on IWSLT14. We refer the reader to \citet{Huang2020BNDynamic} for detailed analysis about the correlation between optimization of neural network and layer-wise training dynamics. We empirically observe that replacing LN with RBN significantly improves the layer-wise conditioning~\cite{Huang2020BNDynamic} of Transformer. We denote the intermediate embedding in Transformer by $\Tilde{\X{}} \in \R^{B\times T\times d}$, each $\Tilde{\X{}}_{i,j,:} \in \R^d$ is a word vector. We reshape $\Tilde{\X{}}$ to a sequence of word vectors to $\X{}=\left[\x{1},\x{2},\ldots, \x{BT} \right] \in \R^{BT\times d}$. We assume $BT>d$ which is satisfied in our experiments. We define the general condition number with respect to the percentage as $C_p(\X{})=\frac{\sigma_1}{\sigma_{\lceil pd \rceil}}$, $0<p \leq 1$. $\lceil a \rceil$ is the smallest integer that is larger than or equal to $a$. Lower $C_p(\X{})$ is usually associated with faster convergence of training. We plot $C_{50\%}$, and $C_{80\%}$ of input features of transformer encoder layer 2/4/6  in \cref{fig:Cp_condition}. We can see that RBN significantly reduces the $C_{50\%}$ and $C_{80\%}$ of LN, usually with orders of magnitude. We also plot the layer-wise $C_{max}(\X{})=\lambda_{max}((\X{}^T\X{})^{\frac{1}{2}})$ in \cref{fig:cmax}. Smaller $C_{max}$ usually permits higher learning rates which leads to faster training and better generalization~\cite{Hardt2016TrainFG}. RBN has much smaller $C_{max}$ than LN. 

\section{Conclusion and Limitation}
\label{sec:conclusion}
In this paper, we defined Training Inference Discrepancy (TID) and showed that TID is a good indicator of BN's performance for Transformers, supported by comprehensive experiments. 
We observed BN performs much better than LN when TID is negligible and proposed Regularized BN (RBN) to alleviate TID when TID is large.
Our RBN has theoretical advantages in optimization and works empirically better by controlling the TID of BN when compared with LN. We hope our work will facilitate a better understanding and application of BN in NLP. 
\vspace{-0.15 in}
\paragraph{Limitation.} Our analyses on TID are almost empirical studies without theoretical guarantee. It is better to further model the geometric distribution of word embedding, evolving along with the training dynamics and information propagation,  with theoretical derivation under mild assumptions. 
Besides, our proposed RBN cannot entirely suppress huge TID in training large-scale datasets with high diversity, leading to degraded performance. One possible direction is to combine RBN and LN for both better optimization properties and small TID, as explored in~\cite{2022_CVPR_Huang, Yao2021LeveragingBN} for CV tasks.
\paragraph{Acknowledgement} 
Ji Wu was sponsored by
National Key Research \& Development Program of China (2021ZD0113402), Tsinghua University Spring Breeze Fund (2021Z99CFZ010), National Key Research \& Development Program of China (Grant Number: 2021YFC2500803), and Tsinghua-Toyota Joint Research Institute Inter-disciplinary Program. Lei Huang was supported by National Natural Science Foundation of China (Grant No. 62106012) and the Fundamental Research Funds for the Central Universities. 
%


\bibliography{main}

\begin{thebibliography}{52}
\providecommand{\natexlab}[1]{#1}
\providecommand{\url}[1]{\texttt{#1}}
\expandafter\ifx\csname urlstyle\endcsname\relax
  \providecommand{\doi}[1]{doi: #1}\else
  \providecommand{\doi}{doi: \begingroup \urlstyle{rm}\Url}\fi

\bibitem[Ba et~al.(2016)Ba, Kiros, and Hinton]{Ba2016LN}
Jimmy~Lei Ba, Jamie~Ryan Kiros, and Geoffrey~E. Hinton.
\newblock Layer normalization, 2016.

\bibitem[Benz et~al.(2021)Benz, Zhang, Karjauv, and Kweon]{2021_WACV_Benz}
Philipp Benz, Chaoning Zhang, Adil Karjauv, and In~So Kweon.
\newblock Revisiting batch normalization for improving corruption robustness.
\newblock In \emph{WACV}, 2021.

\bibitem[Bhardwaj et~al.(2021)Bhardwaj, Majumder, Poria, and
  Hovy]{Bhardwaj2021TextClsTrans}
Rishabh Bhardwaj, Navonil Majumder, Soujanya Poria, and Eduard Hovy.
\newblock More identifiable yet equally performant transformers for text
  classification.
\newblock In \emph{Proceedings of the 59th Annual Meeting of the Association
  for Computational Linguistics and the 11th International Joint Conference on
  Natural Language Processing (Volume 1: Long Papers)}, pages 1172--1182.
  Association for Computational Linguistics, 2021.

\bibitem[Bjorck et~al.(2018)Bjorck, Gomes, Selman, and
  Weinberger]{Bjorck2018BNLargeLr}
Nils Bjorck, Carla~P Gomes, Bart Selman, and Kilian~Q Weinberger.
\newblock Understanding batch normalization.
\newblock In S.~Bengio, H.~Wallach, H.~Larochelle, K.~Grauman, N.~Cesa-Bianchi,
  and R.~Garnett, editors, \emph{Advances in Neural Information Processing
  Systems}, volume~31. Curran Associates, Inc., 2018.

\bibitem[Brown et~al.(2020)Brown, Mann, Ryder, Subbiah, Kaplan, Dhariwal,
  Neelakantan, Shyam, Sastry, Askell, Agarwal, Herbert-Voss, Krueger, Henighan,
  Child, Ramesh, Ziegler, Wu, Winter, Hesse, Chen, Sigler, Litwin, Gray, Chess,
  Clark, Berner, McCandlish, Radford, Sutskever, and Amodei]{Brown2020GPT3}
Tom Brown, Benjamin Mann, Nick Ryder, Melanie Subbiah, Jared~D Kaplan, Prafulla
  Dhariwal, Arvind Neelakantan, Pranav Shyam, Girish Sastry, Amanda Askell,
  Sandhini Agarwal, Ariel Herbert-Voss, Gretchen Krueger, Tom Henighan, Rewon
  Child, Aditya Ramesh, Daniel Ziegler, Jeffrey Wu, Clemens Winter, Chris
  Hesse, Mark Chen, Eric Sigler, Mateusz Litwin, Scott Gray, Benjamin Chess,
  Jack Clark, Christopher Berner, Sam McCandlish, Alec Radford, Ilya Sutskever,
  and Dario Amodei.
\newblock Language models are few-shot learners.
\newblock In \emph{Advances in Neural Information Processing Systems},
  volume~33, pages 1877--1901. Curran Associates, Inc., 2020.

\bibitem[Chiley et~al.(2019)Chiley, Sharapov, Kosson, Koster, Reece, Samaniego
  de~la Fuente, Subbiah, and James]{2019_NeurIPS_Chiley}
Vitaliy Chiley, Ilya Sharapov, Atli Kosson, Urs Koster, Ryan Reece, Sofia
  Samaniego de~la Fuente, Vishal Subbiah, and Michael James.
\newblock Online normalization for training neural networks.
\newblock In \emph{NeurIPS}, 2019.

\bibitem[Dai et~al.(2019)Dai, Yang, Yang, Carbonell, Le, and
  Salakhutdinov]{Dai2019TransXL}
Zihang Dai, Zhilin Yang, Yiming Yang, Jaime Carbonell, Quoc Le, and Ruslan
  Salakhutdinov.
\newblock Transformer-{XL}: Attentive language models beyond a fixed-length
  context.
\newblock In \emph{Proceedings of the 57th Annual Meeting of the Association
  for Computational Linguistics}, pages 2978--2988. Association for
  Computational Linguistics, 2019.

\bibitem[Daneshmand et~al.(2020)Daneshmand, Kohler, Bach, Hofmann, and
  Lucchi]{Dane2020BNRank1}
Hadi Daneshmand, Jonas~Moritz Kohler, Francis Bach, Thomas Hofmann, and
  Aurélien Lucchi.
\newblock Batch normalization provably avoids ranks collapse for randomly
  initialised deep networks.
\newblock In \emph{NeurIPS}, 2020.

\bibitem[Daneshmand et~al.(2021)Daneshmand, Joudaki, and Bach]{Dane2021BNRank2}
Hadi Daneshmand, Amir Joudaki, and Francis Bach.
\newblock Batch normalization orthogonalizes representations in deep random
  networks.
\newblock In \emph{Advances in Neural Information Processing Systems}, 2021.

\bibitem[Devlin et~al.(2019)Devlin, Chang, Lee, and Toutanova]{Devlin2019Bert}
Jacob Devlin, Ming-Wei Chang, Kenton Lee, and Kristina Toutanova.
\newblock {BERT}: Pre-training of deep bidirectional transformers for language
  understanding.
\newblock In \emph{Proceedings of the 2019 Conference of the North {A}merican
  Chapter of the Association for Computational Linguistics: Human Language
  Technologies, Volume 1 (Long and Short Papers)}, pages 4171--4186.
  Association for Computational Linguistics, 2019.

\bibitem[Hardt et~al.(2016)Hardt, Recht, and Singer]{Hardt2016TrainFG}
Moritz Hardt, Benjamin Recht, and Yoram Singer.
\newblock Train faster, generalize better: Stability of stochastic gradient
  descent.
\newblock \emph{ArXiv}, abs/1509.01240, 2016.

\bibitem[Huang et~al.(2019)Huang, Zhou, Zhu, Liu, and Shao]{Huang2019IterNorm}
Lei Huang, Yi~Zhou, Fan Zhu, Li~Liu, and Ling Shao.
\newblock Iterative normalization: Beyond standardization towards efficient
  whitening.
\newblock \emph{2019 IEEE/CVF Conference on Computer Vision and Pattern
  Recognition (CVPR)}, pages 4869--4878, 2019.

\bibitem[Huang et~al.(2020)Huang, Qin, Liu, Zhu, and Shao]{Huang2020BNDynamic}
Lei Huang, Jie Qin, Li~Liu, Fan Zhu, and Ling Shao.
\newblock Layer-wise conditioning analysis in exploring the learning dynamics
  of dnns.
\newblock In Andrea Vedaldi, Horst Bischof, Thomas Brox, and Jan-Michael Frahm,
  editors, \emph{Computer Vision -- ECCV 2020}, pages 384--401. Springer
  International Publishing, 2020.
\newblock ISBN 978-3-030-58536-5.

\bibitem[Huang et~al.(2022)Huang, Zhou, Wang, Luo, and Liu]{2022_CVPR_Huang}
Lei Huang, Yi~Zhou, Tian Wang, Jie Luo, and Xianglong Liu.
\newblock Delving into the estimation shift of batch normalization in a
  network.
\newblock \emph{arXiv preprint arXiv:2203.10778}, 2022.

\bibitem[Huang et~al.(2015)Huang, Xu, and Yu]{Huang2015CRF}
Zhiheng Huang, Wei Xu, and Kai Yu.
\newblock Bidirectional lstm-crf models for sequence tagging.
\newblock \emph{ArXiv}, abs/1508.01991, 2015.

\bibitem[Ioffe(2017)]{2017_NIPS_Ioffe}
Sergey Ioffe.
\newblock Batch renormalization: Towards reducing minibatch dependence in
  batch-normalized models.
\newblock In \emph{NeurIPS}, 2017.

\bibitem[Ioffe and Szegedy(2015)]{Ioffe2015BN}
Sergey Ioffe and Christian Szegedy.
\newblock Batch normalization: Accelerating deep network training by reducing
  internal covariate shift.
\newblock In \emph{Proceedings of the 32nd International Conference on Machine
  Learning}, volume~37 of \emph{Proceedings of Machine Learning Research},
  pages 448--456. PMLR, 2015.

\bibitem[Izmailov et~al.(2018)Izmailov, Podoprikhin, Garipov, Vetrov, and
  Wilson]{2018_UAI_Izmailov}
Pavel Izmailov, Dmitrii Podoprikhin, Timur Garipov, Dmitry Vetrov, and
  Andrew~Gordon Wilson.
\newblock Averaging weights leads to wider optima and better generalization.
\newblock \emph{arXiv preprint arXiv:1803.05407}, 2018.

\bibitem[Krizhevsky(2009)]{Krizhevsky2009CIFAR10}
Alex Krizhevsky.
\newblock Learning multiple layers of features from tiny images.
\newblock 2009.

\bibitem[Krizhevsky et~al.(2012)Krizhevsky, Sutskever, and
  Hinton]{Kri2012AlexNet}
Alex Krizhevsky, Ilya Sutskever, and Geoffrey~E. Hinton.
\newblock Imagenet classification with deep convolutional neural networks.
\newblock In \emph{NIPS}, pages 1106--1114, 2012.

\bibitem[LeCun et~al.(2015)LeCun, Bengio, and Hinton]{Lecun2015DL}
Yann LeCun, Yoshua Bengio, and Geoffrey Hinton.
\newblock Deep learning.
\newblock \emph{Nature}, 521\penalty0 (7553):\penalty0 436--444, 2015.

\bibitem[Li et~al.(2019)Li, Chen, Hu, and Yang]{Li2019BNDropout}
Xiang Li, Shuo Chen, Xiaolin Hu, and Jian Yang.
\newblock Understanding the disharmony between dropout and batch normalization
  by variance shift.
\newblock In \emph{Proceedings of the IEEE/CVF Conference on Computer Vision
  and Pattern Recognition (CVPR)}, June 2019.

\bibitem[Li et~al.(2016)Li, Wang, Shi, Liu, and Hou]{2017_arxiv_Li}
Yanghao Li, Naiyan Wang, Jianping Shi, Jiaying Liu, and Xiaodi Hou.
\newblock Revisiting batch normalization for practical domain adaptation.
\newblock \emph{arXiv preprint arXiv:1603.04779}, 2016.

\bibitem[Luo et~al.(2019{\natexlab{a}})Luo, Ren, Peng, Zhang, and
  Li]{2019_ICLR_Luo}
Ping Luo, Jiamin Ren, Zhanglin Peng, Ruimao Zhang, and Jingyu Li.
\newblock Differentiable learning-to-normalize via switchable normalization.
\newblock In \emph{ICLR}, 2019{\natexlab{a}}.

\bibitem[Luo et~al.(2019{\natexlab{b}})Luo, Wang, Shao, and Peng]{Luo2018BNReg}
Ping Luo, Xinjiang Wang, Wenqi Shao, and Zhanglin Peng.
\newblock Towards understanding regularization in batch normalization.
\newblock In \emph{International Conference on Learning Representations},
  2019{\natexlab{b}}.

\bibitem[Ma et~al.(2019)Ma, Zhang, Zhang, Duan, Hou, Zhou, and
  Song]{Ma2019TensorTrans}
Xindian Ma, Peng Zhang, Shuai Zhang, Nan Duan, Yuexian Hou, Ming Zhou, and
  Dawei Song.
\newblock A tensorized transformer for language modeling.
\newblock In H.~Wallach, H.~Larochelle, A.~Beygelzimer, F.~d\textquotesingle
  Alch\'{e}-Buc, E.~Fox, and R.~Garnett, editors, \emph{Advances in Neural
  Information Processing Systems}, volume~32. Curran Associates, Inc., 2019.

\bibitem[Maas et~al.(2011)Maas, Daly, Pham, Huang, Ng, and Potts]{Maas2011IMDB}
Andrew~L. Maas, Raymond~E. Daly, Peter~T. Pham, Dan Huang, Andrew~Y. Ng, and
  Christopher Potts.
\newblock Learning word vectors for sentiment analysis.
\newblock In \emph{Proceedings of the 49th Annual Meeting of the Association
  for Computational Linguistics: Human Language Technologies}, pages 142--150.
  Association for Computational Linguistics, 2011.

\bibitem[Martens and Grosse(2015)]{Marten2015KFAC}
James Martens and Roger~B. Grosse.
\newblock Optimizing neural networks with kronecker-factored approximate
  curvature.
\newblock \emph{CoRR}, abs/1503.05671, 2015.

\bibitem[Merity et~al.(2016)Merity, Xiong, Bradbury, and
  Socher]{MerityX2016WikiText}
Stephen Merity, Caiming Xiong, James Bradbury, and Richard Socher.
\newblock Pointer sentinel mixture models.
\newblock \emph{CoRR}, abs/1609.07843, 2016.

\bibitem[Mikolov et~al.(2011)Mikolov, Deoras, Kombrink, Burget, and
  Cernocky]{Mikolov2011PTB}
Tomas Mikolov, Anoop Deoras, Stefan Kombrink, Lukas Burget, and Jan~"Honza"
  Cernocky.
\newblock Empirical evaluation and combination of advanced language modeling
  techniques.
\newblock In \emph{Interspeech}. ISCA, August 2011.

\bibitem[Nado et~al.(2020)Nado, Padhy, Sculley, D'Amour, Lakshminarayanan, and
  Snoek]{2020_arxiv_Nado}
Zachary Nado, Shreyas Padhy, D~Sculley, Alexander D'Amour, Balaji
  Lakshminarayanan, and Jasper Snoek.
\newblock Evaluating prediction-time batch normalization for robustness under
  covariate shift.
\newblock \emph{arXiv preprint arXiv:2006.10963}, 2020.

\bibitem[Ott et~al.(2019)Ott, Edunov, Baevski, Fan, Gross, Ng, Grangier, and
  Auli]{Ott2019Fairseq}
Myle Ott, Sergey Edunov, Alexei Baevski, Angela Fan, Sam Gross, Nathan Ng,
  David Grangier, and Michael Auli.
\newblock fairseq: A fast, extensible toolkit for sequence modeling.
\newblock In \emph{Proceedings of the 2019 Conference of the North {A}merican
  Chapter of the Association for Computational Linguistics (Demonstrations)},
  pages 48--53. Association for Computational Linguistics, 2019.

\bibitem[Papineni et~al.(2002)Papineni, Roukos, Ward, and Zhu]{Pap2002Bleu}
Kishore Papineni, Salim Roukos, Todd Ward, and Wei-Jing Zhu.
\newblock Bleu: A method for automatic evaluation of machine translation.
\newblock In \emph{Proceedings of the 40th Annual Meeting on Association for
  Computational Linguistics}, ACL '02, page 311–318. Association for
  Computational Linguistics, 2002.

\bibitem[Radford et~al.(2019)Radford, Wu, Child, Luan, Amodei, and
  Sutskever]{Radford2019GPT2}
Alec Radford, Jeff Wu, Rewon Child, David Luan, Dario Amodei, and Ilya
  Sutskever.
\newblock Language models are unsupervised multitask learners.
\newblock 2019.

\bibitem[Sang and Meulder(2003)]{Sang2003CoNLL}
Erik Tjong~Kim Sang and Fien~De Meulder.
\newblock Introduction to the conll-2003 shared task: Language-independent
  named entity recognition.
\newblock In \emph{CoNLL}, 2003.

\bibitem[Santurkar et~al.(2018)Santurkar, Tsipras, Ilyas, and
  Madry]{San2018BnHelpOpt}
Shibani Santurkar, Dimitris Tsipras, Andrew Ilyas, and Aleksander Madry.
\newblock How does batch normalization help optimization?
\newblock In S.~Bengio, H.~Wallach, H.~Larochelle, K.~Grauman, N.~Cesa-Bianchi,
  and R.~Garnett, editors, \emph{Advances in Neural Information Processing
  Systems}, volume~31, 2018.

\bibitem[Schneider et~al.(2020)Schneider, Rusak, Eck, Bringmann, Brendel, and
  Bethge]{2020_NIPS_Schneider}
Steffen Schneider, Evgenia Rusak, Luisa Eck, Oliver Bringmann, Wieland Brendel,
  and Matthias Bethge.
\newblock Improving robustness against common corruptions by covariate shift
  adaptation.
\newblock In \emph{NeurIPS}, 2020.

\bibitem[Shen et~al.(2020)Shen, Yao, Gholami, Mahoney, and
  Keutzer]{Shen2020Powernorm}
Sheng Shen, Zhewei Yao, Amir Gholami, Michael Mahoney, and Kurt Keutzer.
\newblock Powernorm: Rethinking batch normalization in transformers.
\newblock In \emph{ICML}, 2020.

\bibitem[Singh and Shrivastava(2019)]{2019_ICCV_Singh}
Saurabh Singh and Abhinav Shrivastava.
\newblock Evalnorm: Estimating batch normalization statistics for evaluation.
\newblock In \emph{ICCV}, 2019.

\bibitem[Summers and Dinneen(2020)]{2020_ICLR_Summers}
Cecilia Summers and Michael~J. Dinneen.
\newblock Four things everyone should know to improve batch normalization.
\newblock In \emph{ICLR}, 2020.

\bibitem[Vaswani et~al.(2017)Vaswani, Shazeer, Parmar, Uszkoreit, Jones, Gomez,
  Kaiser, and Polosukhin]{Vaswani2017Transformer}
Ashish Vaswani, Noam Shazeer, Niki Parmar, Jakob Uszkoreit, Llion Jones,
  Aidan~N Gomez, \L~ukasz Kaiser, and Illia Polosukhin.
\newblock Attention is all you need.
\newblock In I.~Guyon, U.~V. Luxburg, S.~Bengio, H.~Wallach, R.~Fergus,
  S.~Vishwanathan, and R.~Garnett, editors, \emph{Advances in Neural
  Information Processing Systems}, volume~30. Curran Associates, Inc., 2017.

\bibitem[Wu and He(2018)]{2018_ECCV_Wu}
Yuxin Wu and Kaiming He.
\newblock Group normalization.
\newblock In \emph{ECCV}, 2018.

\bibitem[Wu and Johnson(2021)]{Wu2021RevistBN}
Yuxin Wu and Justin Johnson.
\newblock Rethinking "batch" in batchnorm.
\newblock \emph{CoRR}, abs/2105.07576, 2021.

\bibitem[Xiong et~al.(2020)Xiong, Yang, He, Zheng, Zheng, Xing, Zhang, Lan,
  Wang, and Liu]{Xiong2020NoWarmUp}
Ruibin Xiong, Yunchang Yang, Di~He, Kai Zheng, Shuxin Zheng, Chen Xing,
  Huishuai Zhang, Yanyan Lan, Liwei Wang, and Tie-Yan Liu.
\newblock On layer normalization in the transformer architecture.
\newblock In \emph{ICML 2020}, 2020.

\bibitem[Xu et~al.(2019)Xu, Sun, Zhang, Zhao, and Lin]{Xu2019UnderstandLN}
Jingjing Xu, Xu~Sun, Zhiyuan Zhang, Guangxiang Zhao, and Junyang Lin.
\newblock Understanding and improving layer normalization.
\newblock In \emph{Advances in Neural Information Processing Systems},
  volume~32, 2019.

\bibitem[Yan et~al.(2019)Yan, Deng, Li, and Qiu]{Yan2019TENER}
Hang Yan, Bocao Deng, Xiaonan Li, and Xipeng Qiu.
\newblock Tener: Adapting transformer encoder for named entity recognition,
  2019.

\bibitem[Yan et~al.(2020)Yan, Wan, Zhang, Zhang, Wei, and Sun]{2020_ICLR_Yan}
Junjie Yan, Ruosi Wan, Xiangyu Zhang, Wei Zhang, Yichen Wei, and Jian Sun.
\newblock Towards stabilizing batch statistics in backward propagation of batch
  normalization.
\newblock In \emph{ICLR}, 2020.

\bibitem[Yao et~al.(2021{\natexlab{a}})Yao, Cao, Lin, Liu, Zhang, and
  Hu]{Yao2021LeveragingBN}
Zhuliang Yao, Yue Cao, Yutong Lin, Ze~Liu, Zheng Zhang, and Han Hu.
\newblock Leveraging batch normalization for vision transformers.
\newblock \emph{2021 IEEE/CVF International Conference on Computer Vision
  Workshops (ICCVW)}, pages 413--422, 2021{\natexlab{a}}.

\bibitem[Yao et~al.(2021{\natexlab{b}})Yao, Cao, Zheng, Huang, and
  Lin]{2021_CVPR_Yao}
Zhuliang Yao, Yue Cao, Shuxin Zheng, Gao Huang, and Stephen Lin.
\newblock Cross-iteration batch normalization.
\newblock In \emph{CVPR}, 2021{\natexlab{b}}.

\bibitem[Yong et~al.(2020)Yong, Huang, Hua, and Zhang]{2020_ECCV_Yong}
Hongwei Yong, Jianqiang Huang, Xiansheng Hua, and Lei Zhang.
\newblock Gradient centralization: A new optimization technique for deep neural
  networks.
\newblock In \emph{ECCV}, 2020.

\bibitem[Zhang and Sennrich(2019)]{Zhang2019RMSNorm}
Biao Zhang and Rico Sennrich.
\newblock Root mean square layer normalization.
\newblock In H.~Wallach, H.~Larochelle, A.~Beygelzimer, F.~d\textquotesingle
  Alch\'{e}-Buc, E.~Fox, and R.~Garnett, editors, \emph{Advances in Neural
  Information Processing Systems}, volume~32. Curran Associates, Inc., 2019.

\bibitem[Zhang and Yang(2018)]{Zhang2018Resume}
Yue Zhang and Jie Yang.
\newblock Chinese ner using lattice lstm.
\newblock In \emph{ACL}, 2018.

\end{thebibliography}

\newpage
\appendix

\section{Experimental Details}
\label{sec:app_exp}
In the main paper, we sketch the experimental settings. Here, we report the details. Our WMT16 experiments require two NVIDIA 3090 GPUs with 24GB memory each, and other experiments are implemented with one NVIDIA 3090 GPU. 
\subsubsection*{Neural Machine Translation}
\paragraph{Datasets}
We use two widely used datasets: IWSLT14 German-to-English (De-En) dataset and WMT16 English-to-German (En-De) dataset. IWSLT14/WMT16 contains 0.15M/4.5M sentence pairs. 
\paragraph{Configurations}
Our code is based on \textit{fairseq}~\cite{Ott2019Fairseq}\footnote{https://github.com/pytorch/fairseq. MIT license.}. We use six encoder layers and six decoder layers for both datasets. We keep the Transformer decoder unchanged and only modify normalization layers in Transformer encoder throughout NMT experiments. We follow the model settings in \cite{Shen2020Powernorm} for IWSLT14 and apply $\text{Transformer}_{base}$ model~\cite{Vaswani2017Transformer} for WMT16. At test phase, we averaged the last six checkpoints and measure case sensitive tokenized BLEU~\cite{Pap2002Bleu} with beam size 4/5 and length penalty 0.6/1.0 for ISWLT14/WMT16. We use Adam with $\left(\beta_1, \beta_2\right)=\left(0.9, 0.98\right)$, an inverse square root learning rate scheduler, and a warmup stage with $8000$ steps. We apply labeling smoothing $\epsilon_{ls}=0.1$. For IWSLT14, we set num-tokens=4096, max-epochs=60, dropout=0.3, attention dropout=0.1, activation dropout=0.1 and lr=$5e^{-4}$/$1.5e^{-3}$ for Post-Norm/Pre-Norm Transformer.  For WMT16, we set num-tokens=8192, update-freq=4, max-epochs=20, dropout=0.1, attention dropout=0.1, activation dropout=0.1 and lr=$7e^{-4}$/$2e^{-3}$ for Post-Norm/Pre-Norm Transformer. 
\subsubsection*{Language Modeling}
\paragraph{Datasets} 
We conduct experiments on PTB (0.93M tokens)~\cite{Mikolov2011PTB} and WikiText-103 (WT103)(100M tokens)~\cite{MerityX2016WikiText}. We follow the evaluation scheme in \cite{Dai2019TransXL} and use perplexity (PPL) of test set to compare the model performance. 
\paragraph{Configurations} 
Following the experimental settings in \cite{Shen2020Powernorm}\cite{Ma2019TensorTrans}, we use three and six layers tensorized transformer core-1 for PTB and Wikitext-103 separately. We use the same hyperparameters for Post-Norm and Pre-Norm Transformer. We use Adam optimizer and set lr=$2.5e^{-4}$ with linear decay. For PTB, we use dropout=0.3, batch size=120, max-steps=20000.  For PTB, we use dropout=0.1, batch size=60, max-steps=200000.
\subsubsection*{Named Entity Recognition}
\paragraph{Datasets} 
We choose two widely used NER datasets: CoNLL2003 (English)~\cite{Sang2003CoNLL} and Resume (Chinese)~\cite{Zhang2018Resume}. CoNLL2003/Resume contains four/eight kinds of named entities. CoNLL2003 contains 14.0k/3.2k/3.5k sentences for train/dev/test split while Resume includes 3.8k/0.5k/0.5k sentences for train/dev/test set.  We use F1 score to measure the model performance. 
\paragraph{Configurations} 
We mainly follow the experimental settings in \cite{Yan2019TENER}. We use two and four layers transformer encoder for CoNLL2003 and Resume, respectively. An additional CRF layer~\cite{Huang2015CRF} is added to model the transition probability. We use SGD optimizer with 0.9 momentum and 1\% total steps as warmup steps. We set the learning rate to be $9e^{-4}$ and $7e^{-4}$ for CoNLL2003 and Resume.
\subsubsection*{Text Classification}
\paragraph{Datasets}
For text classification, we use the code\footnote{https://github.com/declare-lab/identifiable-transformers. Apache-2.0 license.} and most configurations in \cite{Bhardwaj2021TextClsTrans}. We select one small scale dataset (IMDB~\cite{Maas2011IMDB}) and three large scale datasets (Yelp, DBPedia, Sogou News), including two sentiment classification tasks (IMDB, Yelp) and two topic classification tasks (DBPedia, Sogou News). 
\paragraph{Configurations}
We use six Transformer encoder layers with a CLS token at the beginning of each sentence for classification. We extract 30\% of training data as validation set for all datasets. The best checkpoint for the validation is applied in the test phase. We run three times with different random seeds for each setting and report the mean accuracy.
\section{Optimal Hyperparameters}
\label{sec:optimal_hyperpara}
For RBN, we choose $\lambda, \nu$ both from $\left\{0,0.01,0.1,1\right\}$ by validation loss. We empirically find that increasing the mean penalty on NMT for Post-Norm Transformer can further improve the BLEU scores. Thus, we apply $\left( \lambda, \nu \right)=\left(10,0\right)\slash\left(100,0\right)$ on ISWLT14/WMT16 for Post-Norm Transformer exceptionally. 
\begin{table}[h]
	\centering
	\caption{Optimal hyperparameters $\left(\lambda, \nu\right)$ for each experimental setting. $\left(\lambda, \nu\right)$ are mean and variance penalty coefficients separately.}
	\resizebox{\columnwidth}{!}{
		\begin{tabular}{@{}ccccccccccc@{}}
			
			\toprule
			Task      & \multicolumn{2}{c}{NMT}            & \multicolumn{2}{c}{LM}  & \multicolumn{2}{c}{NER}      & \multicolumn{4}{c}{TextCls}                     \\ \midrule
			Datasets  & IWSLT14         & WMT16            & PTB        & WT103      & Resume            & CoNLL    & IMDB     & Sogou      & DBPedia    & Yelp       \\ \midrule
			Post-Norm & (10,0) & (100,0) & (0.1,0.01) & (0.1,0.01) &(0.01,0)& (0.01,0) & (0.1,0)  & (0.1,0.01) & (0.1,0.01) & (0,0.1)    \\
			Pre-Norm  & (0.1,0.01)      & (0.1,0)          & (0.01,0)   & (0.1,0)    & (0.01,0)          & (0.01,0) & (0,0.01) & (0.1,0.01) & (0.1,0.01) & (0.1,0.01) \\ \bottomrule
		\end{tabular}
	}
	
\end{table}

\section{Average TID of BN and RBN}
\label{sec:avg_tid_compare}
In the main paper, we have shown the TID of the last BN (RBN) layer on various NLP datasets with Post-Norm or Pre-Norm Transformer. Here, we report the average TID of all BN (RBN) layers (\cref{tab:tid_avg}). RBN decreases the average TID of BN.
\begin{table}[h]
	\centering
	\caption{Average TID of all BN/RBN layer in Post-Norm and Pre-Norm Transformers on various natural language tasks at the end of training. RBN reduces the TID of BN effectively.}
	\resizebox{\columnwidth}{!}{
		\begin{tabular}{@{}ccccccccccc@{}}
			\toprule
			Task                  & \multicolumn{2}{c}{NMT}       & \multicolumn{2}{c}{LM}        & \multicolumn{2}{c}{NER}       & \multicolumn{4}{c}{TextCls}                                   \\ \midrule
			Datasets              & IWSLT14       & WMT16         & PTB           & WT103         & Resume        & CoNLL         & IMDB          & Sogou         & DBPedia       & Yelp          \\ \midrule
			\multicolumn{11}{c}{Post-Norm Transformer} \\ \midrule
			Mean TID of BN\_avg   & 2.8\%         & 3.0\%         & 1.0\%         & 0.9\%         & 5.0\%         & 9.9\%         & 1.5\%         & 1.9\%         & 1.3\%         & 2.7\%         \\
			Mean TID of RBN\_avg  & 0.8\%         & 1.5\%         & 1.0\%         & 1.0\%         & 4.4\%         & 5.6\%         & 0.4\%         & 0.4\%         & 0.4\%         & 0.4\%         \\
			Var TID of BN\_avg    & 7.1\%         & 4.9\%         & 1.1\%         & 2.2\%         & 4.2\%         & 9.8\%         & 3.1\%         & 2.1\%         & 1.4\%         & 3.9\%         \\
			Var TID of RBN\_avg   & 2.6\%         & 2.8\%         & 1.0\%         & 1.0\%         & 4.0\%         & 6.5\%         & 1.7\%         & 0.3\%         & 0.3\%         & 0.2\%         \\
			\midrule \midrule
			\multicolumn{11}{c}{Pre-Norm Transformer} \\ \midrule 
			Mean TID of BN\_avg   & 3.2\%         & 7.6\%         & 1.6\%         & 2.1\%         & 10.3\%        & 10.8\%        & 2.1\%         & 4.2\%         & 2.5\%         & 4.2\%         \\
			Mean TID of RBN\_avg  & 3.0\%         & 1.6\%         & 1.6\%         & 2.0\%         & 6.7\%         & 5.9\%         & 0.8\%         & 1.2\%         & 1.2\%         & 1.1\%         \\
			Var TID of BN\_avg    & 5.5\%         & 12.8\%        & 1.6\%         & 1.9\%         & 8.1\%         & 7.4\%         & 3.6\%         & 3.7\%         & 2.1\%         & 5.1\%         \\
			Var TID of RBN\_avg   & 1.6\%         & 5.6\%         & 1.6\%         & 1.8\%         & 7.5\%         & 6.1\%         & 1.9\%         & 0.5\%         & 0.5\%         & 0.5\%         \\
			\bottomrule
		\end{tabular}
	}
	\label{tab:tid_avg}
\end{table}

\section{Figures of TID Through Training}
\label{sec:train_tid}
\begin{figure}[h]
	\captionsetup[subfigure]{justification=centering}
	\centering
	\begin{subfigure}[b]{0.24\columnwidth}
		\centering
		\centerline{\includegraphics[width=\textwidth]{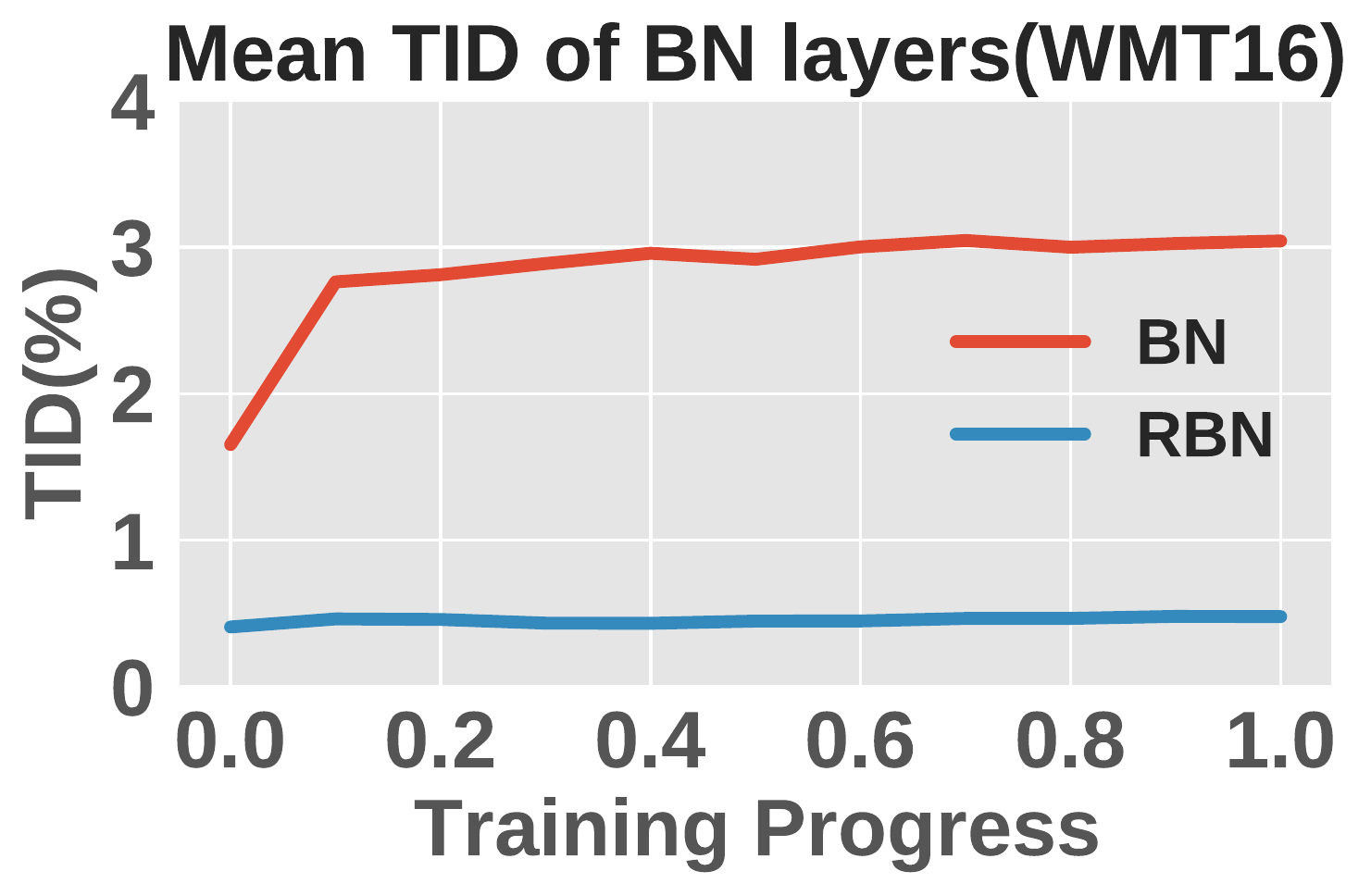}}
	\end{subfigure}
	\hfill
	\begin{subfigure}[b]{0.24\columnwidth}
		\centering
		\includegraphics[width=\textwidth]{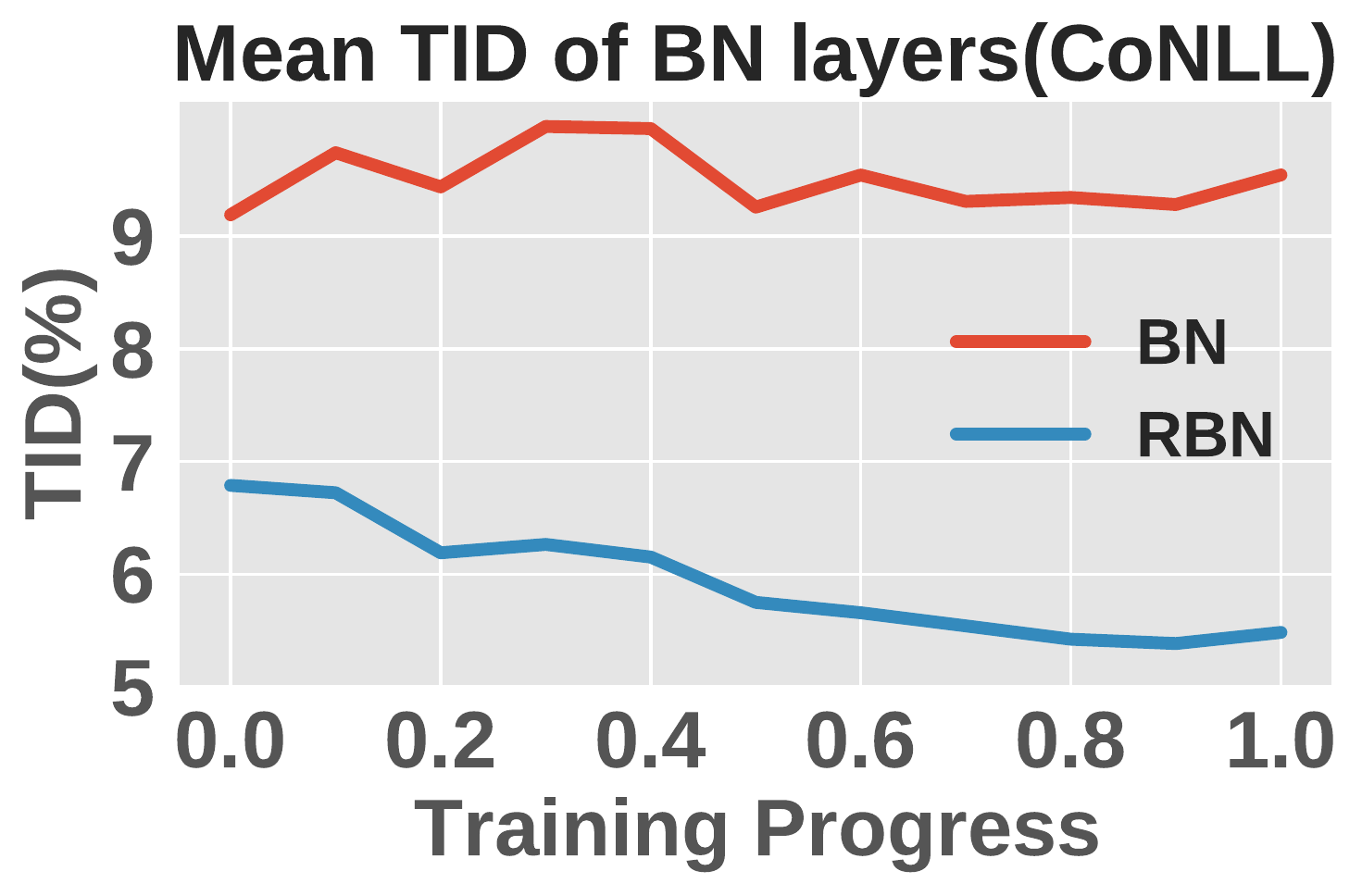}
	\end{subfigure}
	\hfill
	\begin{subfigure}[b]{0.24\columnwidth}
		\centering
		\includegraphics[width=\textwidth]{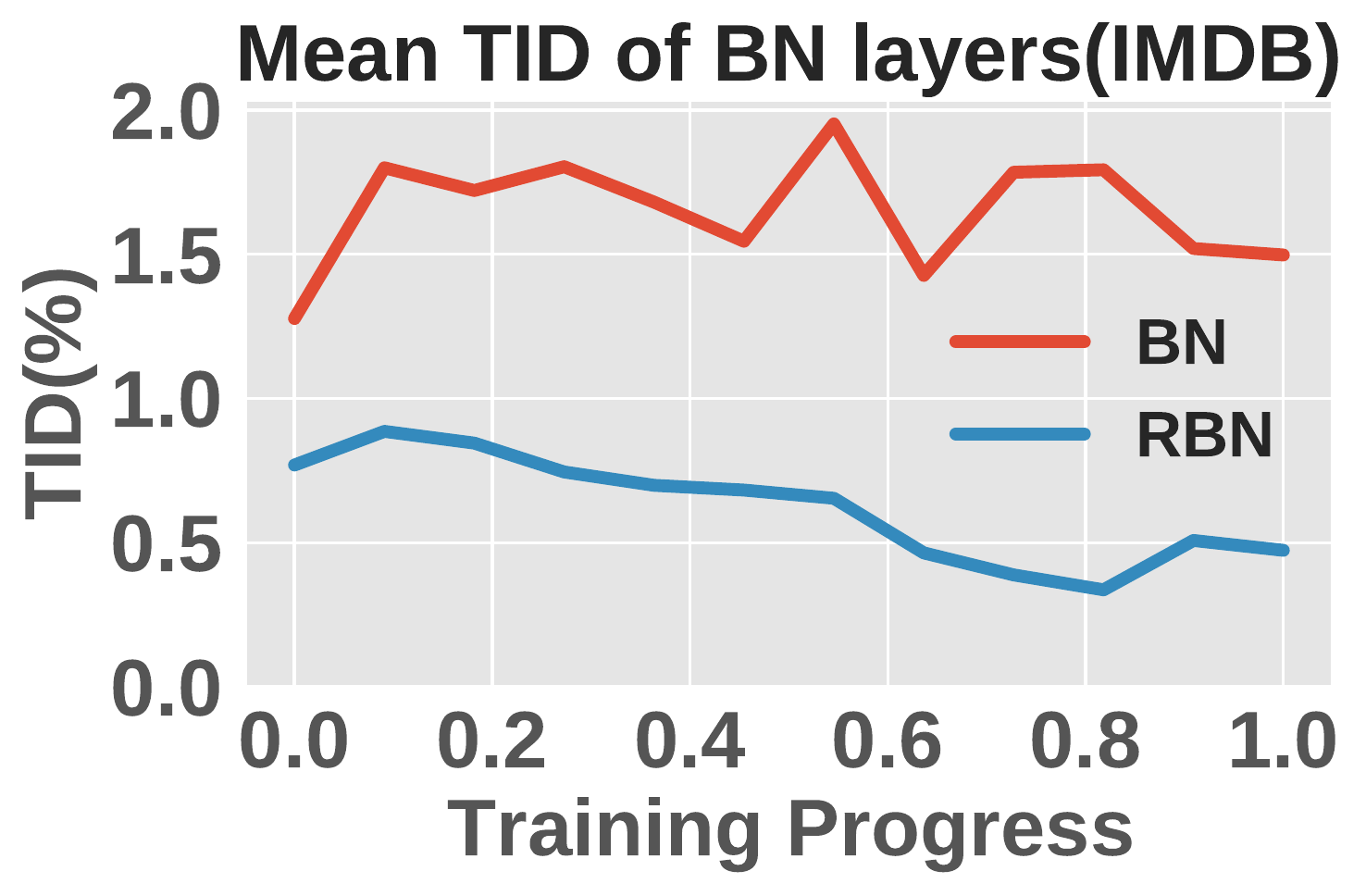}
	\end{subfigure}
	\hfill
	\begin{subfigure}[b]{0.24\columnwidth}
		\centering
		\includegraphics[width=\textwidth]{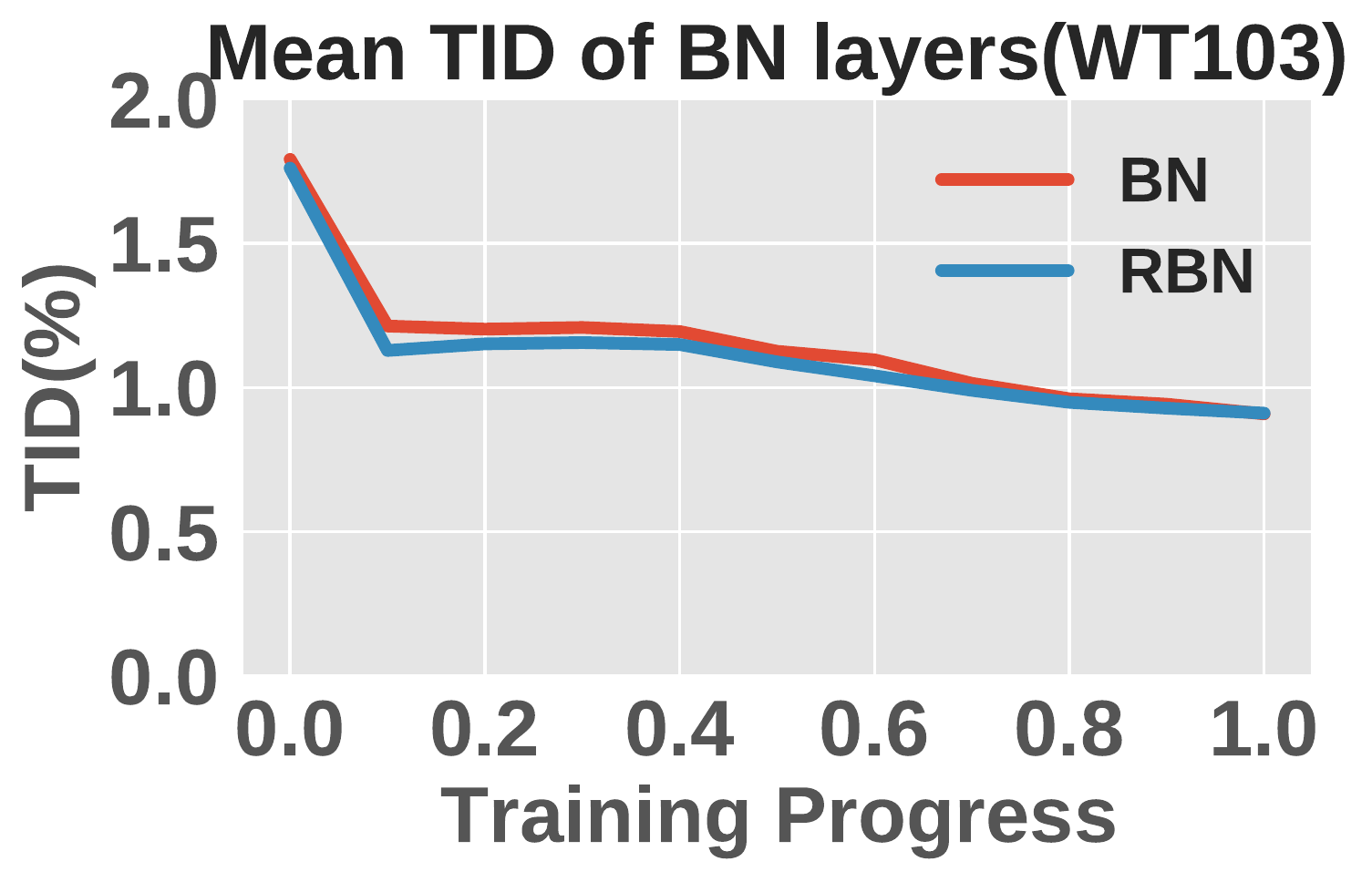}
	\end{subfigure}
	\begin{subfigure}[b]{0.24\columnwidth}
		\centering
		\centerline{\includegraphics[width=\textwidth]{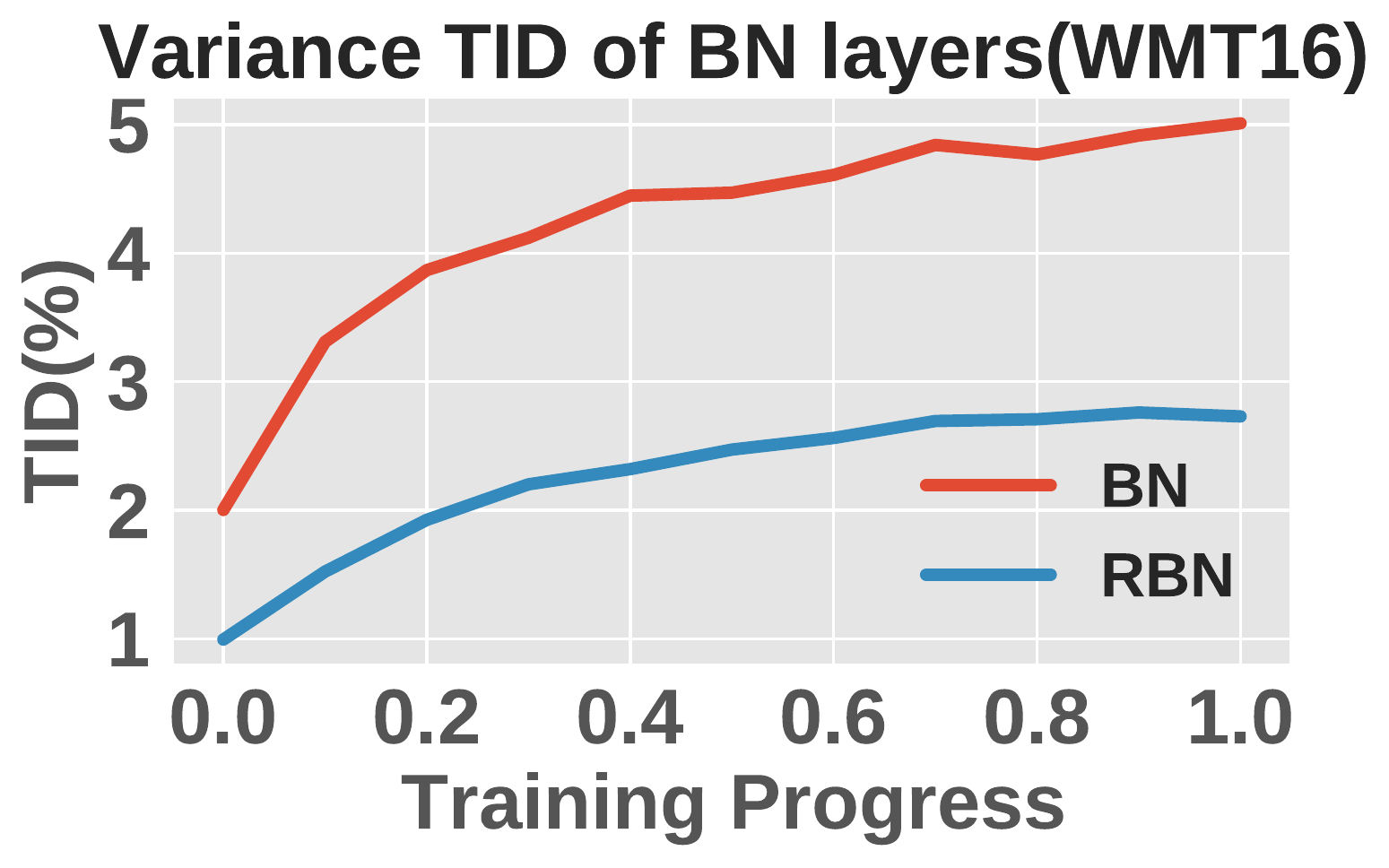}}
	\end{subfigure}
	\hfill
	\begin{subfigure}[b]{0.24\columnwidth}
		\centering
		\includegraphics[width=\textwidth]{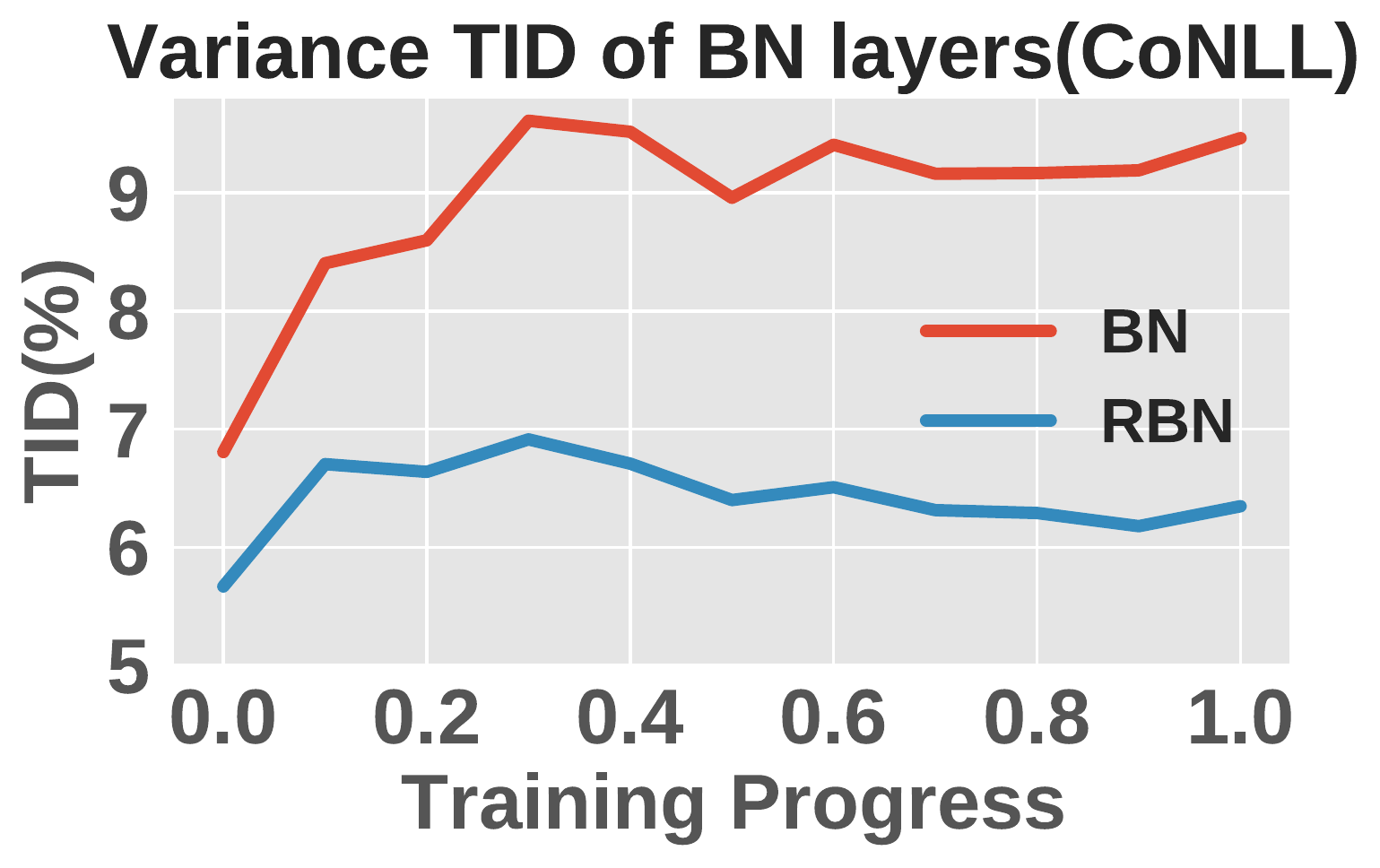}
	\end{subfigure}
	\hfill
	\begin{subfigure}[b]{0.24\columnwidth}
		\centering
		\includegraphics[width=\textwidth]{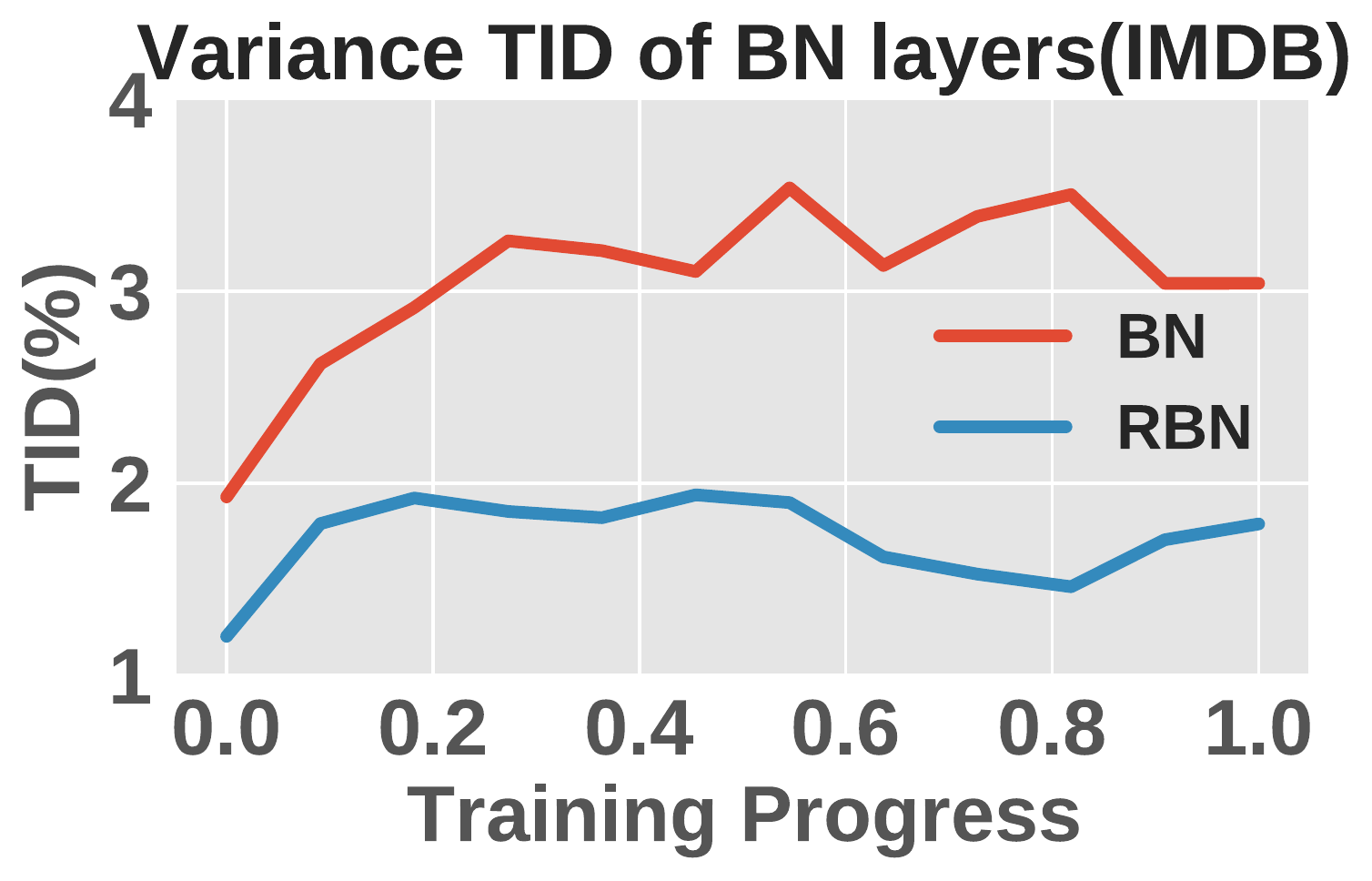}
	\end{subfigure}
	\hfill
	\begin{subfigure}[b]{0.24\columnwidth}
		\centering
		\includegraphics[width=\textwidth]{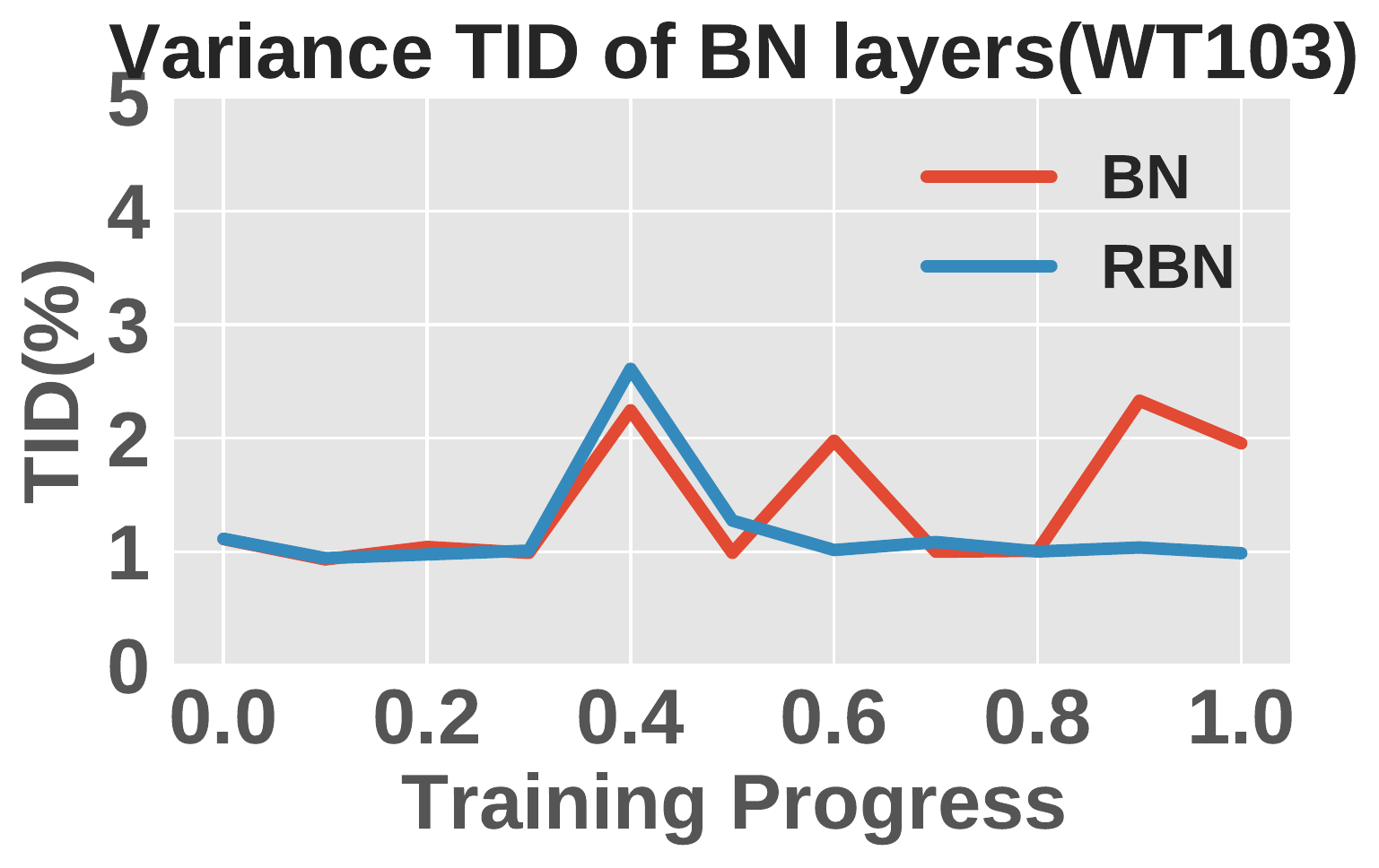}
	\end{subfigure}
	\caption{Averaged Mean and Variance TID on WMT16/CoNLL/IMDB/WT103 for Post-Norm Transformer with BN and RBN.}
	\label{fig:mean_var_tid_post}
\end{figure}

\begin{figure}[h]
	\captionsetup[subfigure]{justification=centering}
	\centering
	\begin{subfigure}[b]{0.24\columnwidth}
		\centering
		\centerline{\includegraphics[width=\textwidth]{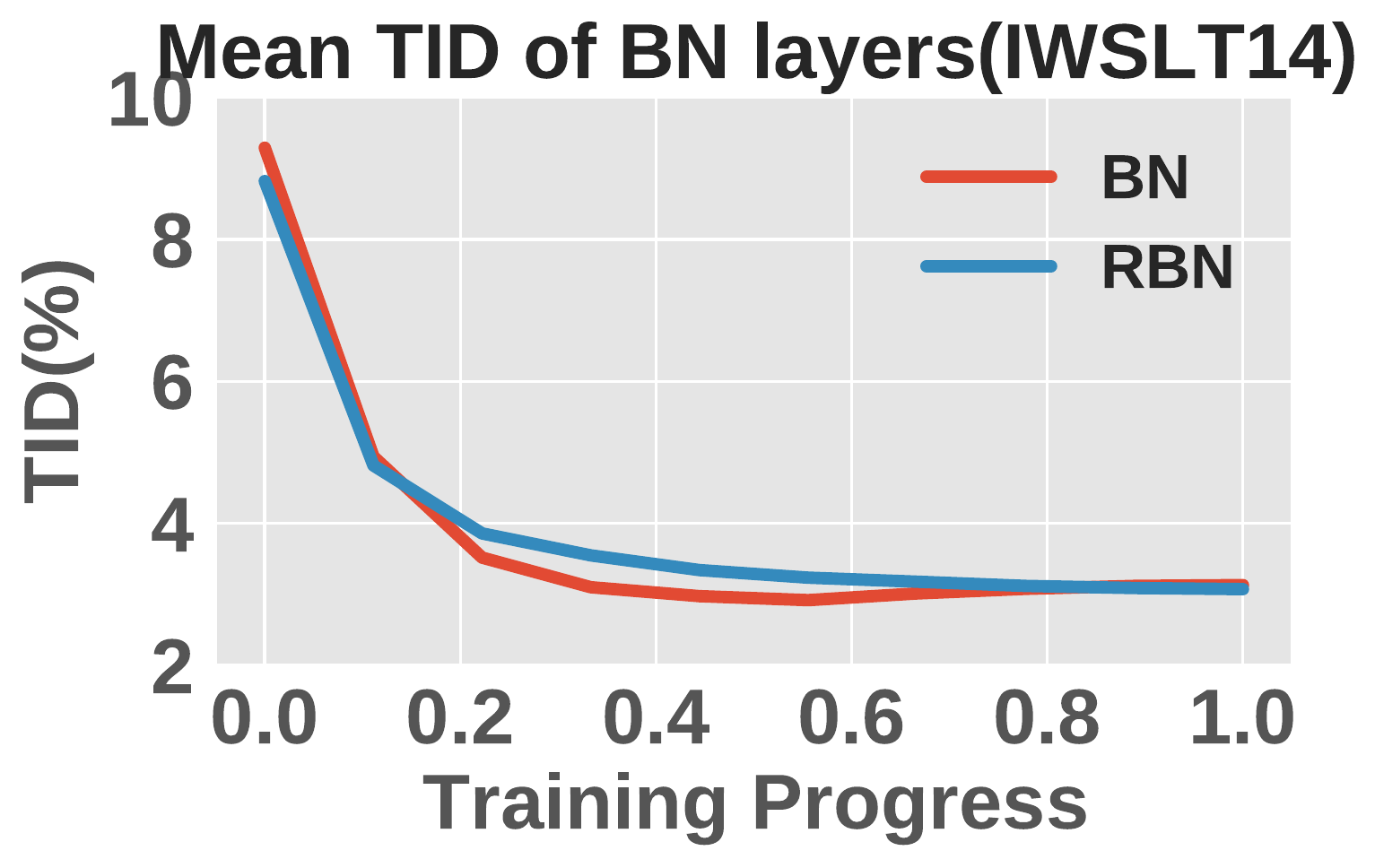}}
	\end{subfigure}
	\hfill
	\begin{subfigure}[b]{0.24\columnwidth}
		\centering
		\includegraphics[width=\textwidth]{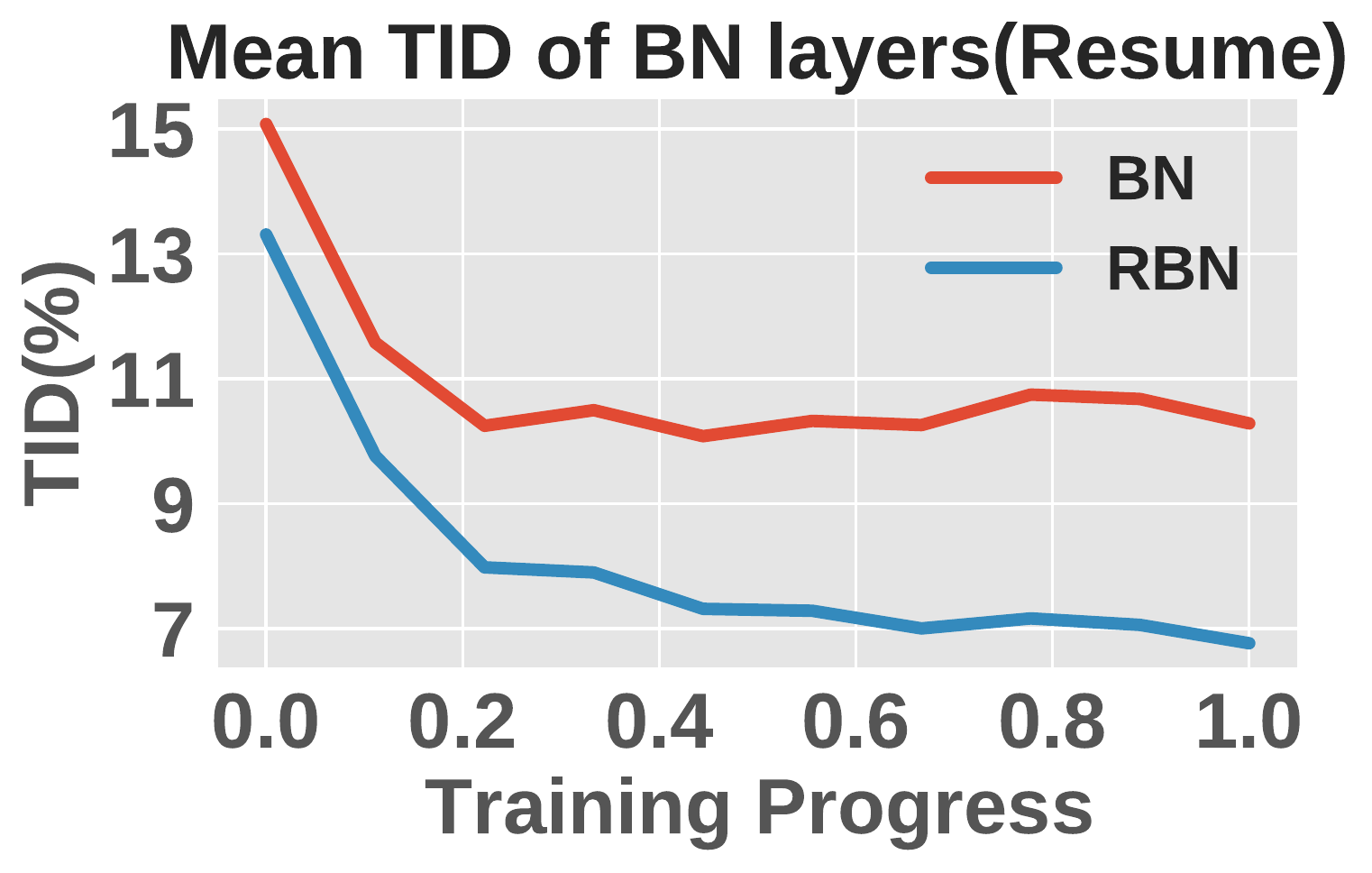}
	\end{subfigure}
	\hfill
	\begin{subfigure}[b]{0.24\columnwidth}
		\centering
		\includegraphics[width=\textwidth]{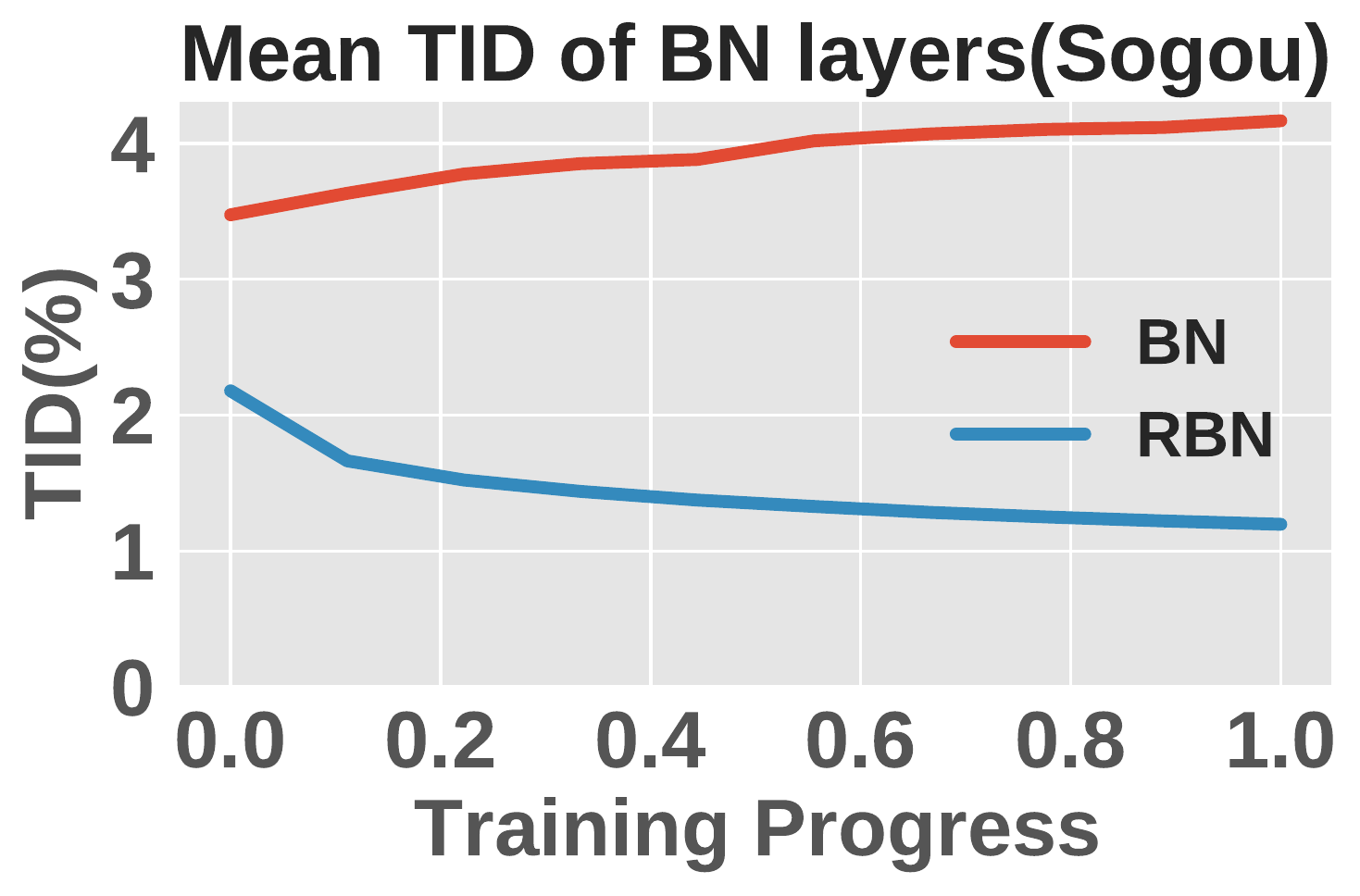}
	\end{subfigure}
	\hfill
	\begin{subfigure}[b]{0.24\columnwidth}
		\centering
		\includegraphics[width=\textwidth]{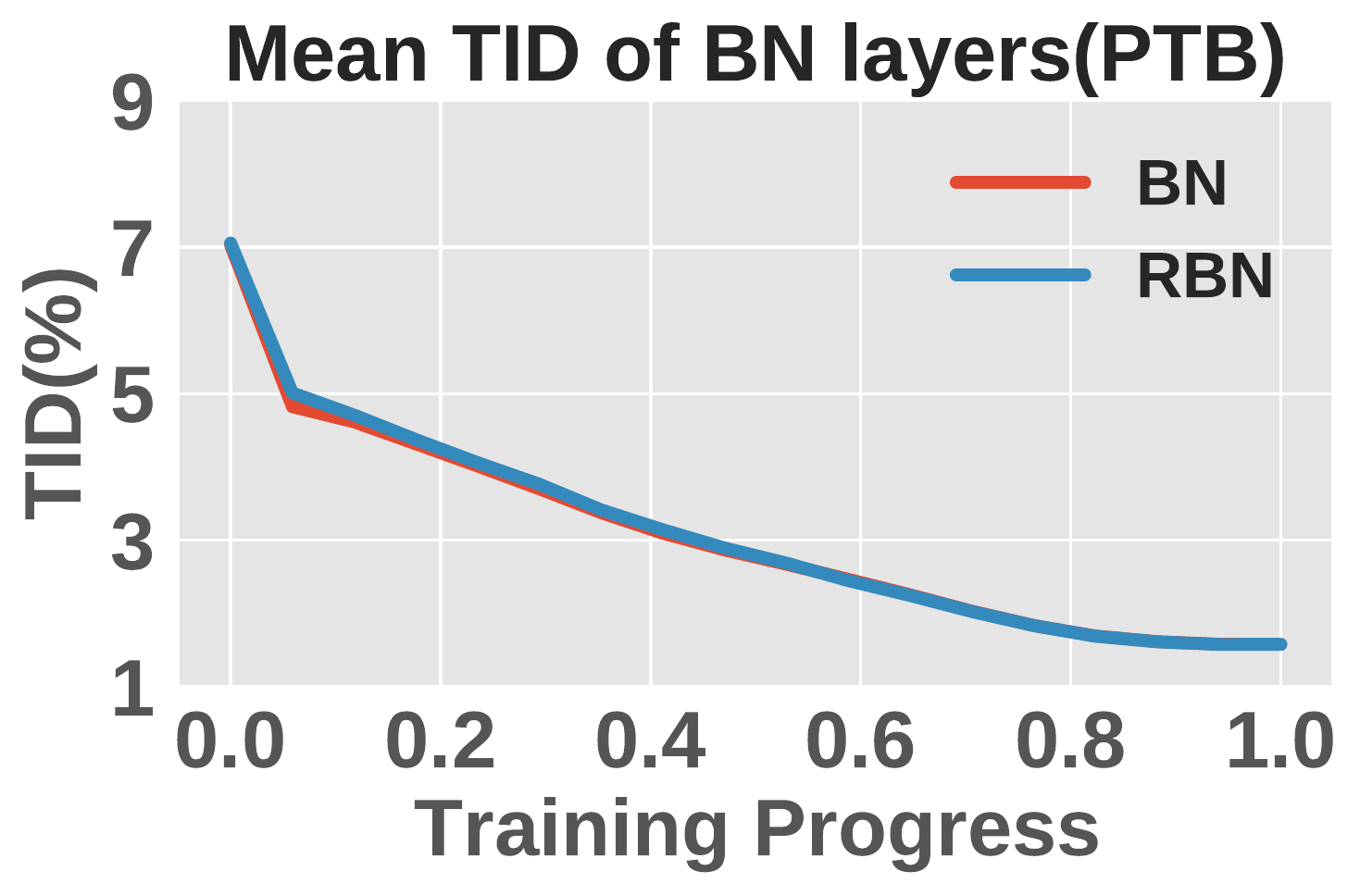}
	\end{subfigure}
	\begin{subfigure}[b]{0.24\columnwidth}
		\centering
		\centerline{\includegraphics[width=\textwidth]{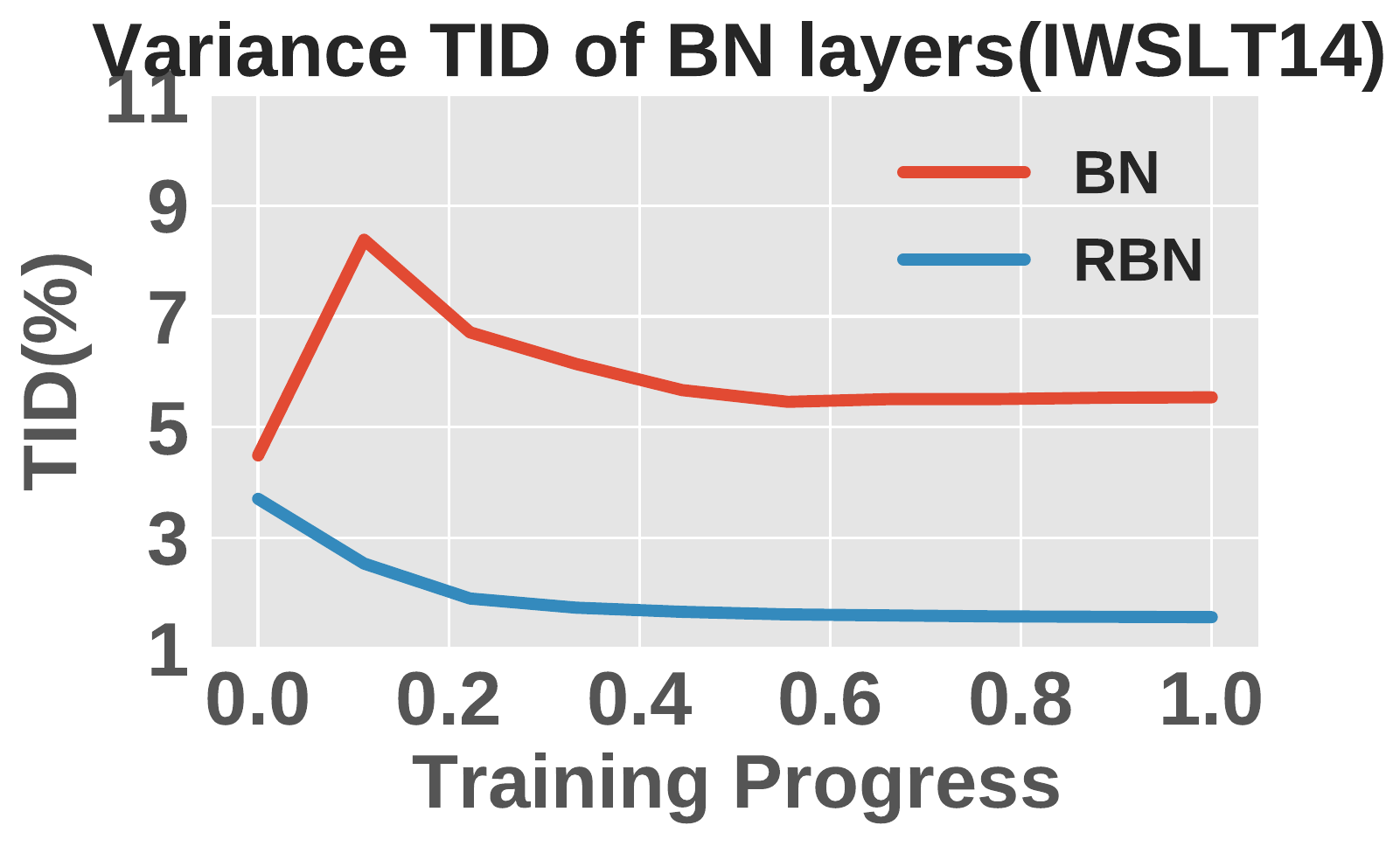}}
	\end{subfigure}
	\hfill
	\begin{subfigure}[b]{0.24\columnwidth}
		\centering
		\includegraphics[width=\textwidth]{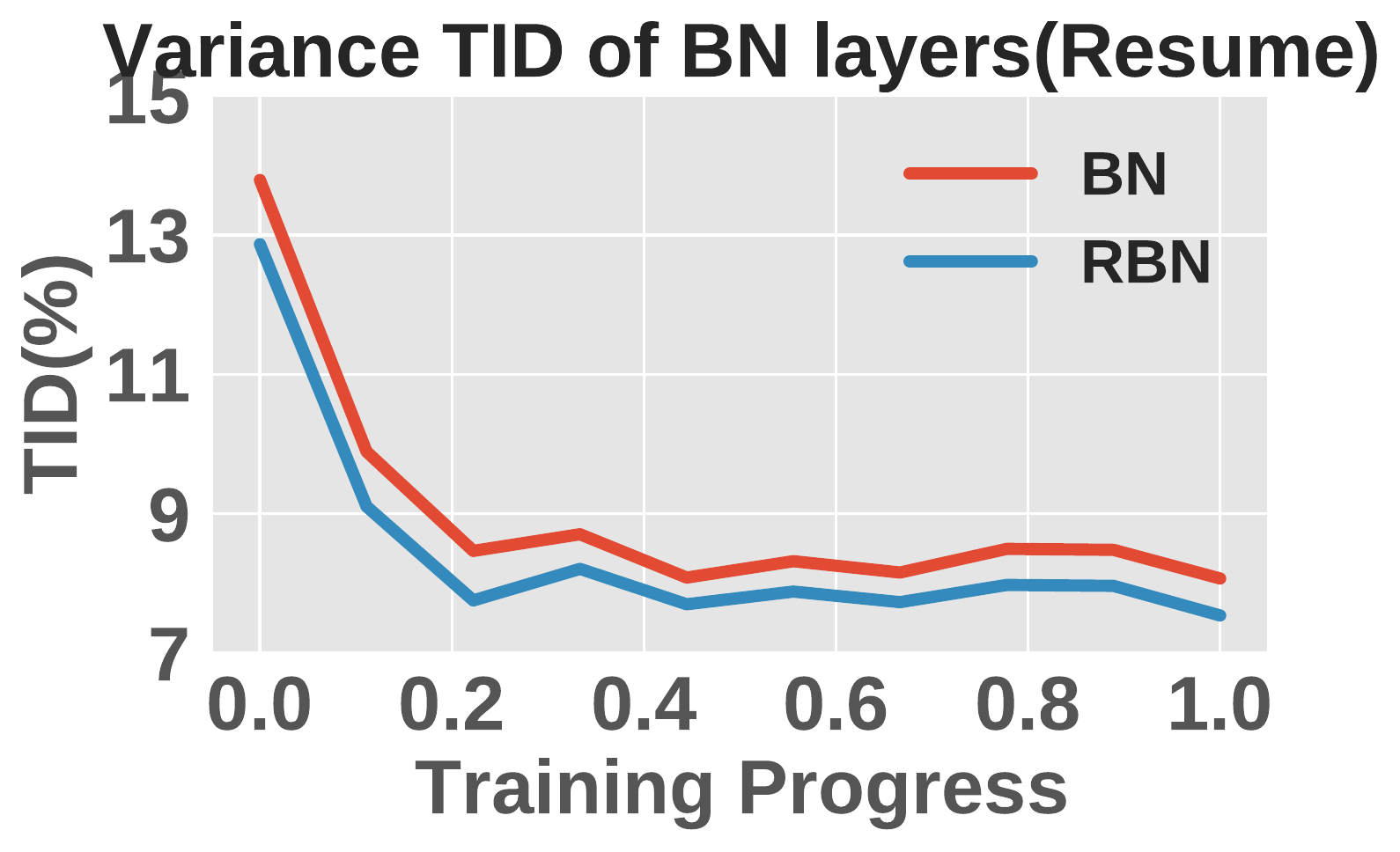}
	\end{subfigure}
	\hfill
	\begin{subfigure}[b]{0.24\columnwidth}
		\centering
		\includegraphics[width=\textwidth]{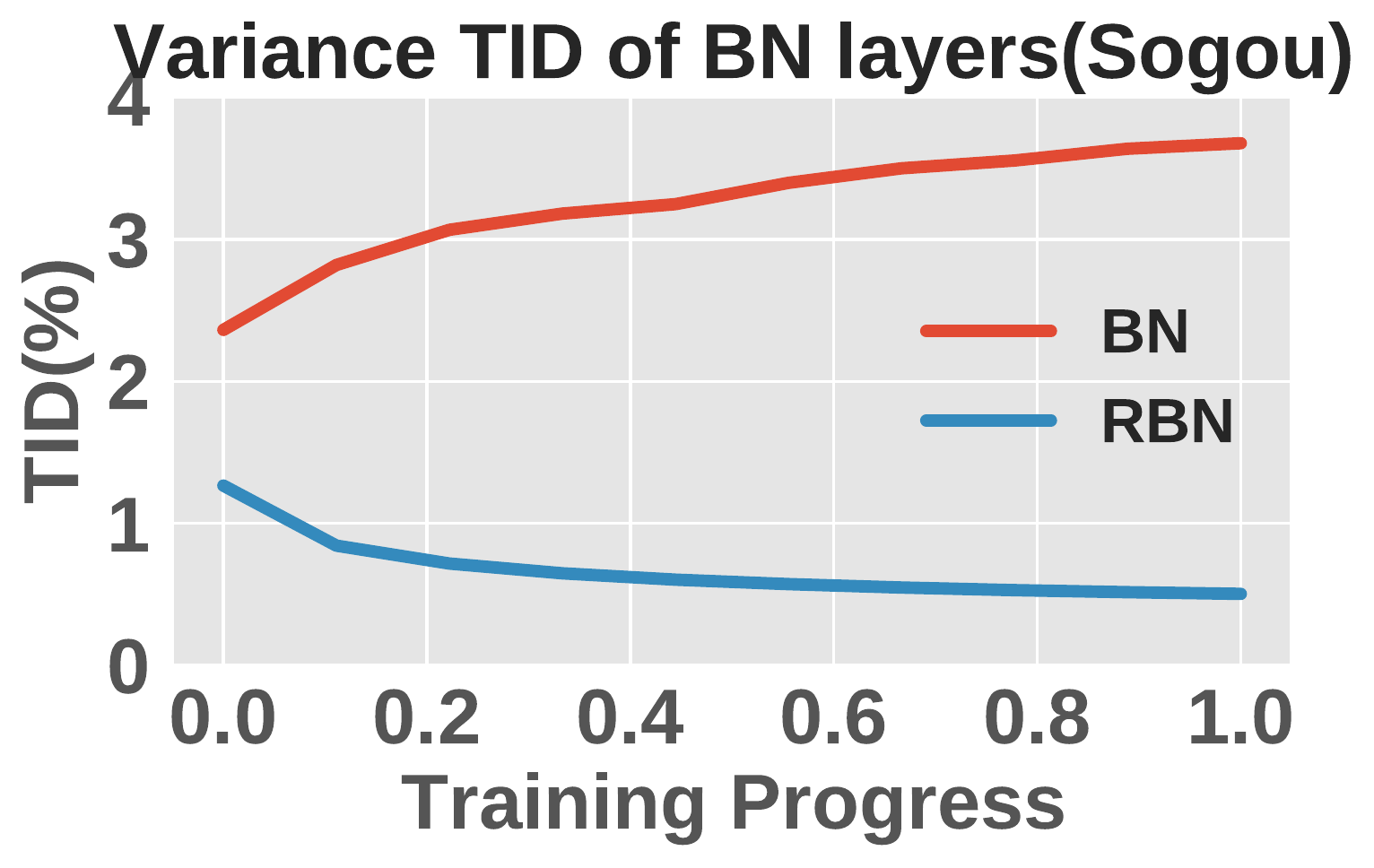}
	\end{subfigure}
	\hfill
	\begin{subfigure}[b]{0.24\columnwidth}
		\centering
		\includegraphics[width=\textwidth]{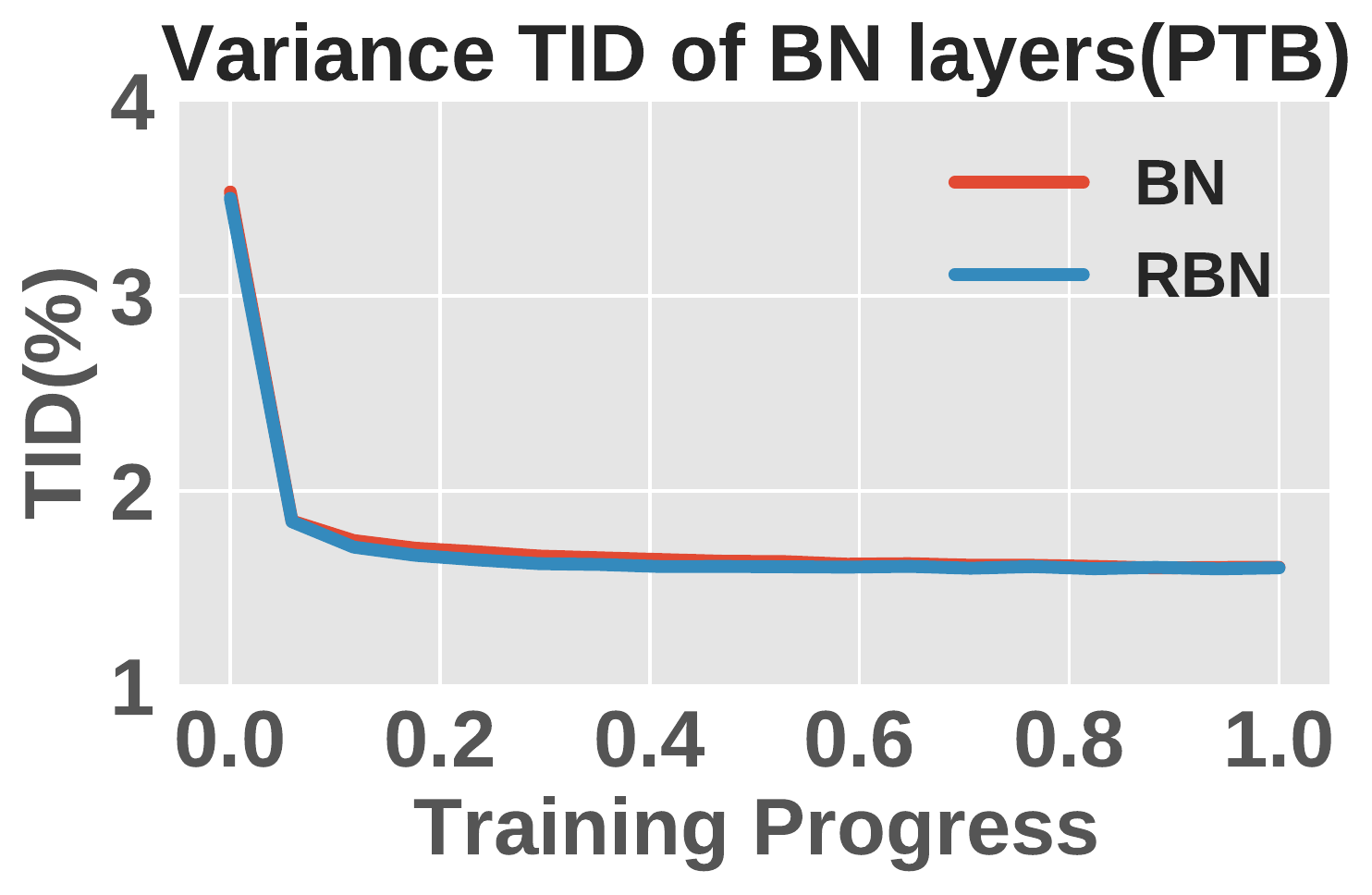}
	\end{subfigure}
	\caption{Averaged Mean and Variance TID on IWSLT14/Resume/Sogou/PTB for Pre-Norm Transformer with BN and RBN.}
	\label{fig:mean_var_tid_four_other_pre}
\end{figure}

\begin{figure}[h]
	\captionsetup[subfigure]{justification=centering}
	\centering
	\begin{subfigure}[b]{0.24\columnwidth}
		\centering
		\centerline{\includegraphics[width=\textwidth]{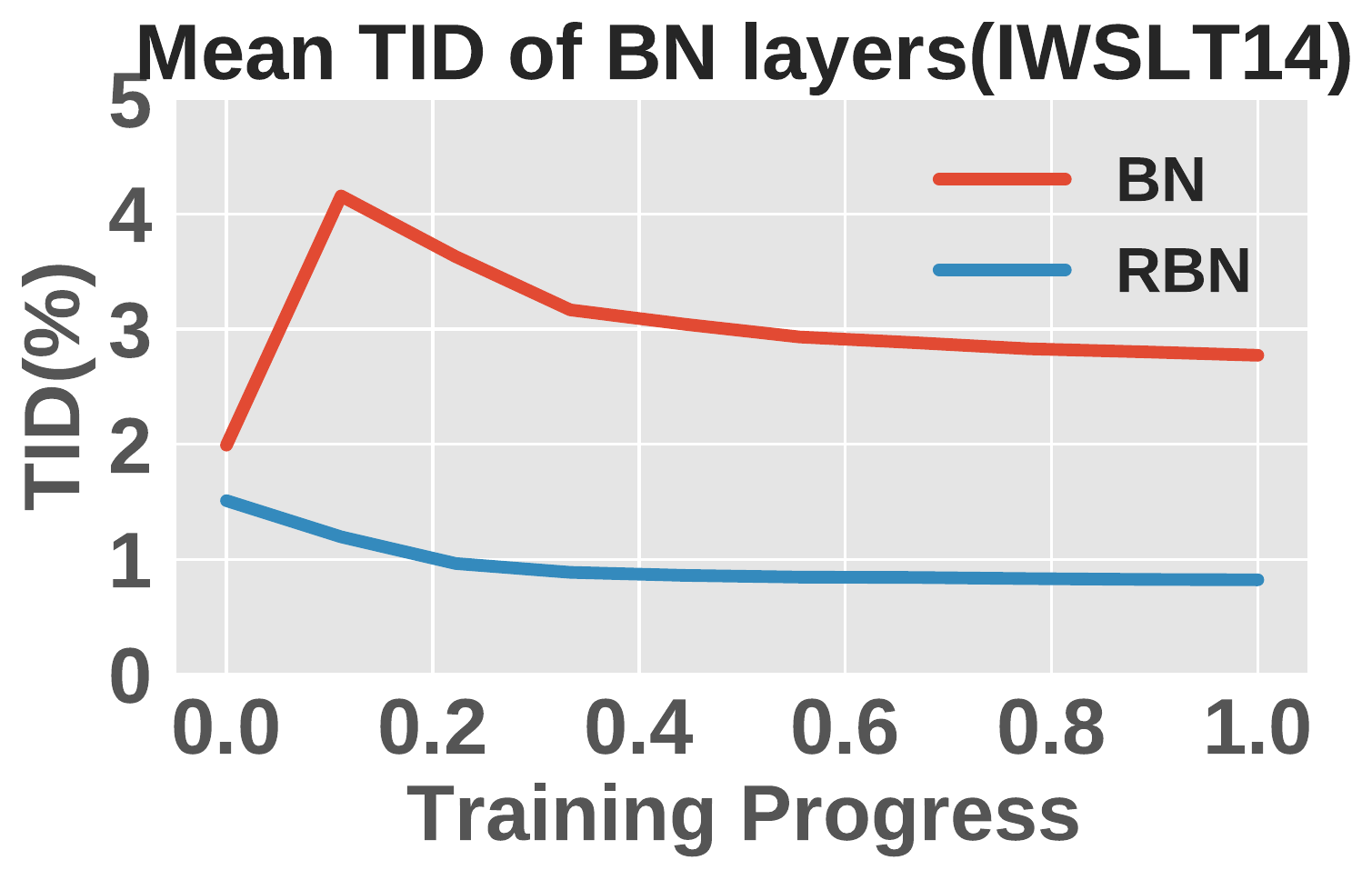}}
	\end{subfigure}
	\hfill
	\begin{subfigure}[b]{0.24\columnwidth}
		\centering
		\includegraphics[width=\textwidth]{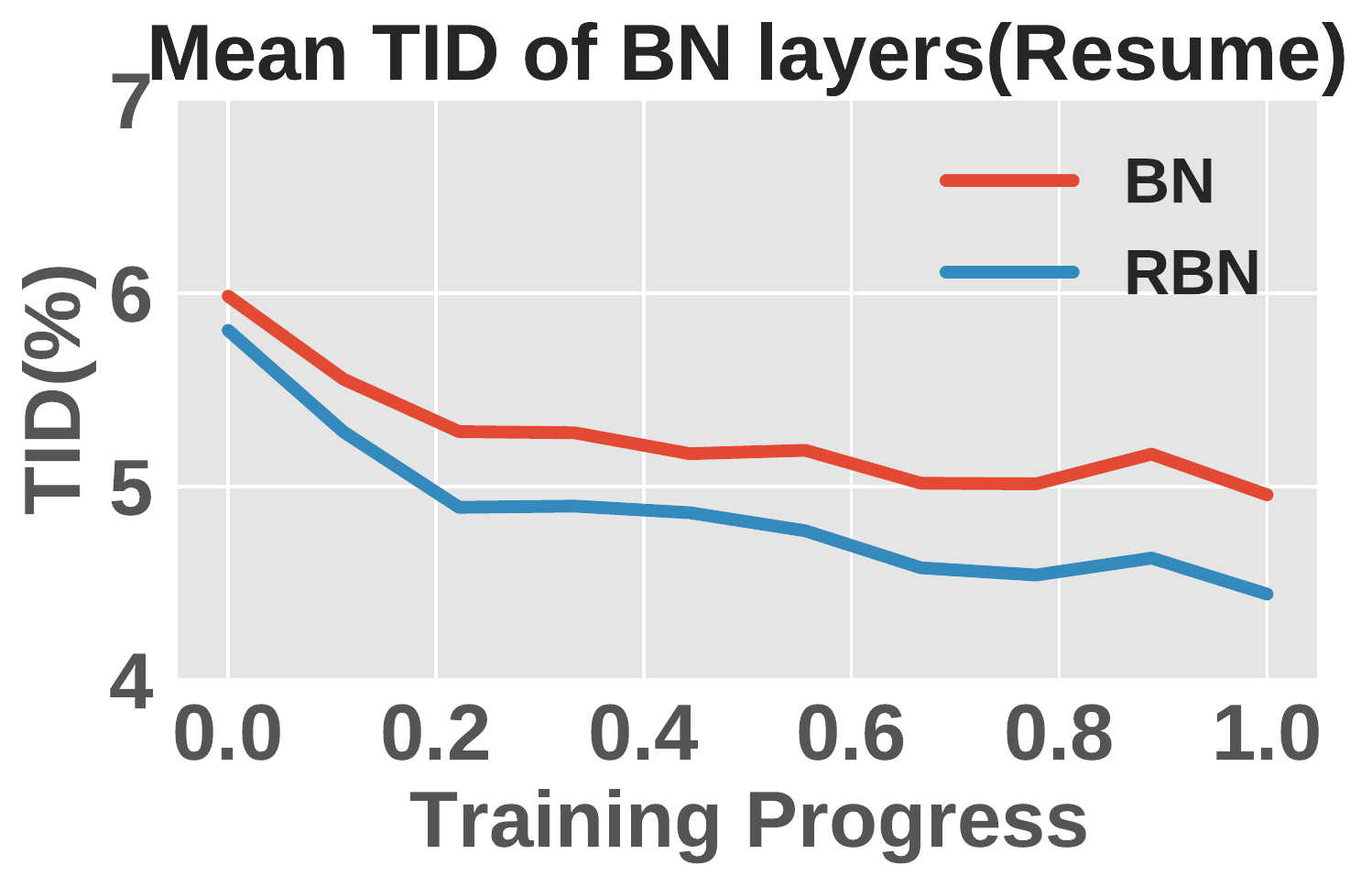}
	\end{subfigure}
	\hfill
	\begin{subfigure}[b]{0.24\columnwidth}
		\centering
		\includegraphics[width=\textwidth]{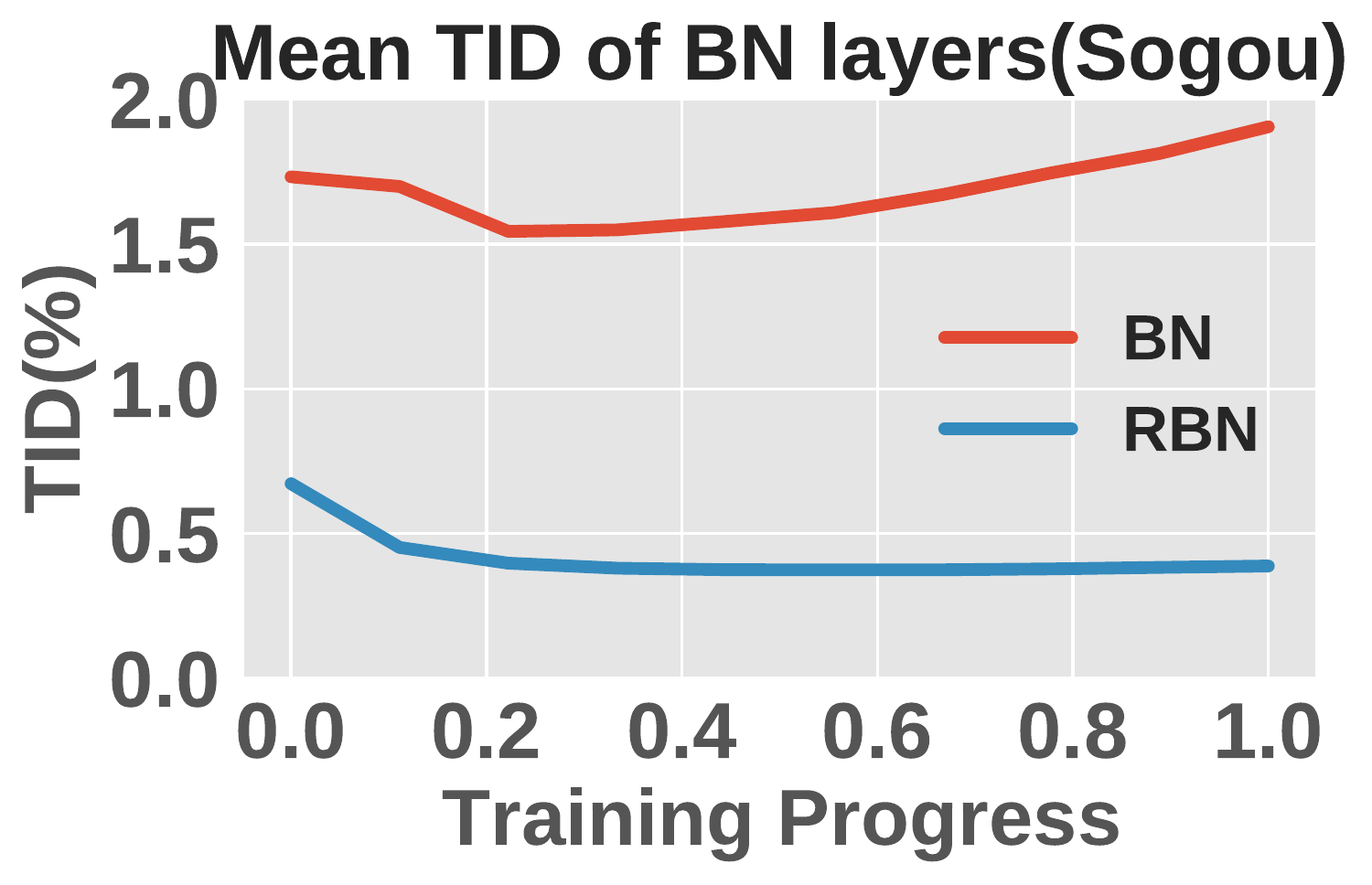}
	\end{subfigure}
	\hfill
	\begin{subfigure}[b]{0.24\columnwidth}
		\centering
		\includegraphics[width=\textwidth]{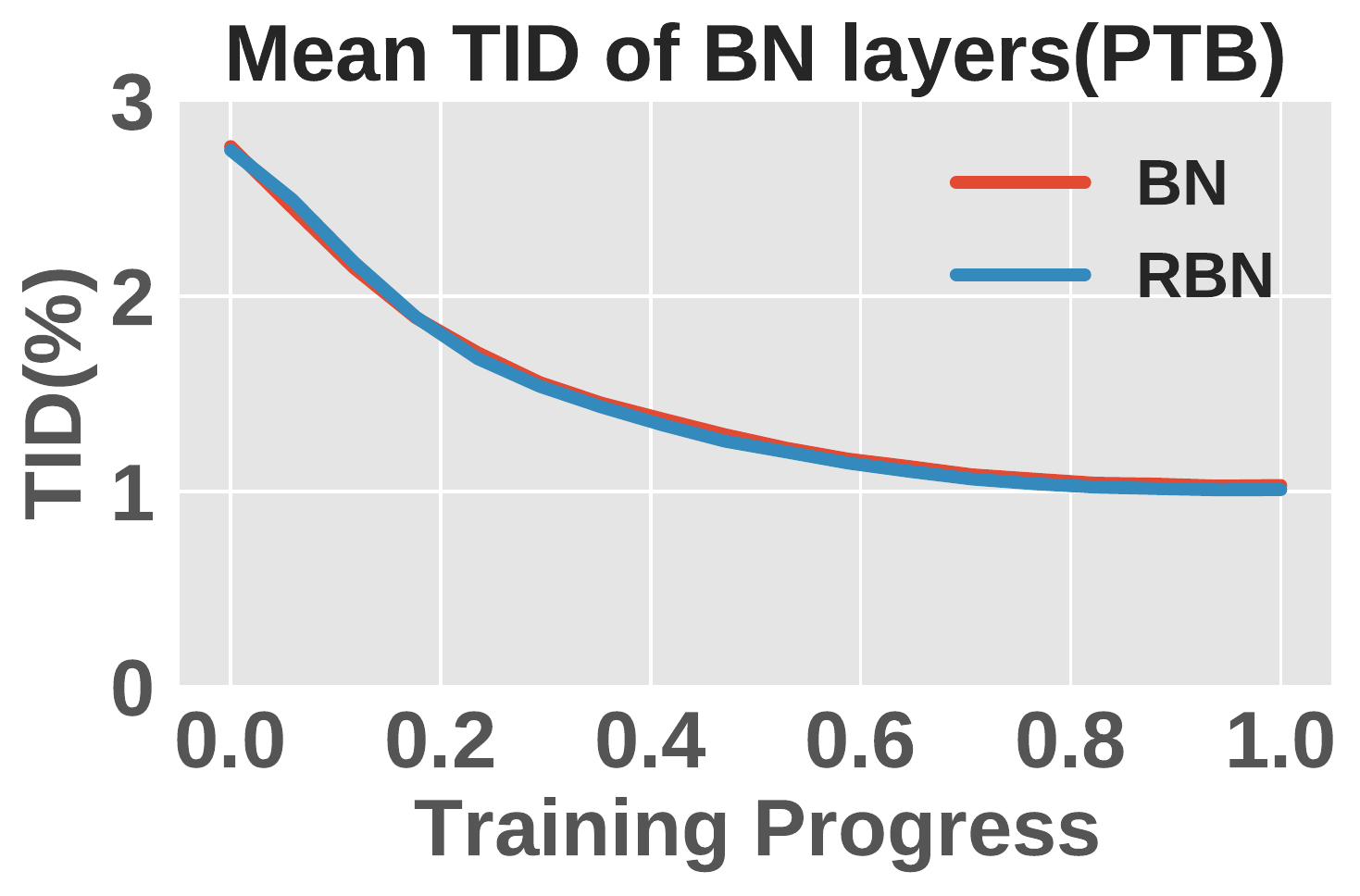}
	\end{subfigure}
	\begin{subfigure}[b]{0.24\columnwidth}
		\centering
		\centerline{\includegraphics[width=\textwidth]{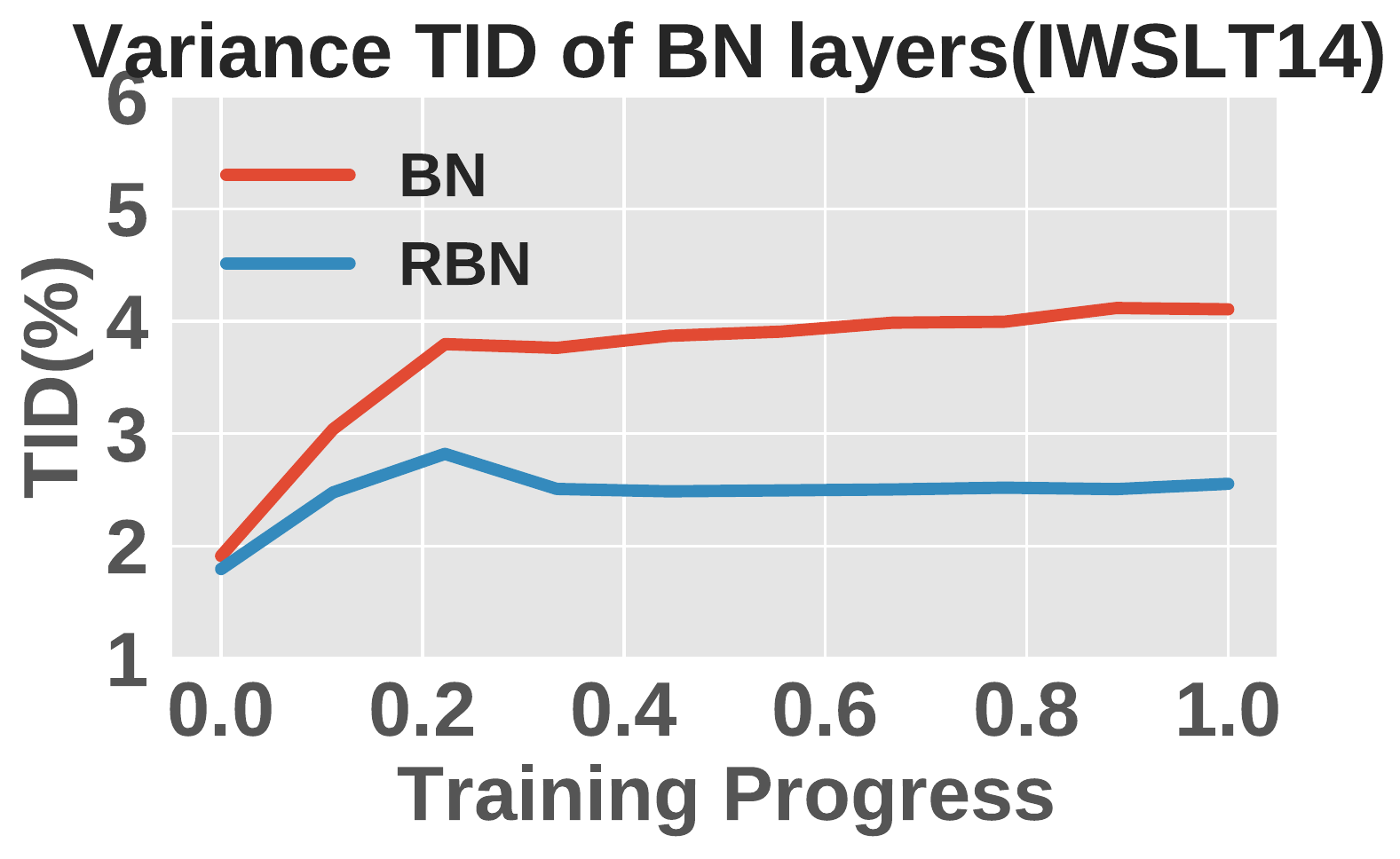}}
	\end{subfigure}
	\hfill
	\begin{subfigure}[b]{0.24\columnwidth}
		\centering
		\includegraphics[width=\textwidth]{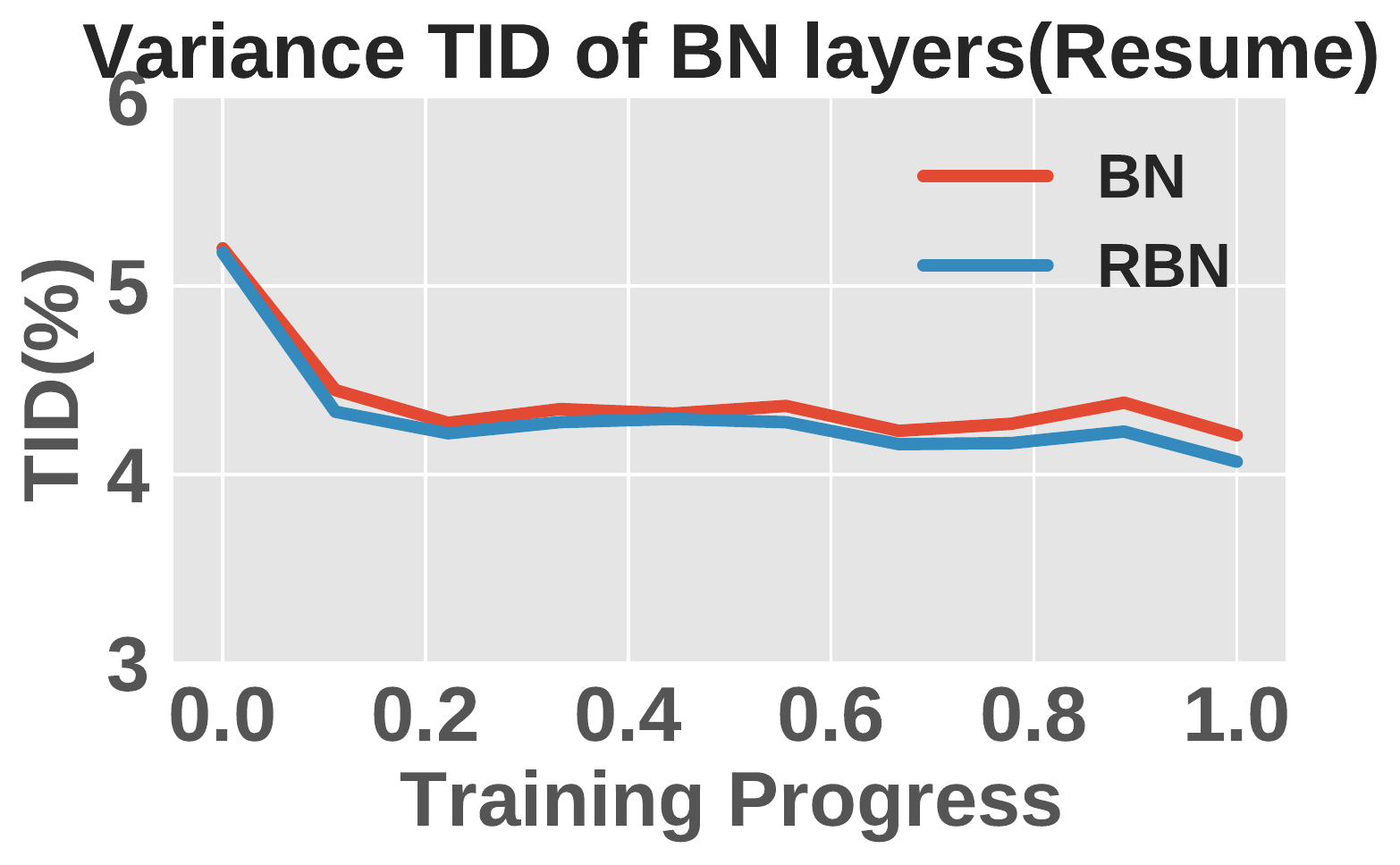}
	\end{subfigure}
	\hfill
	\begin{subfigure}[b]{0.24\columnwidth}
		\centering
		\includegraphics[width=\textwidth]{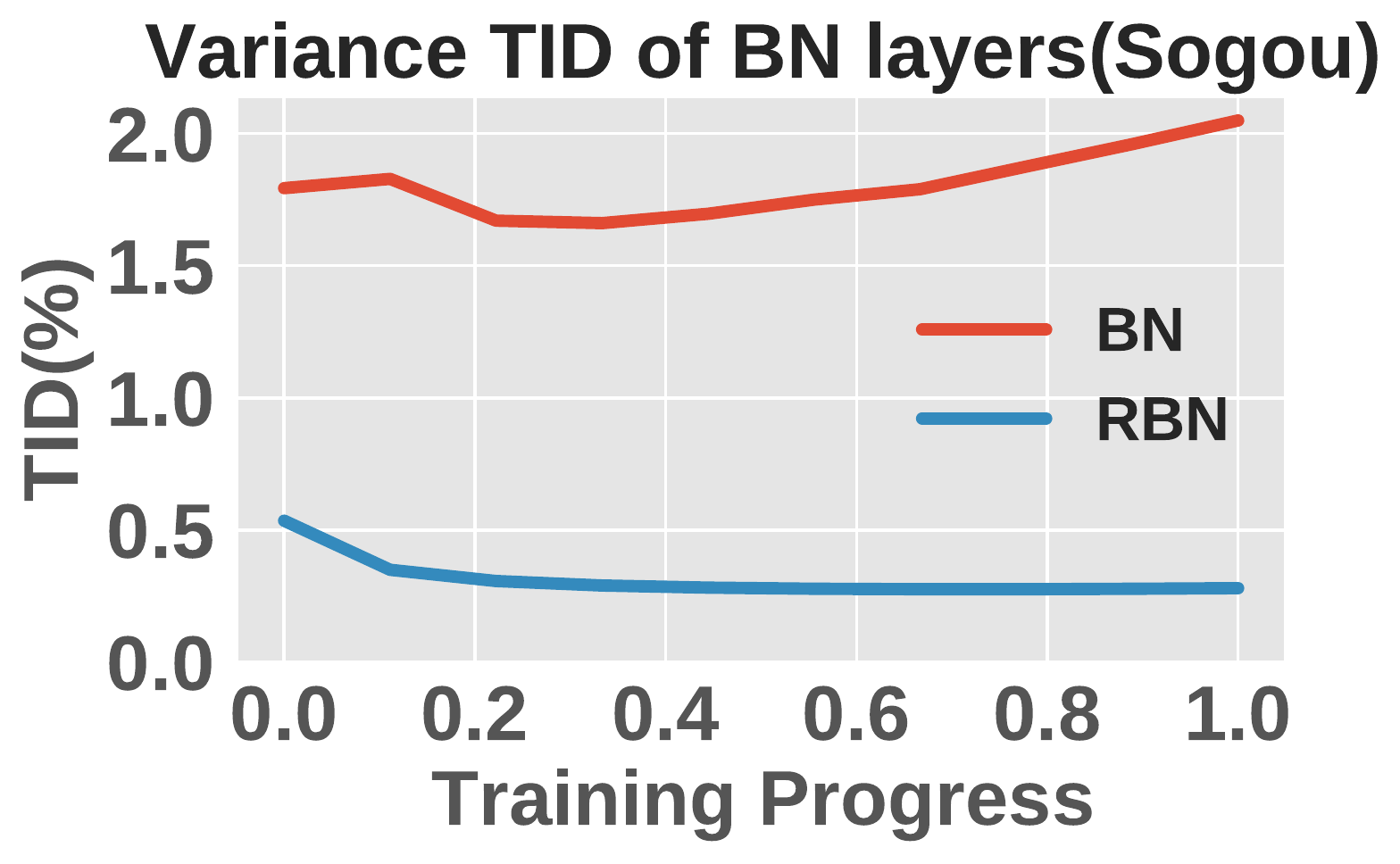}
	\end{subfigure}
	\hfill
	\begin{subfigure}[b]{0.24\columnwidth}
		\centering
		\includegraphics[width=\textwidth]{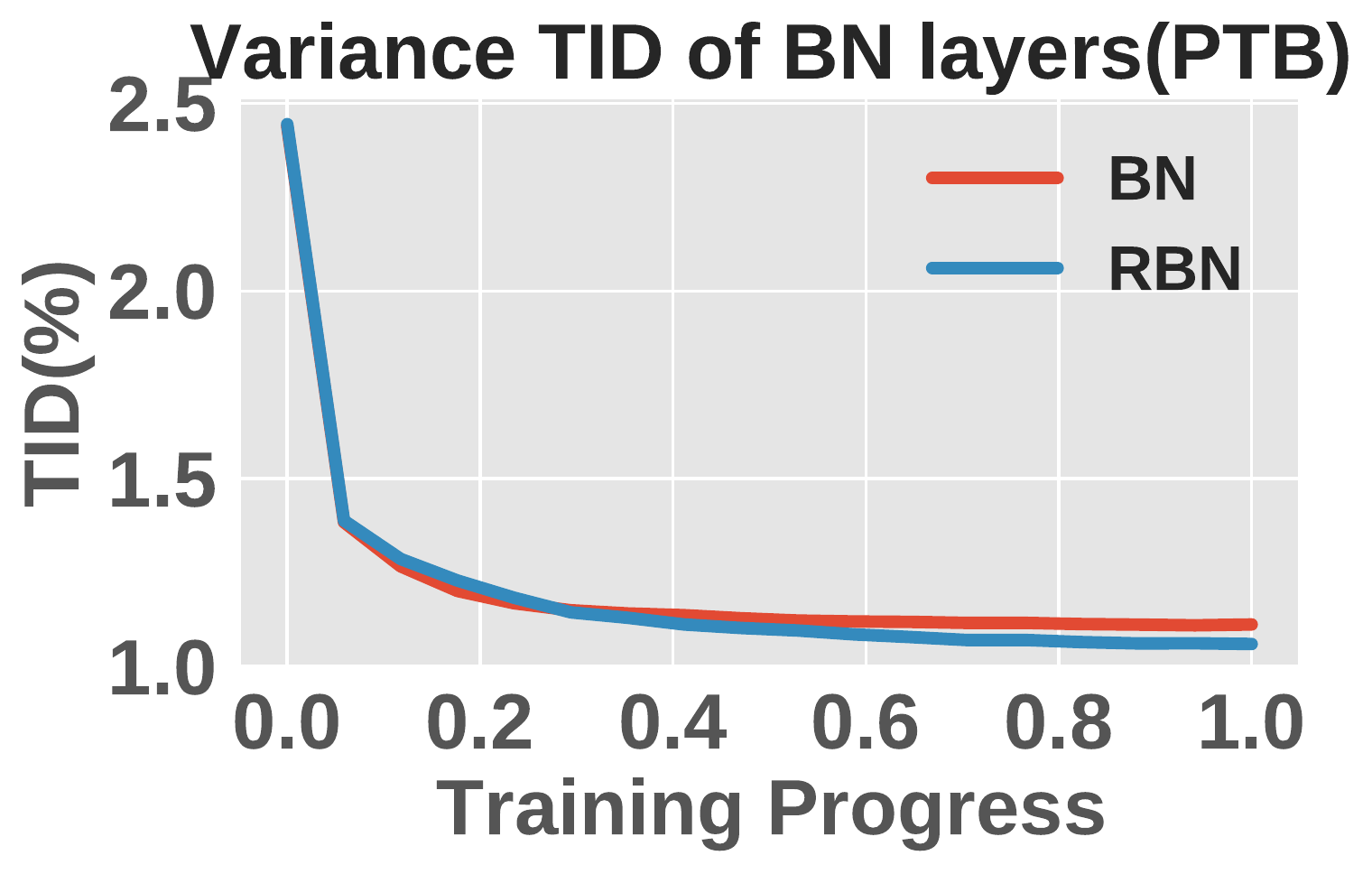}
	\end{subfigure}
	\caption{Averaged Mean and Variance TID on IWSLT14/Resume/Sogou/PTB for Post-Norm Transformer with BN and RBN.}
	\label{fig:mean_var_tid_four_other_post}
\end{figure}

\begin{figure}[h]
	
	\captionsetup[subfigure]{justification=centering}
	\centering
	\begin{subfigure}[b]{0.24\columnwidth}
		\centering
		\centerline{\includegraphics[width=\textwidth]{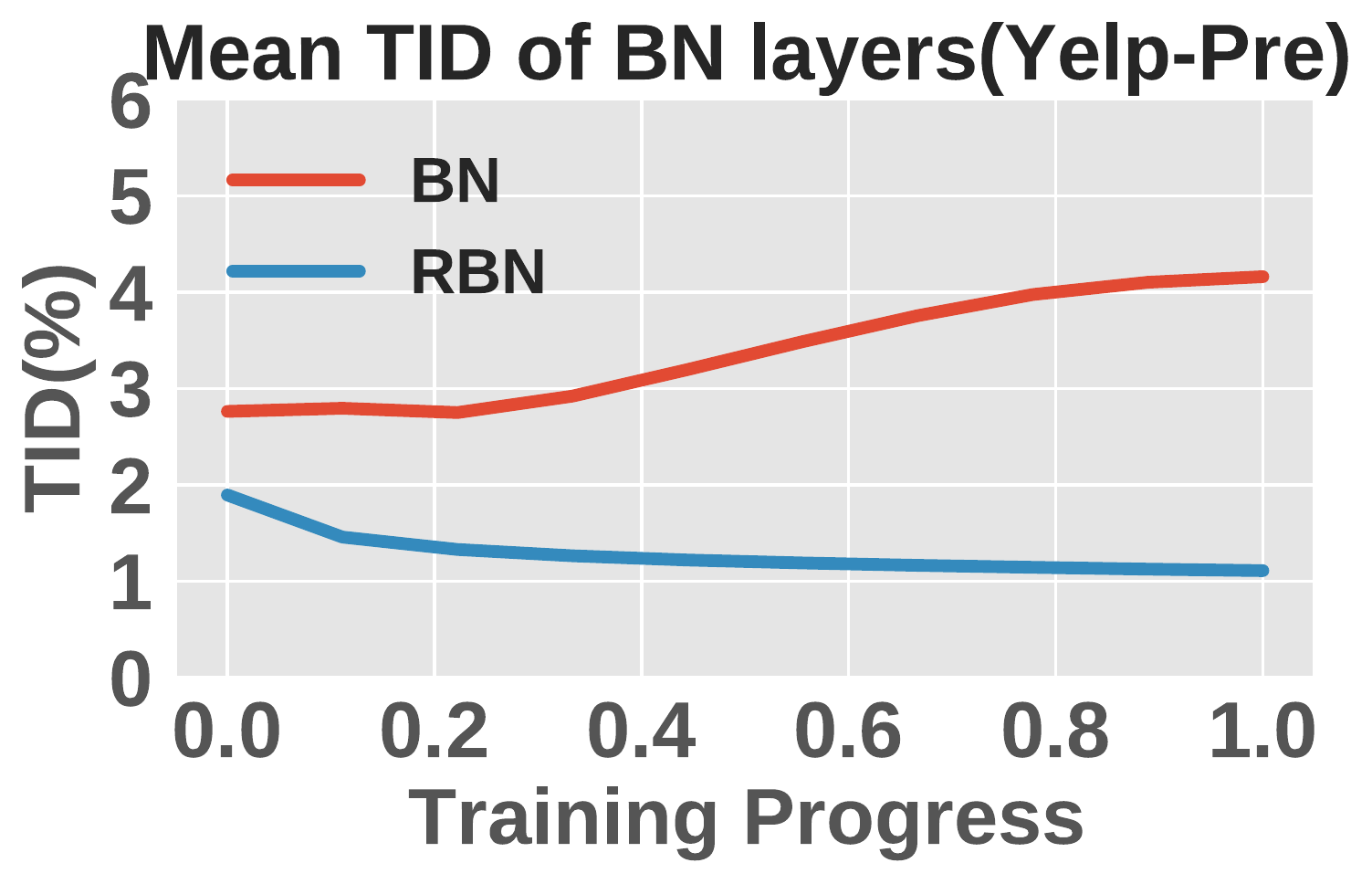}}
	\end{subfigure}
	\hfill
	\begin{subfigure}[b]{0.24\columnwidth}
		\centering
		\includegraphics[width=\textwidth]{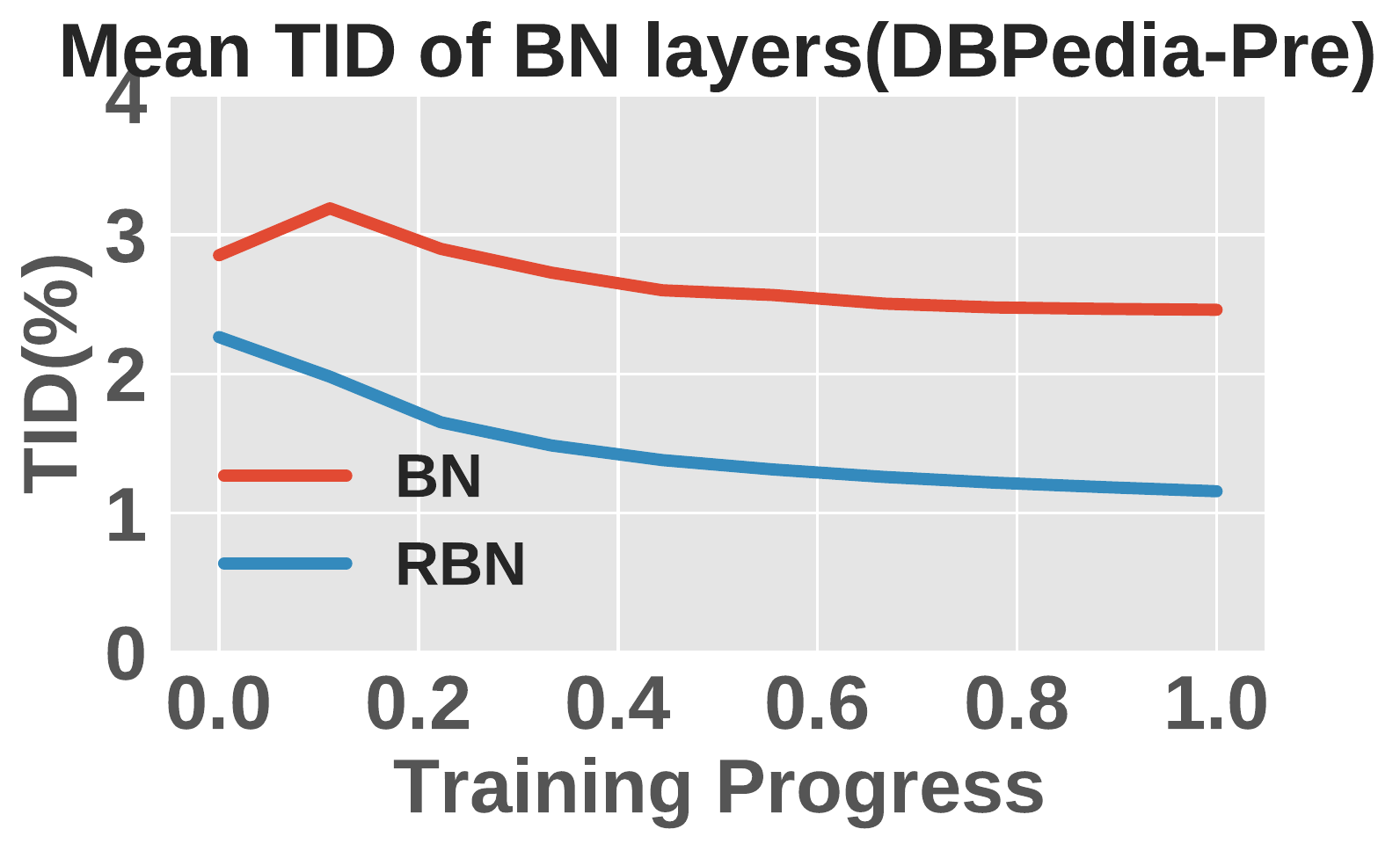}
	\end{subfigure}
	\hfill
	\begin{subfigure}[b]{0.24\columnwidth}
		\centering
		\includegraphics[width=\textwidth]{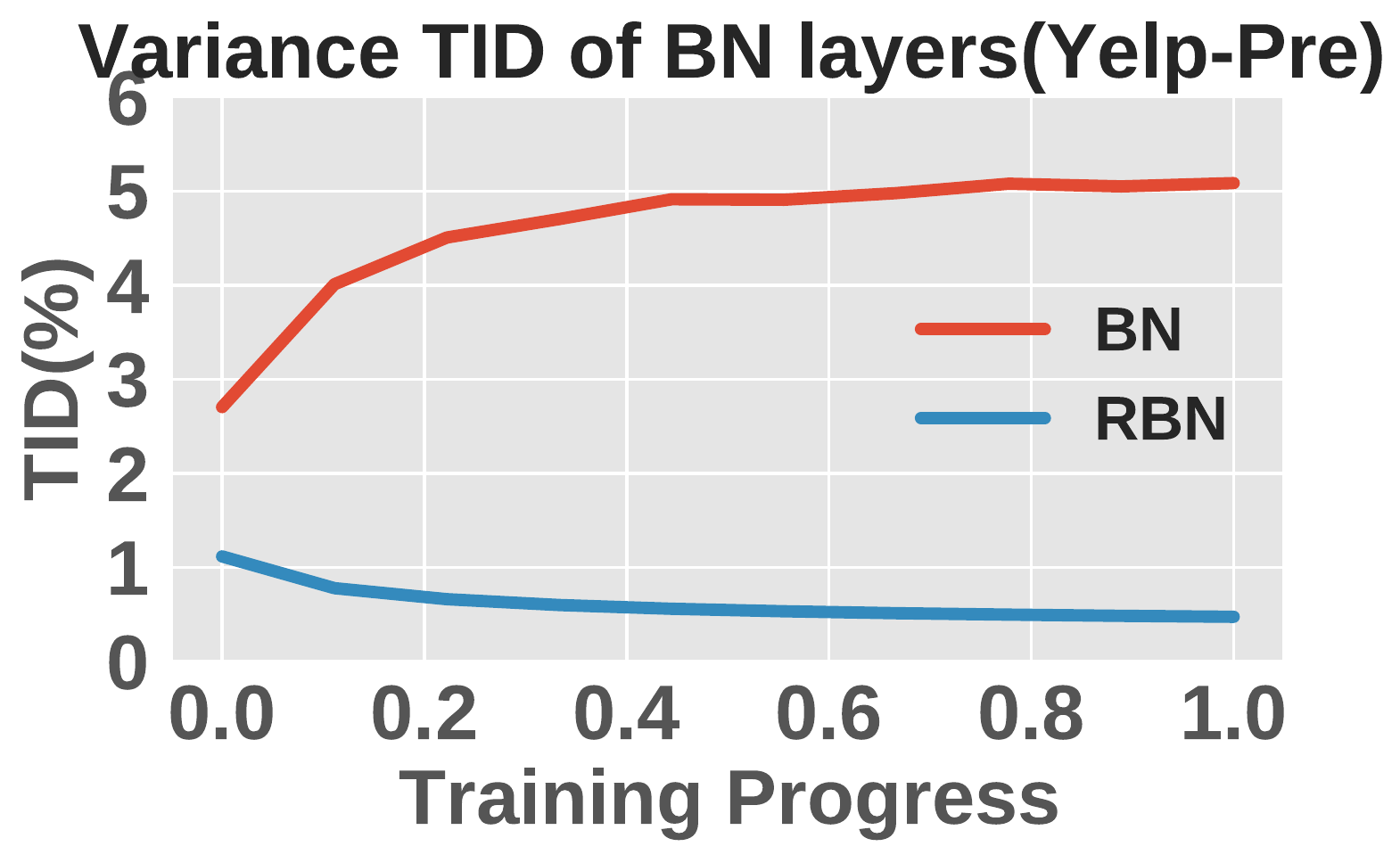}
	\end{subfigure}
	\hfill
	\begin{subfigure}[b]{0.24\columnwidth}
		\centering
		\includegraphics[width=\textwidth]{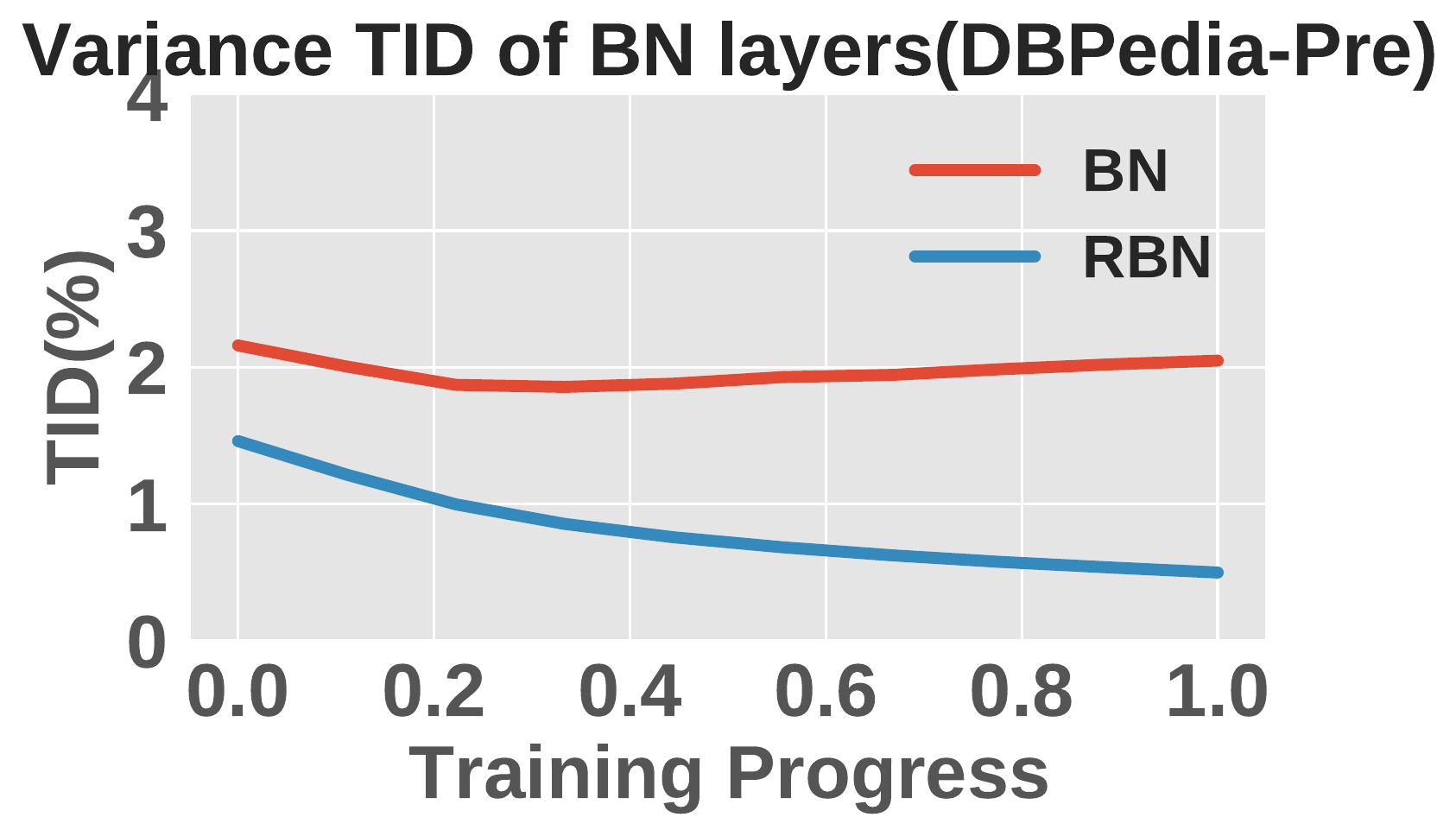}
	\end{subfigure}
	\begin{subfigure}[b]{0.24\columnwidth}
		\centering
		\centerline{\includegraphics[width=\textwidth]{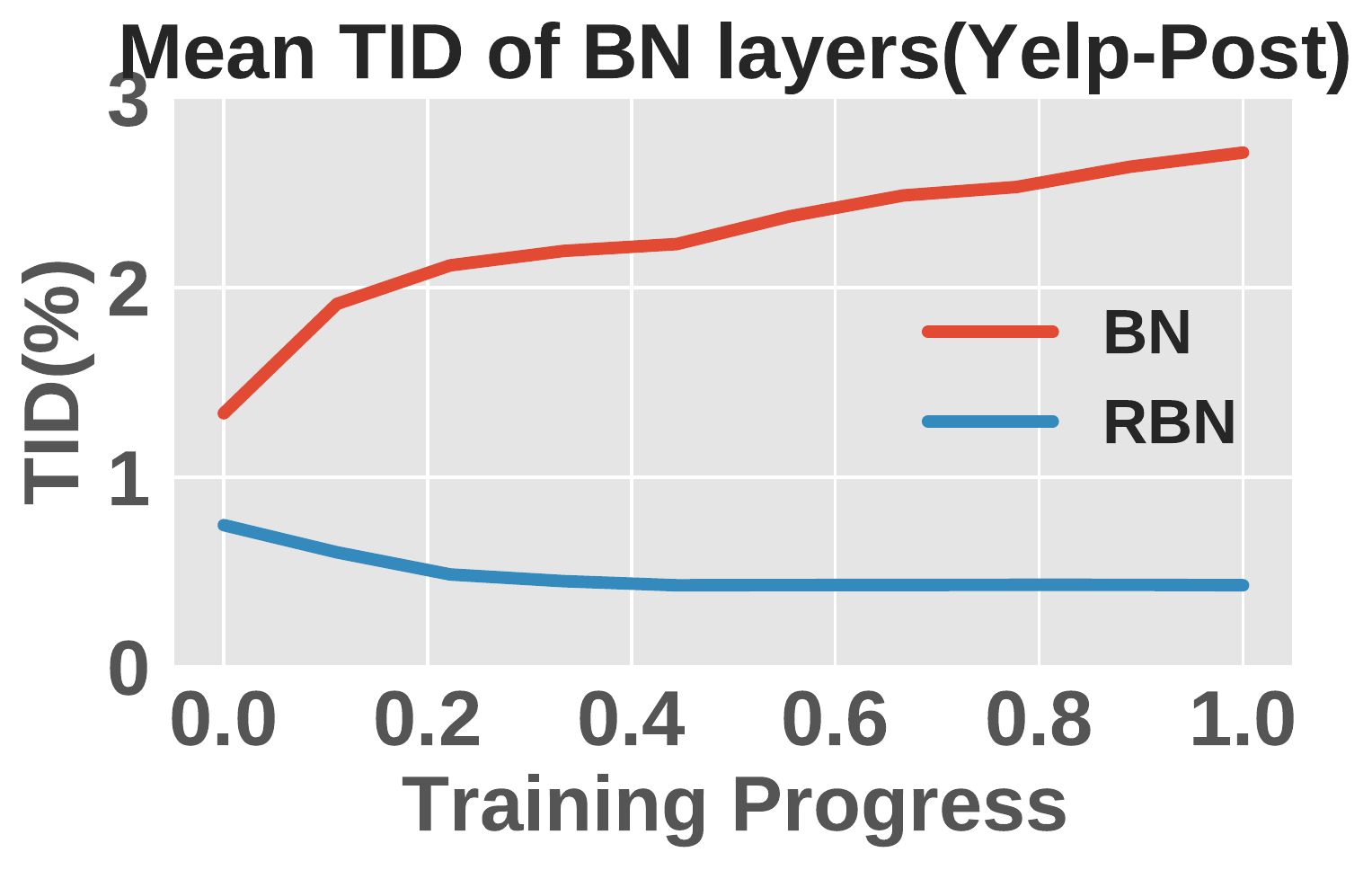}}
	\end{subfigure}
	\hfill
	\begin{subfigure}[b]{0.24\columnwidth}
		\centering
		\includegraphics[width=\textwidth]{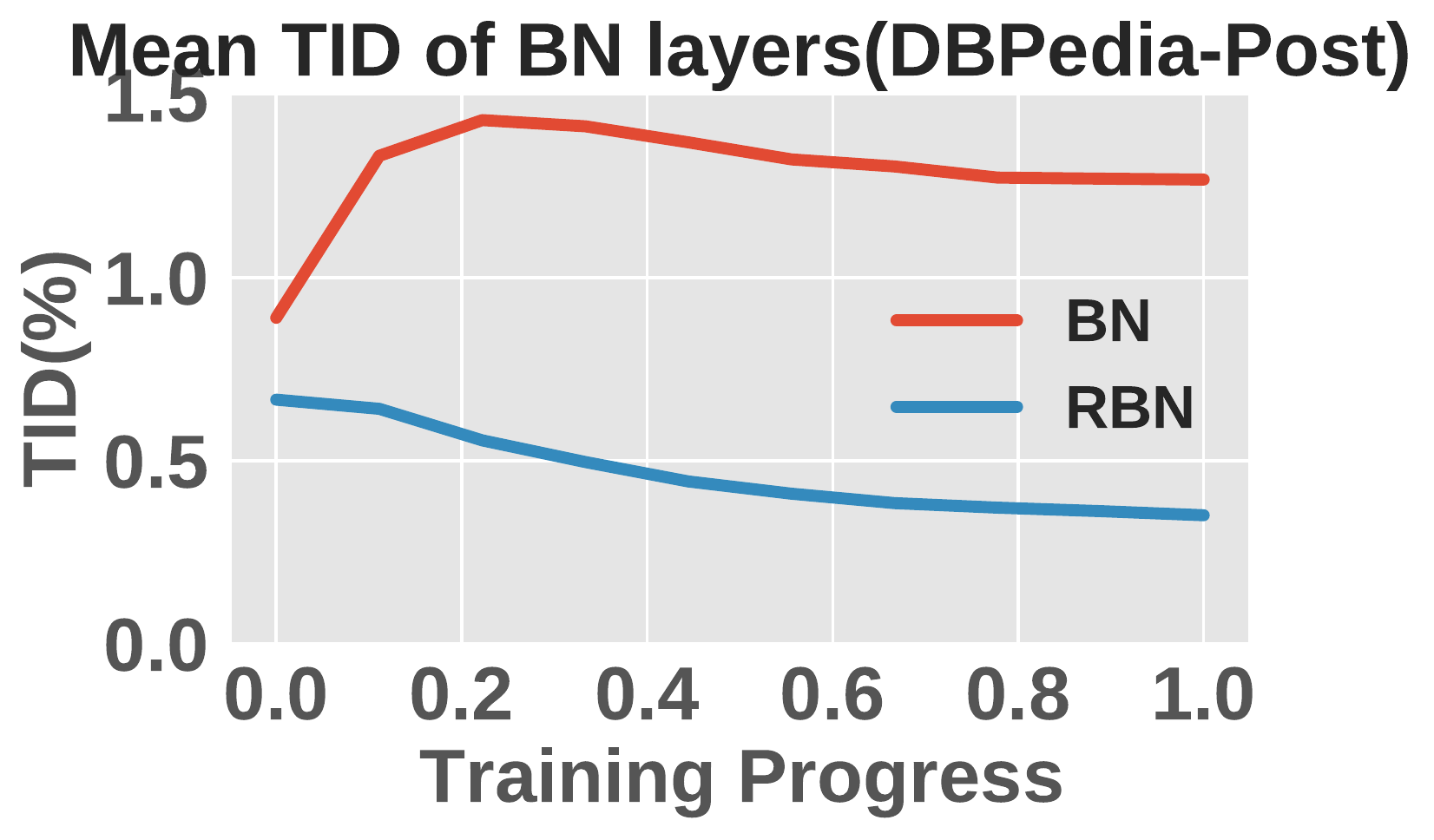}
	\end{subfigure}
	\hfill
	\begin{subfigure}[b]{0.24\columnwidth}
		\centering
		\includegraphics[width=\textwidth]{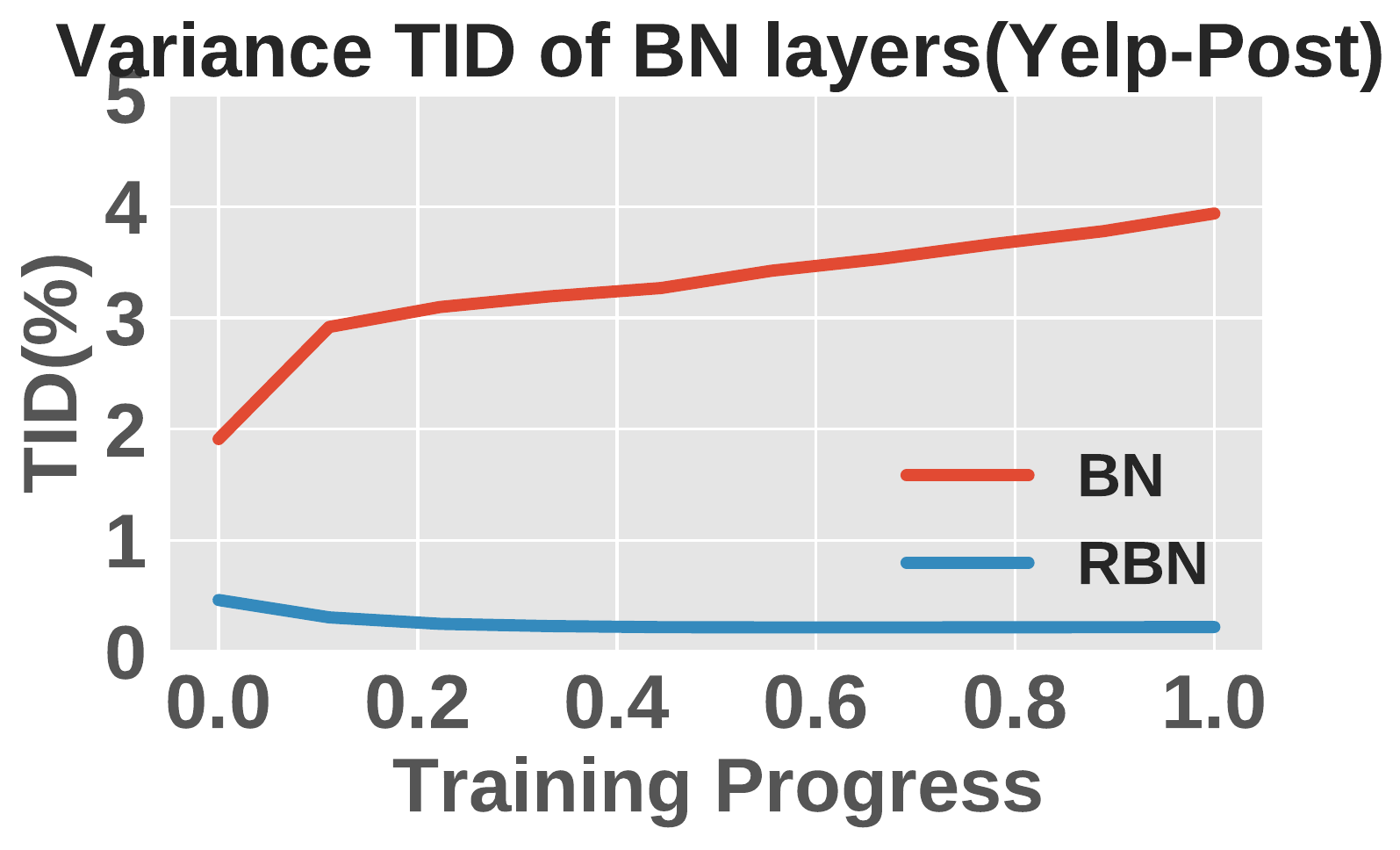}
	\end{subfigure}
	\hfill
	\begin{subfigure}[b]{0.24\columnwidth}
		\centering
		\includegraphics[width=\textwidth]{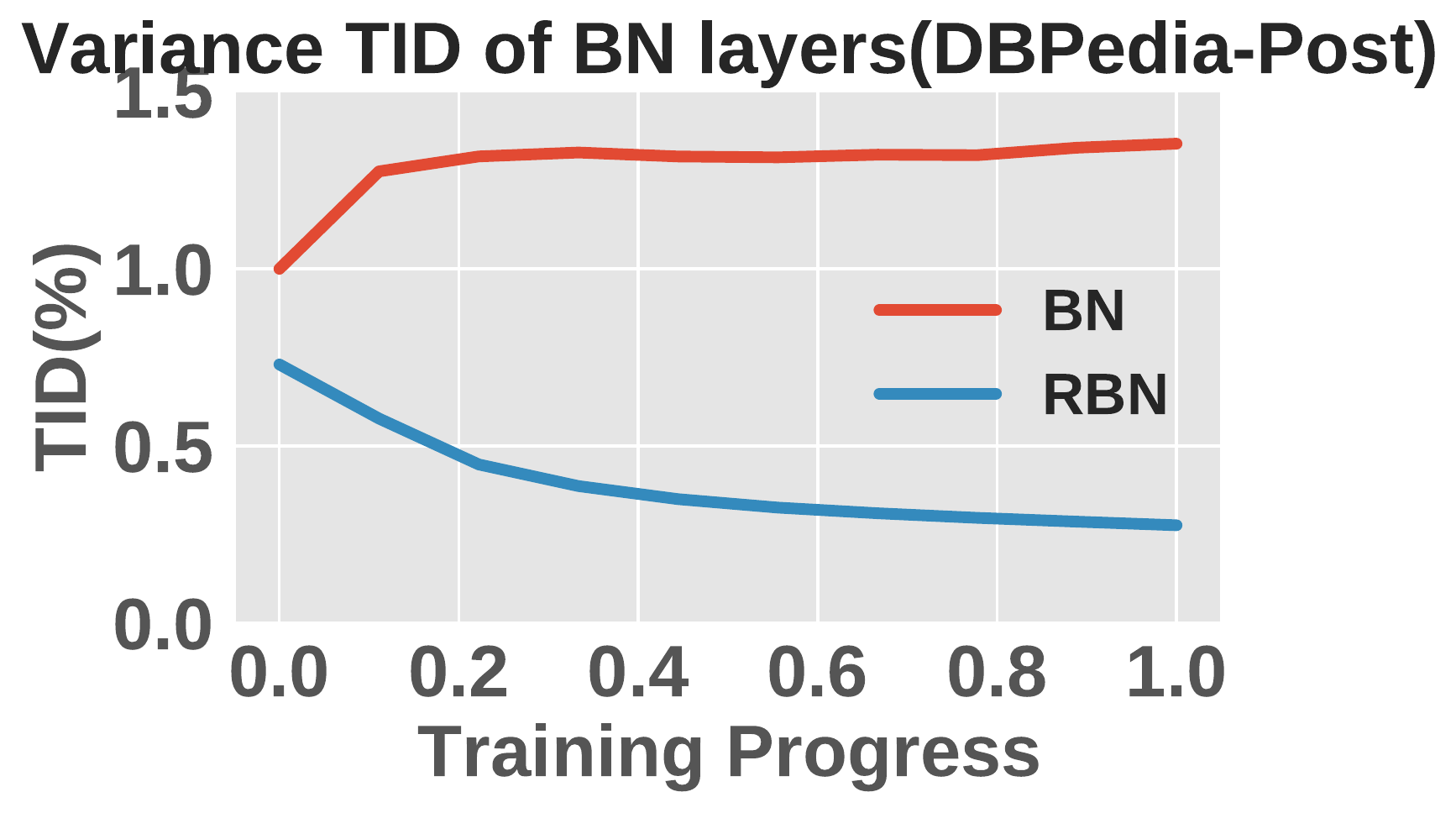}
	\end{subfigure}
	\caption{Averaged Mean and Variance TID on Yelp/DBPedia for Pre-Norm Transformer (upper) and Post-Norm Transformer (bottom) with BN and RBN.}
	\label{fig:mean_var_tid_two_other}
\end{figure}

We have shown the average mean and variance TID on WMT16/CoNLL/IMDB/WT103 for Pre-Norm Transformer with BN and RBN in the main paper. Here, we plot the average mean and variance TID on other datasets with Pre-Norm and Post-Norm Transformers in \cref{fig:mean_var_tid_post,fig:mean_var_tid_four_other_pre,fig:mean_var_tid_four_other_post,fig:mean_var_tid_two_other}. RBN reduces the TID of BN effectively.

\section{Other settings of BN in Test Stage}
\label{sec:bn_test_variant}
In the main paper, we test BN (RBN) with population statistics estimated by Exponential Moving Average (EMA). Here, we test two other settings of BN on IWSLT14 dataset. In the first setting (\cref{tab:first}), we reestimate the population statistics of BN by running two more epochs with zero learning rate. In the second setting (\cref{tab:second}), we test BN with batch statistics of different effective batch size (max-tokens). \par
From \cref{tab:first}, we can see that reestimating the population statistics leads to similar results as EMA. However, using batch statistics instead of population statistics boosts the performance of Post-Norm $\text{Transformer}_{BN}$, but hurts the performance of Pre-Norm $\text{Transformer}_{BN}$ and Post-Norm $\text{Transformer}_{RBN}$ (\cref{tab:second}). Pre-Norm $\text{Transformer}_{RBN}$ is robust to different max-tokens. Note that $\text{Transformer}_{LN}$ achieves 35.5 BLEU in both Pre-Norm and Post-Norm settings. BN can not match the performance of LN by changing the test settings.
\begin{table}[h]
	\centering
	\caption{EMA: Use population statistics estimated by EMA. Reestimation: Run two more epochs with zero learning rate to update population statistics. The performance metric is BLEU. }
	\begin{tabular}{@{}ccccc@{}}
		\toprule
		& Post-BN & Post-RBN & Pre-BN & Pre-RBN \\ \midrule
		EMA        & 34.0    & 35.5     & 34.8   & 35.6    \\
		Reestimation & 33.8    & 35.6     & 34.9   & 35.4    \\ \bottomrule
	\end{tabular}
	\label{tab:first}
\end{table}

\section{Configurations of BN's Variants}
\label{sec:bn_variants_config}
We compare the performance of RBN with Power Normalization (PN)~\cite{Shen2020Powernorm}, Batch Renormalization (BRN)~\cite{2017_NIPS_Ioffe} and Moving Averaing Batch Normaliazation
(MABN)~\cite{2020_ICLR_Yan} in the main paper. We mainly follow the hyperparameters in their papers. For PN, we use 4000 warmups,
and set foward and backward momentum as $0.9$. For BRN, we use one epoch BN as warmup and linearly increase $r$ to 3 
and $d$ to 5. $r$ and $d$ are
renormalizing factors. For MABN, we use $16$ mini-batches to compute simple moving
average statistics and momentum $\alpha=0.98$ to compute exponential
moving average statistics.

\section{Potential Negative Societal Impact}
\label{sec:social_impact}
We spend many GPU hours on running experiments which may negatively impact the environment. 

\begin{table}[h]
	\centering
	\caption{Testing BN (RBN) with batch statistics of different max-tokens. }
	\begin{tabular}{@{}ccccc@{}}
		\toprule
		Max-tokens & Post-BN & Post-RBN & Pre-BN & Pre-RBN \\ \midrule
		EMA        & 34.0    & 35.5     & 34.8   & 35.6    \\
		512        & 34.8    & 34.9     & 27.2   & 35.6    \\
		1024       & 35.1    & 34.9     & 30.9   & 35.6    \\
		2048       & 35.2    & 34.9     & 32.3   & 35.6    \\
		4096       & 35.2    & 34.9     & 32.7   & 35.6    \\
		8192       & 35.2    & 34.9     & 32.8   & 35.6    \\
		16384      & 35.2    & 34.8     & 32.9   & 35.6    \\ \bottomrule
	\end{tabular}
	\label{tab:second}
\end{table}



\end{document}